%% file: wise_arxiv.tex
\documentclass[10pt,twocolumn,letterpaper]{article}

\usepackage[pagenumbers]{./paper_styles/paper_style} % To force page numbers, e.g. for an arXiv version

\input{paper_styles/preamble}
\definecolor{cvprblue}{rgb}{0.21,0.49,0.74}
\usepackage[pagebackref,breaklinks,colorlinks,allcolors=cvprblue]{hyperref}

\usepackage{epigraph} 
\usepackage[utf8]{inputenc} % allow utf-8 input
\usepackage[T1]{fontenc}    % use 8-bit T1 fonts
\usepackage{hyperref}       % hyperlinks
\usepackage{url}            % simple URL typesetting
\usepackage{booktabs}       % professional-quality tables
\usepackage{amsfonts}       % blackboard math symbols
\usepackage{nicefrac}       % compact symbols for 1/2, etc.
\usepackage{microtype}      % microtypography
\usepackage{xcolor}         % colors
\usepackage{xspace}
\usepackage{graphicx}
\usepackage{amsmath,amsthm}
\usepackage{tabularx}     % in preamble
\usepackage{array}        % for \newcolumntype
\newcolumntype{L}[1]{>{\raggedright\arraybackslash}p{#1}} % fixed-width left
\newcolumntype{Y}{>{\raggedright\arraybackslash}X}    % wrapping left
\usepackage{subcaption}
\usepackage{multirow}
\usepackage{multicol}
\usepackage{makecell}

\usepackage[utf8]{inputenc}
\usepackage[T1]{fontenc}
\usepackage{listingsutf8}
\usepackage{listings}
\include{definitions}

\input{math_commands.tex}

\usepackage{hyperref}
\usepackage{url}

\title{WISE: Weighted Iterative Society-of-Experts for\\
Robust Multimodal Multi-Agent  Debate}

\author{%
 Anoop Cherian$^1$\quad River Doyle$^2$\thanks{Equal Contribution.}\quad  Eyal Ben-Dov$^{2*}$\quad  Suhas Lohit$^1$\quad  Kuan-Chuan Peng$^1$\\
 $^1$Mitsubishi Electric Research Labs, Cambridge, MA\\ 
 $^2$Cambridge Rindge and Latin School, Cambridge, MA 
}

\begin{document}

\maketitle

\begin{abstract}
Recent large language models (LLMs) are trained on diverse corpora and tasks, leading them to develop complementary strengths. Multi-agent debate (MAD) has emerged as a popular way to leverage these strengths for robust reasoning, though it has mostly been applied to language-only tasks, leaving its efficacy on multimodal problems underexplored. In this paper, we study MAD for solving vision-and-language reasoning problems. Our setup enables generalizing the debate protocol with heterogeneous experts that possess single- and multi-modal capabilities. To this end, we present \fullname (\name), a generalized and modular MAD framework that partitions the agents into \emph{Solvers}, that generate solutions, and \emph{Reflectors}, that verify correctness, assign weights, and provide natural language feedback. To aggregate the agents’ solutions across debate rounds, while accounting for variance in their responses and the feedback weights, we present a modified Dawid–Skene algorithm for post-processing that integrates our two-stage debate model. We evaluate \name on SMART-840, VisualPuzzles, EvoChart-QA, and a new SMART-840++ dataset with programmatically generated problem instances of controlled difficulty. Our results show that \name consistently improves accuracy by 2–7\% over the state-of-the-art MAD setups and aggregation methods across diverse multimodal tasks and LLM configurations.
\end{abstract}

\section{Introduction}
\epigraph{``\small{\emph{What magical trick makes us intelligent? The trick is that there is no trick. The power of intelligence stems from our vast diversity, not from any single, perfect principle.}}''}{\textit{Marvin Minsky, The Society of Mind, p. 308}}

\begin{figure}[t]
\centering
\includegraphics[width=8cm,trim={1cm 2.6cm 1.5cm 6cm},clip]{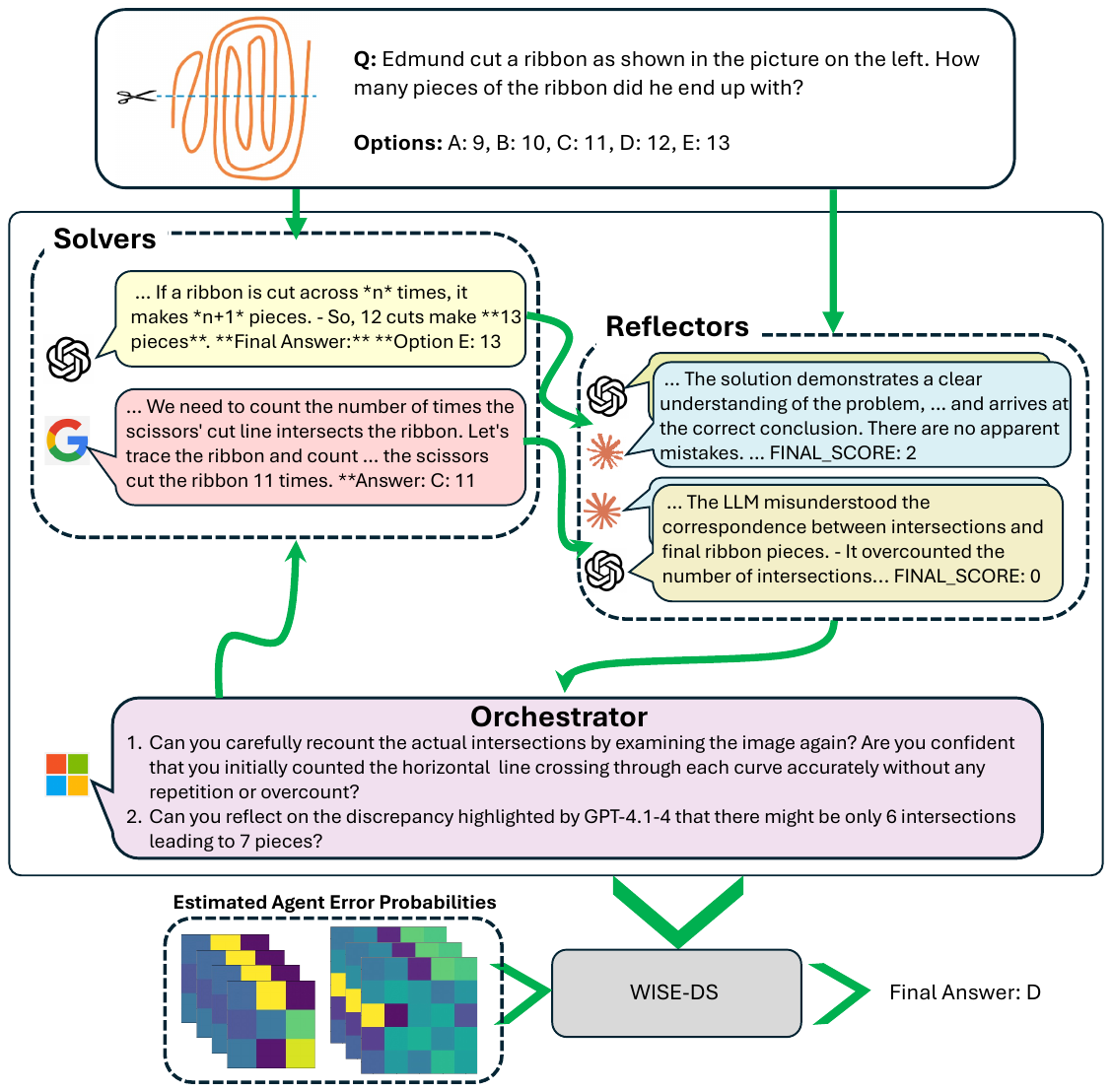}
\caption{An illustration of \name message passing on a problem from the SMART-840 dataset.}
\label{fig:first_page}
\end{figure}
In recent years, large language models (LLMs) have been widely adopted as agents for a broad spectrum of tasks, spanning everyday applications to domains requiring advanced capabilities such as scientific discovery and mathematical reasoning~\cite{luo2025llm4sr,taylor2022galactica,ahn2024large}. Alongside these advances, LLMs have also revealed remarkable signs of extreme generalization~\cite{bubeck2023sparks,pfister2025understanding,chollet2025arc}. A central challenge in making LLM agents effective for reasoning tasks lies in ensuring that their chains of thought (CoT) are both logical and correct. However, due to a range of internal and external factors—including spurious data correlations, sensitivity to sampling temperature, computational errors, and context-length limitations—their predictions are often brittle.

To address this, a variety of ensembling mechanisms have been explored~\cite{chen2023reconcile,wang2022self,shinn2023reflexion,liang2023encouraging}. Multi-agent debate (MAD)~\cite{du2023improving,liang2023encouraging} generalizes these approaches by allowing single or heterogeneous LLMs to critique and refine each other’s solutions, thereby reducing errors and improving final responses. While MAD has shown considerable benefits in purely language-based tasks, where LLM agents are relatively mature, its potential for enhancing multimodal large language models (MLLMs) remains largely unexplored. Given that MLLMs still lag significantly behind LLMs on popular benchmarks, we investigate whether MAD can improve reasoning in multimodal problem solving, specifically those involving vision and language. 

The prospect of applying MAD to multimodal problems introduces both challenges and opportunities, including: (i) improving reasoning across modalities given the weaker abilities of MLLMs in handling non-linguistic data, (ii) integrating LLMs and MLLMs with varying skills (\eg, language-only vs. vision–language) into a cohesive framework, and (iii) exploring whether LLMs can enhance the reasoning capabilities of MLLMs. Inspired by the classical cognitive science view of the mind as a society of agents, we propose \fullname (\name)—a generalized multimodal MAD framework where heterogeneous LLM and MLLM agents collaborate via distributed reasoning to solve complex problems. Each agent is pretrained on diverse corpora for specific downstream tasks (\eg, mathematical reasoning, coding, general knowledge), offering complementary skills that can be harnessed collectively.

A natural approach to designing \name is to adopt frameworks such as RECONCILE~\cite{chen2023reconcile}, where agents are placed in a debate and report their confidence in their proposed solutions. While intuitive, this strategy faces several limitations in the multimodal setting: (i) it assumes all agents have similar multimodal capabilities, (ii) modern MLLMs are often heavily trained and tend to be overly confident—even when hallucinating, (iii) agents with diverse abilities may excel at solving tasks but not at critiquing (or vice versa), and (iv) involving all agents in debate can generate a quadratic number of messages, causing significant computational overhead.

To address these concerns, \name partitions agents into two roles: Solvers and Reflectors (see Figure~\ref{fig:first_page}). Solvers generate reasoning chains to tackle the problem, while Reflectors evaluate these solutions—verifying correctness, assigning weights, and providing feedback. To balance independent reasoning with the benefits of collective intelligence, we introduce a model orchestrator (itself an LLM) that coordinates the exchange of thoughts among agents. Within \name, agents engage in multi-round debates: in each round, Solvers independently propose solutions, Reflectors assess and weigh them, and feedback is provided to correct inferred mistakes. The orchestrator then aggregates the Reflectors’ responses, summarizes feedback, and formulates actionable questions for Solvers to refine their reasoning. This iterative process promotes error correction, robustness, and the emergence of collective reasoning patterns beyond the reach of a single model while allowing partitioning of the agents based on their (multimodal) capabilities.

Another key challenge in MAD design is determining how to extract a final response from agents’ solutions across multiple rounds. A common approach is majority voting, while weighted voting schemes incorporate agents’ self-reported confidence (\eg, via Brier scores or upper-calibrated bounds). However, there remains no principled method for estimating confidence or error probabilities for each agent. To address this, we revisit the classic Dawid–Skene (DS) model~\cite{dawid1979maximum} for estimating agent error matrices. Under the assumption of agent independence, DS employs expectation–maximization over a mixture of multinomial distributions to infer each agent’s error matrices and posterior probabilities. We extend this model to \name, incorporating the joint error of both Solvers and Reflectors in a post-processing phase.

To evaluate the efficacy of \name, we conduct experiments on three datasets: (i) SMART-840~\cite{cherian2024evaluating}, (ii) VisualPuzzles~\cite{song2025visualpuzzles}, and (iii) EvoChart-QA~\cite{huang2025evochart}. The first two focus on vision-and-language (VL) reasoning in a multiple-choice format, while the latter involves free-form short answers. In addition, we propose SMART-840++, an extension of SMART-840 comprising 55 VL problems with programmatically generated puzzle instances at controlled difficulty levels towards assessing the robustness of MAD formulations. Across these datasets and a varied MAD configurations, we show that \name delivers consistent improvements, beating the state-of-the-art results with substantial margins (2–7\%). 

We summarize our key contributions below:
\begin{enumerate}
\item \textbf{Setup:} We study MAD for zero-shot multimodal reasoning problems, a novel and unexplored setting.
\item \textbf{Formulation:} We propose a multimodal MAD framework that partitions agents into possibly overlapping \emph{Solvers} and \emph{Reflectors}, a previously unexplored aspect.
\item \textbf{Solution Aggregation:} We propose the \name-Dawid–Skene algorithm for estimating agent error probabilities and deriving consensus solutions.
\item \textbf{Results:} \name outperforms the state-of-the-art (SOTA) performance across three VL benchmarks, covering both multiple-choice and free-form answer formats.
\item \textbf{Dataset:} We propose SMART-840++, an extension of SMART-840 with new programmatically generated puzzle instances designed to test MLLM MAD formulations under controlled levels of complexity (which will be publicly released on acceptance).
\end{enumerate}

\section{Related Works}
LLM advances expose their fragile outputs, which motivates the methods for improving robustness and consistency. we review the key directions below to contrast with \name.

\noindent\textbf{Self-Correction Methods.} These methods improve responses either by sampling multiple outputs~\cite{wang2022self,liang2023encouraging} or by incorporating self/external feedback. Sampling-based methods seek robustness through majority voting, while feedback-based methods assume that reflection enables models to identify and fix their own errors. Representative methods include Self-Refine~\cite{madaan2023self}, Self-Correction~\cite{welleck2022generating}, Reflexion~\cite{shinn2023reflexion}, Recursively Criticize and Improve~\cite{kim2023language}, RL from AI Feedback~\cite{bai2022constitutional}, and others~\cite{ganguli2023capacity,jiang2023active,dhuliawala2023chain,pan2023logic}. However, empirical studies show that without external feedback, self-correction often fails to improve—and can even harm—performance~\cite{huang2023large,kamoi2024can,valmeekam2023can}.

\noindent\textbf{Multi-Agent Debate.} MAD generalizes correction and feedback by enabling agents to debate solutions, \eg,~\cite{du2023improving} studies debates among identical LLMs, while~\cite{liang2023encouraging} introduces external feedback and a judge model to manage the process. Heterogeneous setups include collaboration/competition schemes~\cite{feng2024don}, secretary-style orchestration~\cite{wang2024rethinking}, and RECONCILE~\cite{chen2023reconcile}, where agents share uncertainty scores and iteratively refine answers. Explain-Analyze-Generate~\cite{gu2025explain} highlights risks of misleading agents and advocates task decomposition. \cite{subramaniam2025multiagent} fine-tunes MAD models for improved convergence. In contrast, \name introduces explicit agent roles (Solvers vs. Reflectors), avoids reliance on self-estimated confidence (which often needs external calibration~\cite{thind2025optimai,khan2024debating}), and naturally supports abstention through weighted scoring.

\noindent\textbf{Mixture of LLMs.} Ensemble-based methods combine multiple LLMs in various ways. Mixture-of-Agents (MoA)~\cite{wang2406mixture} treats ensembles as layers controlled by prompts and gating models, while Self-MoA~\cite{li2025rethinking} selects only top-performing outputs for downstream processing. Other works explore heterogeneous mixtures resolved by majority voting~\cite{li2024more}, or orchestrated systems such as Magnetic-One~\cite{fourney2024magentic}. Unlike them, \name embeds agents in an iterative self-reflective loop, using feedback to progressively refine solutions.

\noindent\textbf{Multimodal Methods.} MAD has rarely been studied in the multimodal setting. For example,~\cite{wang2025multi} uses multimodal MAD for knowledge transfer to smaller models but does not address vision–language reasoning tasks. Process reward models have been applied for multimodal reasoning~\cite{wang2025athena}, and multi-domain reward models have also been explored~\cite{zeng2025versaprm}. To our knowledge, \name is the first to systematically study MAD for multimodal reasoning with heterogeneous agents.

\noindent\textbf{Aggregation Approaches.} Most MAD systems aggregate solutions via majority voting~\cite{du2023improving} or weighted averaging using confidence estimates~\cite{chen2023reconcile}. From a crowdsourcing perspective, agents can be viewed as workers prone to systematic errors, motivating the use of quality-control methods for statistically sound consensus estimation~\cite{dawid1979maximum,hovy2013learning,ma2020adversarial,majdi2023crowdcertainlabelaggregationcrowdsourced,ustalov2021learning}. \name extends this line by adapting Dawid–Skene to account for both Solver and Reflector roles.

\section{Proposed Method -- \name}
We detail our MAD problem setup, our \name architecture, control flow, and solution aggregation from the debate below.

\subsection{Problem Setup}
Given a problem $\prob$ from a multimodal domain $\probset$, our goal is to design a model $\model: \probset\to\answerset$ that generates the correct answer $\ans\in\answerset$ to $\prob$. We mainly consider $\probset$ to be vision and language problems, where each $\prob$ is a pair of an image and a question text, answering the question needs both visuo-linguistc understanding. We use the notation $\hans=\model(\prob)$ to denote the predicted answer to $\prob$ by $\model$, and our goal is to get to $\ans=\hans$. When considering problems of multiple choice nature, we assume $\ans$ and $\hans$ are one of $K$ candidate options, while for free-form answers, we use an external model to evaluate the equality.

A standard machine learning way to approach our problem setup is to consider a dataset with pairs of problems and their ground truth solutions, where $\model$ is trained to minimize the prediction error. We assume neither training data nor in-context examples are available, instead have access to a set $\llmset$ of heterogeneous LLMs (either unimodal or multimodal) that have been pretrained on diverse corpora of language or multimodal tasks, not necessarily the tasks in $\probset$. Our goal is to use $\llmset$ to design $\model$ to solve $\probset$ in a zero-shot setting.

\begin{figure*}[t]
    \centering
    \includegraphics[width=14cm,trim={2cm 8.5cm 4cm 6cm},clip]{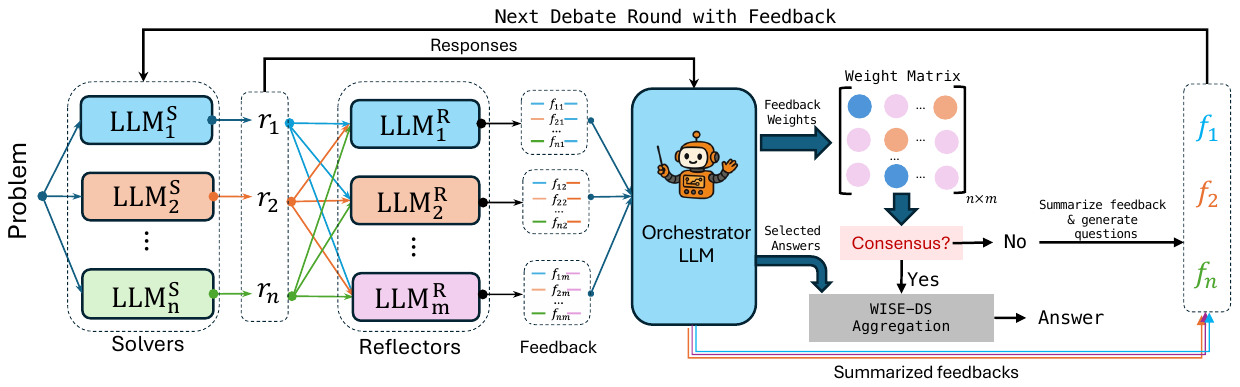}
    \vspace{-1em}
    \caption{\name architecture demonstrating the control flow, agent settings, feedback, responses, and the \name-Dawid-Skene solution aggregation scheme to produce the final response.}
    \label{fig:arch}
\end{figure*}

% \subsection{Heterogeneous Multi-Agent Debate}
As alluded to before, our key insight is that different LLMs / MLLMs may be trained on different types of data and could thus possess disparate abilities, \eg, Qwen-VL~\cite{wang2024qwen2} is trained on general purpose vision-and-language tasks while Phi-4~\cite{abdin2024phi} is trained for coding and math reasoning, and GPT-4.1\footnote{https://openai.com/index/gpt-4-1/} is trained with focus towards bridging vision with language with coding abilities. Since it is hard to judge what skill sets are needed to solve a given problem (and thus deciding which LLM configuration works best), it may be easier if LLMs of different capabilities debate on various aspects of the problem, towards reaching a consensus. With this intuition, we present our \fullname (\name) architecture, illustrated in Fig.~\ref{fig:arch}.

\subsection{\name Architecture}
Given a set of $|\llmset|$ unique agents, a standard MAD implementation~\cite{du2023improving,chen2023reconcile} needs agents to communicate with one another, resulting in nearly $|\llmset|^2$ messages. This quadratic growth can be redundant, costly when models are accessed through API calls, and computationally intensive when hosted locally. Moreover, the volume of messages may quickly exceed the input context size of the models. In the multimodal setting we consider, further restricting all agents to possess identical multimodal capabilities can be limiting; for example, some models may excel at language-based reasoning while others may have stronger vision–language alignment.

To address these challenges, \name partitions the agents into two groups: a set of $n$ Solvers $\solvers=\set{\llms_1,\llms_2,\dots,\llms_n}$ and a set of $m$ Reflectors $\reflectors=\set{\llmr_1,\llmr_2,\dots,\llmr_m}$, where $\solvers \cup \reflectors = \llmset$ and $\solvers \cap \reflectors$ may be non-empty. The Solvers are assumed to be multimodal agents, while the Reflectors may include both unimodal and multimodal models.

\noindent\textbf{Solver Models (First Round):} For a problem $\prob$, each solver agent $\llms_i \in \solvers$ receives as input the tuple $\tup^S_1=\tuple{\sprompt_1, \prob, \answerset_\prob, \fb=\phi}$, where $\sprompt_1$ is a textual prompt describing the task, and $\answerset_\prob$ denotes the candidate answer set (with $\answerset_\prob=\phi$ for free-form answers). The tuple also includes a feedback message $\fb$, which is null in the first round and populated with feedback from the reflectors and orchestrator in subsequent rounds. For vision–and–language problems, $\prob$ also contains the associated images. The exact prompts used are provided in the Appendix. Each solver $\llms_i$ generates a response $\response^i_k$ to $\prob$ at round $k$ ($k=1$ for the first round)\footnote{We will drop $k$ when the round index is not important.}. The response is expected to include both a detailed explanation of the solution steps and a final selected answer from the candidate set, \ie, $\hans \in \answerset_\prob$ (if a multiple choice problem). 

\noindent\textbf{Reflector Models:} At the $k$-th round, a reflector $\llmr_j \in \reflectors$ takes as input the tuple $\tup^R_k=\tuple{\rprompt, \prob, \response_k}$, where $\rprompt$ specifies the reflector’s task. The task is twofold: (i) judge the correctness of the solver’s response $\response_k$, and (ii) provide textual feedback. For correctness, the reflector outputs a numerical weight $\weight^k_{ij}\in\set{-1,0,1,2}$, with $\weight=-1$ indicating reflector failure, $0$ for an incorrect answer, $1$ for uncertainty/abstention, and $2$ for a correct response. For feedback, the reflector produces a detailed explanation justifying its assigned weight, highlighting correct aspects and pointing out errors in $\response_k$. Each solver’s response is thus evaluated by all reflectors, yielding: (i) an $n\times m$ feedback matrix $\fset^k$ whose $(i,j)$-th entry $\fb^k_{ij}$ stores the textual feedback, and (ii) an $n\times m$ weight matrix $\wmat^k$ whose $(i,j)$-th entry is $\weight^k_{ij}$.

\noindent\textbf{Orchestrator:}
Similar to~\cite{chen2023reconcile,wang2024rethinking,feng2024don}, a central component of \name is the orchestrator agent $\orch$, implemented as an LLM, which governs the debate process and message exchange among agents. However, in contrast, \name orchestrator has two extra responsibilities: (i) aggregating the feedback matrix $\fset^k$ and weight matrix $\wmat^k$ to decide whether the debate should continue, and (ii) summarizing the collected feedback into actionable guidance for the solvers.

If the debate is to continue, the orchestrator produces consolidated feedback for each solver. Specifically, for solver $i$, it generates a feedback summary $\fb^k_i$ as:
\begin{equation} \fb^k_i = \llmo\left(\oprompt, \!\!\!\!\!\!\!\!\concat_{j\in[n],\weight_{ij}\neq -1}\!\!\!\!\!\!\!\!\fset^k_{ij}\right), i\in[n], \end{equation}
where $\llmo$ denotes the orchestrator’s internal LLM, and the input consists of all non-null feedback from reflectors on solver $i$’s response $\response^i$, concatenated with the orchestrator prompt $\oprompt$. The prompt $\oprompt$ instructs $\llmo$ to summarize the feedback and formulate a set of \emph{challenge questions} that will force the solver to (re-)consider aspects of the image or language part of the problem in more detail, toward refining its solution in the next round.

The orchestrator then prepares a solver-specific feedback tuple $\tuple{\sprompt_k, \prob, \answerset_\prob, \response^i \oplus \fb^k_i}$, where $\sprompt_k$ is the updated solver prompt and $\oplus$ denotes concatenation of the solver’s response with the orchestrator feedback. This tuple is forwarded to the $i$-th solver, and the process is repeated for all $i \in [n]$, thereby completing one debate round. The debate continues until either a maximum number of rounds is reached or consensus is achieved, as described earlier.

\subsection{\name Dawid-Skene Response Aggregation}
Given $N$ worker responses, each selecting one of $K$ possible answers for a given problem, the classic Dawid–Skene (DS) algorithm employs expectation–maximization (EM) to jointly estimate the error rates of workers and the latent ground-truth label distribution. The DS method and its variants are widely used in the crowdsourcing literature to infer consensus from noisy annotations~\cite{majdi2023crowdcertainlabelaggregationcrowdsourced}. In our context, assuming conditional independence of agents’ responses, we cast our MAD formulation within the DS framework to model agent-specific error rates and compute the posterior over solution responses. This down-weights the agents behaving randomly and amplifies the influence of those whose outputs are consistently aligned with the majority.

Suppose $\prob_{\alpha\beta}$ is the probability of selecting answer $\beta$ for the true answer $\alpha$ by a model, and if $\zeta_\alpha$ is the true prior probability of selecting an option $\alpha$, then given $\Lambda$ problems and the agents' responses, \cite{dawid1979maximum} models the likelihood of the data as a mixture of multinomial distributions:
\begin{equation}
\prob(\text{data}) \propto \prod_{i=1}^\Lambda\sum_{\alpha=1}^K\zeta_\alpha\left[\prod_{j=1}^N\prod_{\beta=1}^K (\prob^j_{\alpha\beta})^{\lambda^j_{i\beta}}\right],
\end{equation}
where $\lambda^j_{i\beta}$ is the number of times $j$-th agent produced $\beta$ as the answer against the true answer of $\alpha$ if the model is run multiple times on the same problem $i$. As both the true priors $\zeta_\alpha$ and the error rates $\prob_{\alpha\beta}$ are unknown, DS uses EM to optimize the likelihood iteratively towards convergence. 

In \name, we have two sources of errors: i) solvers making errors in selecting the correct answers from the $K$ choices and the reflectors selecting one of $J (=3)$ weights. Suppose $\prob^{t}_{\alpha_1\beta_1}$ and $\prob^{c}_{\alpha_2\beta_2}$ denote the error matrices for the solver and the reflectors respectively, in selecting option $\beta_1\to\alpha_1$ (for $\alpha_1,\beta_1\in[K]$) and selecting wrong weights for a provided answer, \ie, selecting weight $\beta_2\to\alpha_2$ (for $\alpha_2,\beta_2\in[J]$), then the joint probability for estimating the combined error matrices and their joint true priors $\zeta_{\alpha\beta}=\zeta_\alpha\zeta_\beta$ could be modeled for debate round $k$ as a product of two mixtures of multinomial distributions, given by: 
\fontsize{9pt}{9pt}
\begin{align}
&\prob^k(\text{data}) \propto \nonumber \\
& \prod_{i=1}^\Lambda \sum_{\alpha=1}^J\sum_{\beta=1}^K\zeta_{\alpha\beta}\left[\prod_{c=1}^{|\reflectors|}\prod_{\beta_2=1}^J \left(\prob^c_{\alpha_2\beta_2}\right)^{\lambda^{R_c}_{i\beta_2}} \prod_{t=1}^{|\solvers|}\prod_{\beta_1=1}^K\left(\prob^t_{\alpha_1\beta_1}\right)^{\lambda^{S_t}_{i\beta_1}}\right],\notag
\end{align}
\normalsize
subject to $\sum_\alpha \zeta_\alpha=\sum_\beta \zeta_\beta = 1$.
We use EM to estimate the joint error matrices and true probabilities, following which the posteriors are computed, which is then used to recompute the selected answers and the respective weights. See the Appendix for detailed derivations.  

If $\wwmat$ is the updated weight matrix given by the \name-DS method above after the EM convergence, then we aggregate these weights across rounds towards finding the final answer. Specifically,
given updated weight matrices from $k$ rounds, we generate aggregated weight $\awmat$ for an answer $\hans$ given by:
\begin{equation}
\awmat^k_{ij}(\hans) = \sum_{\ell=1}^k{\wwmat^\ell_{ij} \frac{\ell}{k(k+1)}},~~\text{if } \response_i^\ell = \hans.
\label{eq:aweight}
\end{equation}
The aggregated weighting scheme in~\eqref{eq:aweight} favors higher weights for more recent rounds while also normalizing them to [0,1] (recall that the weights in $\wwmat$ are in $\set{0,1,2}$).   We also maintain a distribution of weights across the answer candidates. Specifically, assume $|\answerset_\prob|$ updated answer options (after the DS step) for a given problem $\prob$, then the accumulated weight $\aw_{\hans}$ for option $\hans\in\answerset_\prob$ is given by:
\begin{equation}
    \aw^k_{\hans} = \sum_{i\in[n],j\in[m]} \awmat_{ij}^k(\hans),
\end{equation}
and we select the highest ranking answer $\ans^*$ as the output from the final $\aw$, where $\ans^* = \argmax_{\hans} \set{\aw_{\hans}}.$

\section{Experiments}
We conduct experiments to validate \name for robust multi-agent debate. We first review the datasets we used, describe the agents we used, and then present numerical results. 

\noindent\textbf{Datasets.} We evaluate on three existing datasets: (i) SMART-840~\cite{cherian2024evaluating}, (ii) Visual Puzzles~\cite{song2025visualpuzzles}, and (iii) EvoChart-QA~\cite{huang2025evochart}, all of which involve vision–and–language mathematical reasoning. SMART-840 contains 840 problems from children’s mathematical Olympiads (grades 1–12), grouped into six categories spanning consecutive grade pairs, enabling analysis of MLLM performance on age-appropriate problems. Visual Puzzles consists of 1168 problems designed to test algorithmic, analogical, deductive, inductive, and spatial multimodal reasoning. Both datasets use multiple-choice answers, with 5 options in SMART-840 and 4 in Visual Puzzles. EvoChart-QA is a subset of the ChartQA benchmark comprising vision–and–language questions over charts (line, bar, scatter, and pie plots). We use its “complex” subset of 320 problems, considered the most challenging and where SOTA MLLMs perform poorly. Unlike the other two datasets, EvoChart-QA requires free-form short answers, allowing us to evaluate \name in non–multiple-choice settings.

\begin{figure*}[ht]
    \centering
    \includegraphics[width=14cm, trim={0.5cm 11.5cm 1cm 3cm}, clip]{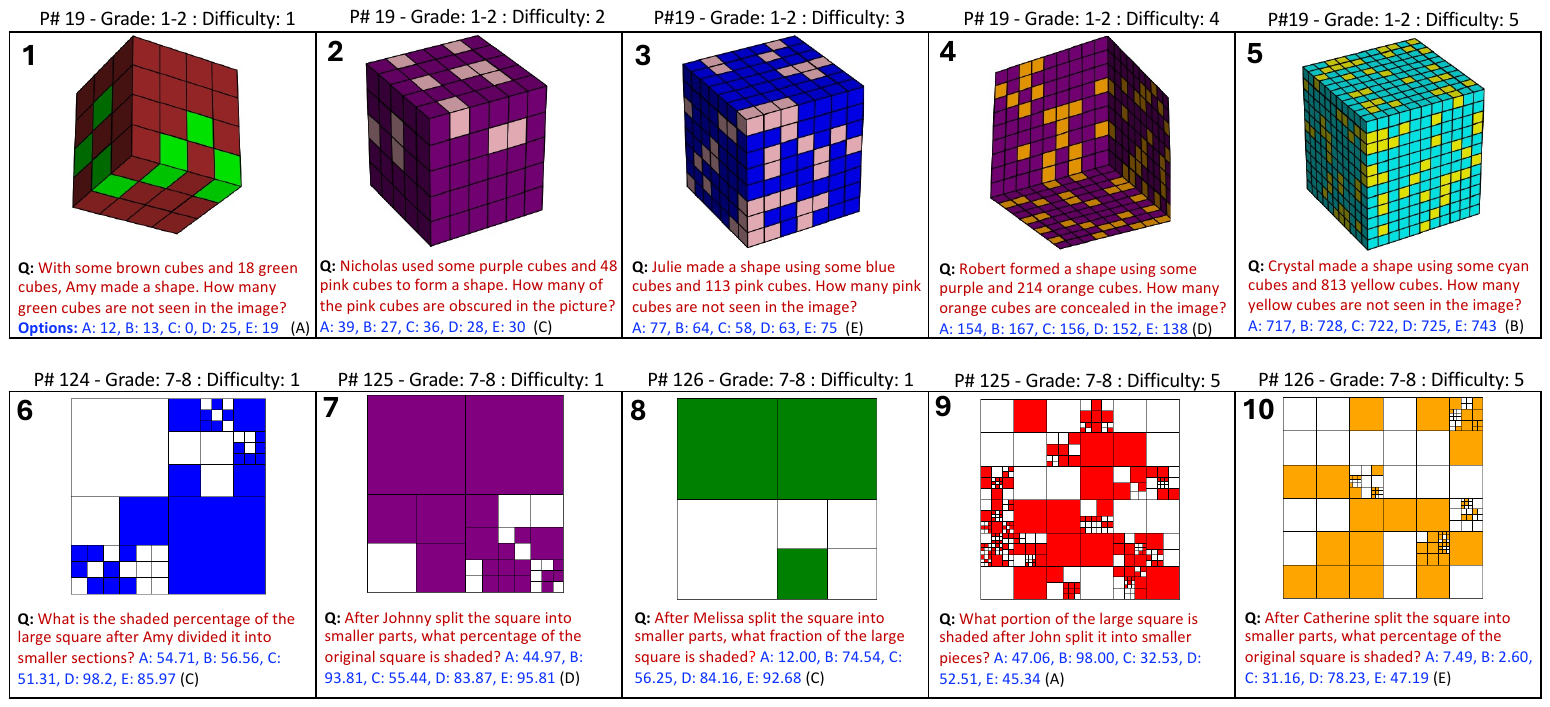}
    \includegraphics[width=14cm, trim={0.5cm 10cm 1cm 4cm}, clip]{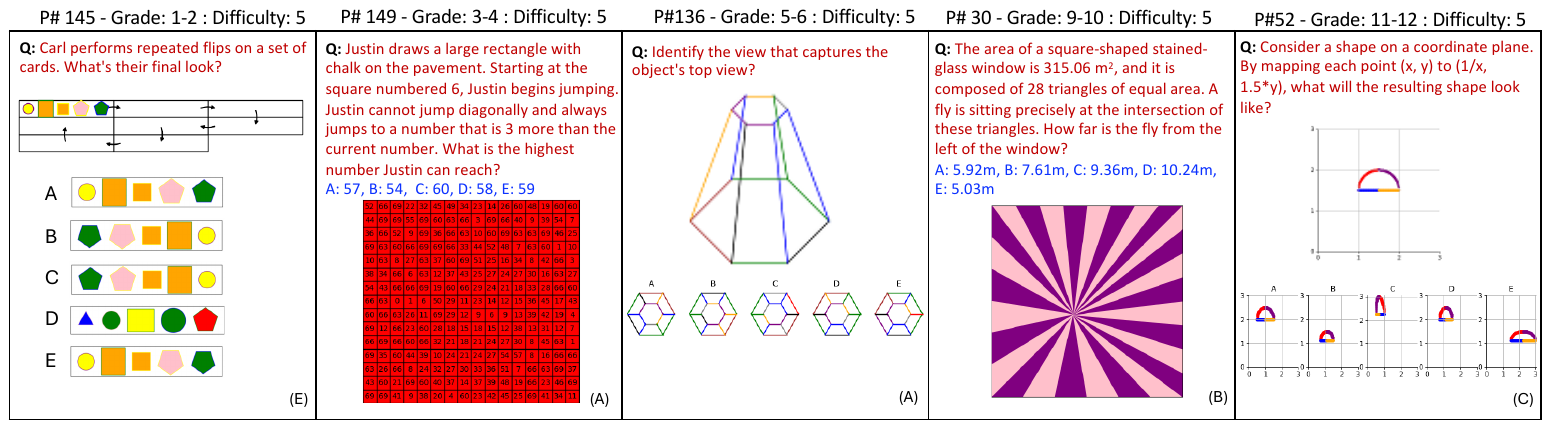}
    \vspace{-1em}
    \caption{Example problems from our proposed SMART-840++ dataset. Cells 1-5 show the same puzzle but with increasing difficulties. Cells 6-8 show puzzle instances (of grade 1-2) of the same difficulty level (1) but with different configurations. Cells 9-10 show diversity over instances for a puzzle from grade 7-8. Last row shows an ensemble of puzzles across grades of the highest difficulty level.}
    \label{fig:smart840++}
\end{figure*}

\noindent\textbf{SMART-840++ Dataset.} In addition to the three datasets above, we introduce SMART-840++, a novel extension of SMART-840 that programmatically augments 55 selected vision–language puzzles for 5 difficulty levels, each level with three new instances each, resulting in a total of 825 problems. The dataset has two primary goals: (i) to assess the performance of the SOTA MLLMs as problem complexity increases, and (ii) to evaluate agents on unseen problem instances, thereby reducing the risk of data contamination. To construct SMART-840++, we developed Python scripts that recreate each puzzle and generate novel variants by modifying image attributes and problem settings while preserving the underlying solution algorithm. Our approach builds on~\cite{cherian2023deep}, but targets the more advanced reasoning challenges in SMART-840 rather than elementary-level tasks. Fig.~\ref{fig:smart840++} shows some example puzzle instances of SMART-840++. We provide more details and examples in the Appendix.

\noindent\textbf{Agents and Debate Notation.}
We use diverse LLM/MLLM agents in our experiments, listed below with their abbreviations: GPT-4o (G4o), GPT-4.1 (G41), o1-mini (o1m) o4-mini (o4m), Claude-Sonnet-3.5 (CS35), Claude-Sonnet-3.7 (CS37), Qwen2-VL-2.5-7B (QVL), Gemma3-8B (Ge3), and Phi4-3.4B ($\phi4$). To represent a debate, we use the notation $(\llms_1+ \llms_2+\cdots)\times(\llmr_1+\llmr_2+\cdots)|\text{orchestrator}$, where the first set is the solver models and the second set is the reflectors. For example, (G41+CS37)$\times$(o4m+$\phi4$)|G4o means we use GPT-4.1 and Claude-Sonnet-3.7 as the solvers, (o4-mini, Phi-4) as reflectors, and GPT-4o as the orchestrator. We use the compact notation $(\llm_1+\llm_2+\cdots)^2|O$ for a configuration that uses $\llm_1+\llm_2+\cdots$ as both solvers and reflectors.

\noindent\textbf{Debate Configuration:} We ran up to 8 debate rounds, with early stopping when the orchestrator terminated the debate upon reaching consensus. Our analysis considers debate setups inspired by prior work, using models closest to those previously studied. Wherever possible, we report numbers from the original papers and supplement them with our own reproduced results. For solution aggregation, we compare against methods implemented in the Crowd-Kit toolbox.\footnote{https://github.com/Toloka/crowd-kit}

\noindent\textbf{Evaluation and Implementation Details:} We use multiple-choice accuracy on SMART-840, SMART-840++, and VisualPuzzles, and use the Phi-4 model to compare predicted answers with ground truth on EvoChart-QA. \name is implemented in PyTorch with MLLMs sourced from HuggingFace or accessed via API calls. Our experiments were conducted on an A100 node, assigning each MLLM to a dedicated GPU. The distributed setup employed asynchronous communication for message passing among MLLMs. We accessed GPT and Claude models via the OpenAI and Claude-3 APIs. Across all debate steps and models, we used a consistent set of prompts, formatted according to each model’s template.

\begin{table*}
    \caption{Performance of varied MAD approaches on the SMART-840 dataset using varied LLM/MLLM configurations. We report performance on every pair of problem grades from 1--12, on VL and L-only subsets, and the average over all grades. }
    \vspace{-.5em}
    \centering
    \resizebox{0.9\textwidth}{!}{%
    \begin{tabular}{c|c|c|c|c|c|c|c|c|c|c}
        Method & LLMs / MLLMs & 1\_2 & 3\_4 & 5\_6 & 7\_8 & 9\_10 & 11\_12 & Image+Text & Text & Overall  \\
        \hline
        \multirow{3}{*}{\parbox[c][0.1cm][c]{3cm}{\centering Self-Reflect\\\cite{shinn2023reflexion}}} & GPT-4.1 & 54.2 & 55.8 & 51.3 & 70.0 & 66.7 & 68.7 & 50.9 & 88.9 & 61.2\\
                & Gemma3 & 38.8 & 36.3 & 36.2 & 46.1 & 53.5 & 64.9 & 30.0 & 58.2 & 46.0\\
                & Qwen-VL-2.5 & 33.3 & 28.6 & 23.3 & 45.6 & 35.6 & 50.7 & 28.5 & 55.4 & 36.2\\
        \hline
        \multirow{4}{*}{\parbox[c][1cm][c]{2.5cm}{\centering Self-Consistency\\\cite{wang2022self}}} & GPT-4.1 & 60.0 & 50.8 & 51.3 & \textbf{76.0} & 66.7 & 73.3 & 48.6 & 90.3 & 63.2\\
                 & Sonnet-3.5 & 49.2 & 37.3 & 35.4 & 48.0 & 46.3 & 48.7 & 35.3 & 68.4 & 44.2\\
                 & Gemma3 & 39.8 & 29.8 & 36.1 & 50.8 & 49.2 & 57.4 & 24.8 & 56.8 & 43.9\\
                 & Qwen-VL-2.5 & 33.6 & 30.8 & 33.1 & 41.1 & 47.2 & 41.1 & 28.9 & 59.2 & 37.8\\
        \hline
        \multirow{4}{*}{\parbox[c][1cm][c]{3cm}{\centering Homogeneous\\\cite{liang2023encouraging}}} & GPT-4.1 & 53.3 & 50.8 & 50.0 & 70.0 & 68.7 & 65.3 & 49.5 & 88.4 & 59.7\\
                 & Sonnet-3.5 & 49.1 & 35.3 & 38.4 & 53.0 & 46.9 & 48.7 & 34.9 & 71.0 & 45.2\\
                 & Gemma3 & 39.7 & 38.7 & 35.3 & 45.2 & 50.0 & 57.7 & 30.9 & 74.2 & 44.4\\
                 & Qwen-VL-2.5 & 42.2 & 33.6 & 27.5 & 45.7 & 39.6 & 44.0 & 33.3 & 60.6 & 38.8\\
        \hline
        \multirow{3}{*}{\parbox{2.5cm}{\centering RECONCILE\\\cite{chen2023reconcile}}} & G41 + Ge3 + S35 & 55.0 & 59.2 & 50.0 & 67.3 & 68.7 & 72.0 & 50.9 & 90.1 & 62.0\\
                 & G4o + Ge3 + QVL  & 50.8 & 41.7 & 36.7 & 54.7 & 56.0 & 62.7 & 39.6 & 76.1 & 50.4\\
                 & Ge3 + QVL	 & 41.7 & 38.7 & 34.0 & 50.7 & 50.0 & 54.7 & 35.8 & 65.5 & 45.0\\
        \hline	
        \multirow{3}{*}{WISE (ours)} & G4o $\times$ G4o|G4o & 45.0 & 45.0 & 38.0 & 54.0 & 52.0 & 52.0 & 38.4 & 68.1 & 47.7\\
                & G4o $\times$ (o1m+$\phi4$)|G4o & 49.2 & 45.4 & 47.1 & 54.6 & 56.9 & 55.8 & 39.7 & 78.6 & 51.5\\
                % & (G4o+Q)$\times$(Q+$\phi4$+L3+G3)|G4o & 0.514 & 0.394 & 0.41 & 0.537 & 0.539 & 0.587 & X8 & X9 & 0.497\\
                % & (G41+S35)$\times$(G4o+S35+$\phi4$)|G4o & 0.592 & 0.583 & 0.493 & 0.733 & 0.607 & 0.693 & X8 & X9 & 0.617\\
                & (G41+Ge3+S35)$^2$|G4o & \textbf{65.0} & \textbf{61.8} & \textbf{56.6} & 74.3 & \textbf{75.5} & \textbf{73.0} & \textbf{55.8} & \textbf{91.4} & \textbf{68.1}\\
    \end{tabular}
    }
    \label{tab:smart-840}

\vspace{0.25cm}
\caption{Comparisons on the Visual Puzzles Dataset. $^*$indicates the overall performance reported in the original paper, unless it is our method. $^\dagger$ indicates results obtained when we ran the model. }
\vspace{-.5em}
\centering
    \resizebox{0.9\textwidth}{!}{%
    \begin{tabular}{c|c|c|c|c|c|c|c|c}
    Method & LLMs/MLLMs & Algorithmic & Analogical & Deductive & Inductive & Spatial & Overall$^*$ & Our Run$^\dagger$\\
    \hline
    Random Choice & NA & 25.0 & 25.0 & 25.0 & 25.0 & 25.0 & 25.0 & NA\\
    Human (50th \%-ile) & NA & 88.0 & 66.0 & 80.0 & 50.0 & 90.0 & 75.0 & NA\\
    \hline
    \multirow{3}{*}{Single} & GPT-4o & 49.2 & 58.3 & 49.0 &  27.3 & 26.2 & 41.3 & 42.1\\
                 & o4-mini & 65.3 & 68.7 & \textbf{75.5} & 33.0 & 45.5 & 57.0 & 53.3\\
                 & CS37 & 64.5 & 48.3 & 65.0 & 26.8 & 37.4 & 48.3 & 46.6\\
                 & Gemma3 & 40.4 & 36.4 & 34.7 & 27.5 & 15.3 & NA & 31.6 \\
                 % & Qwen-VL-2.5 & 53.4 & 46.9 & 58.0 & 25.8 & 29.5 & 42.3 & NA\\
    \hline 
    WISE (ours) & (o4m+CS37+Ge3)$^2$|G4o & \textbf{70.9} & \textbf{69.0} & 75.1 & \textbf{37.0} & \textbf{47.1} & \textbf{59.8} & \textbf{59.8}  \\
    \end{tabular}
    }
    \label{tab:visual_puzzles}
\end{table*}
\begin{table}
\caption{Comparisons on SMART-840++ over difficulty levels.}
\vspace{-.5em}
\centering
    \resizebox{0.48\textwidth}{!}{%
    \begin{tabular}{c|c|c|c|c|c|c|c}
    Method & LLMs/MLLMs & Level-1 & Level-2 & Level-3 & Level-4 & Level-5 & Overall\\
    \hline
    Single & GPT-4.1 & 16.0 & 19.2 & 19.7 & 18.1 & 15.4  & 17.7 \\
    Single & Sonnet-3.7 & 17.4 & 19.5 & 16.3 & 19.4 & 17.8 & 18.1\\
    \hline 
    \name & (G41)$^2$|G4o & 15.3 & 18.4 & 20.1 & 17.4 & 19.6 & 18.2\\
    \name & (CS37)$^2$|G4o & 16.1 &	17.5 &	19.4 &	19.4 & 15.3 & 17.5\\
    \name & (G41+CS37)$^2$|G4o & 26.5 & \textbf{31.4} & 31.7 & \textbf{21.7} & \textbf{23.0} & \textbf{26.9}\\
    \name & (Ge3+QVL)$^2$|G4o	& 16.1 & 19.6 & 21.2 & 19.6 & 17.4 & 18.8\\
    \name & \makecell{(o4m+CS37)$ \times $\\(G41+CS37+o1m)|G4o} & \textbf{28.8} & 21.9 & \textbf{35.5} & 17.5 & 15.3 & 23.8\\
    \end{tabular}
    }
    \label{tab:smart840++}
    \vspace*{-0.5cm}
\end{table}

\begin{table*}[ht]
\caption{Comparisons of \name-DS against methods from \cite{majdi2023crowdcertainlabelaggregationcrowdsourced}. DS: Dawid-Skene, Maj. V: majority voting, Wt. Maj.V uses Eq.~\ref{eq:aweight}.}
%\SL{Can't see the full table, can we move it to the next page?}
\vspace{-.5em}
\centering
    \resizebox{.9\textwidth}{!}{%
    \begin{tabular}{c|c|c|c|c|c|c|c|c|c}
    Dataset & LLMs/MLLMs & Maj. V & DS & MACE & Wawa & Zero-Skill & MMSR & Wt. Maj.V & WISE\\
    \hline
    \multirow{3}{*}{SMART-840} & (G41+CS35+Ge3)$^2$|G4o	 &  63.9 & 65.3 & 67.0 & 65.3 & 66.4 & 65.1 & 66.2 & \textbf{68.1}\\ 
    & (G41+S35)$\times$(G4o+S35+$\phi4$)|G4o	 & 61.1 & 62.6 & 61.7 & \textbf{62.6} & 62.3 & 62.6 & 59.1 & 62.3\\
    &(G4o+Q)$\times$(Q+$\phi$4+Ll3+Ge3)|$\phi$4 & 49.1 & 49.2 & 49.7  & 50.3 & 50.4 & 49.6 & 50.0 & \textbf{51.1}\\
    &G4o$\times$G4o|$\phi$4 &  46.8 & 46.9 & 47.1 & 47.5 & 47.5 & 46.5 & 47.4 & \textbf{48.2}\\
    \hline
    Visual Puzzles & (o4m+Ge3+C37)$^2$|G4o & 57.3 & 56.1 & 56.9 & 57.2 & 57.2 & 56.6 & 58.1 & \textbf{59.8}\\
    \hline
    EvoChart-QA & (o4m+C37)$^2$|Go & 74.4 & NA & NA & NA & NA & NA & \textbf{75.4} & NA\\
    \hline
    \multirow{2}{*}{SMART-840++} & (G41+CS37)$^2$|G4o & 24.8 & 25.1 & 25.4 & 24.8 & 25.2 & 25.0 & 26.0 & \textbf{26.9}\\
                & (o4m+CS37)$\times$(G41+CS37+o1m)|G4o & 21.6 & 22.5 & 22.7 & 22.7 & 22.5 & 22.7 & 22.1 & \textbf{23.8}\\
    \end{tabular}
    }
    \label{tab:agg}
\end{table*}

\subsection{Experimental Results}
\begin{figure*}[h]
    \centering
    \begin{subfigure}{5cm}
        \centering
        \includegraphics[width=5cm]{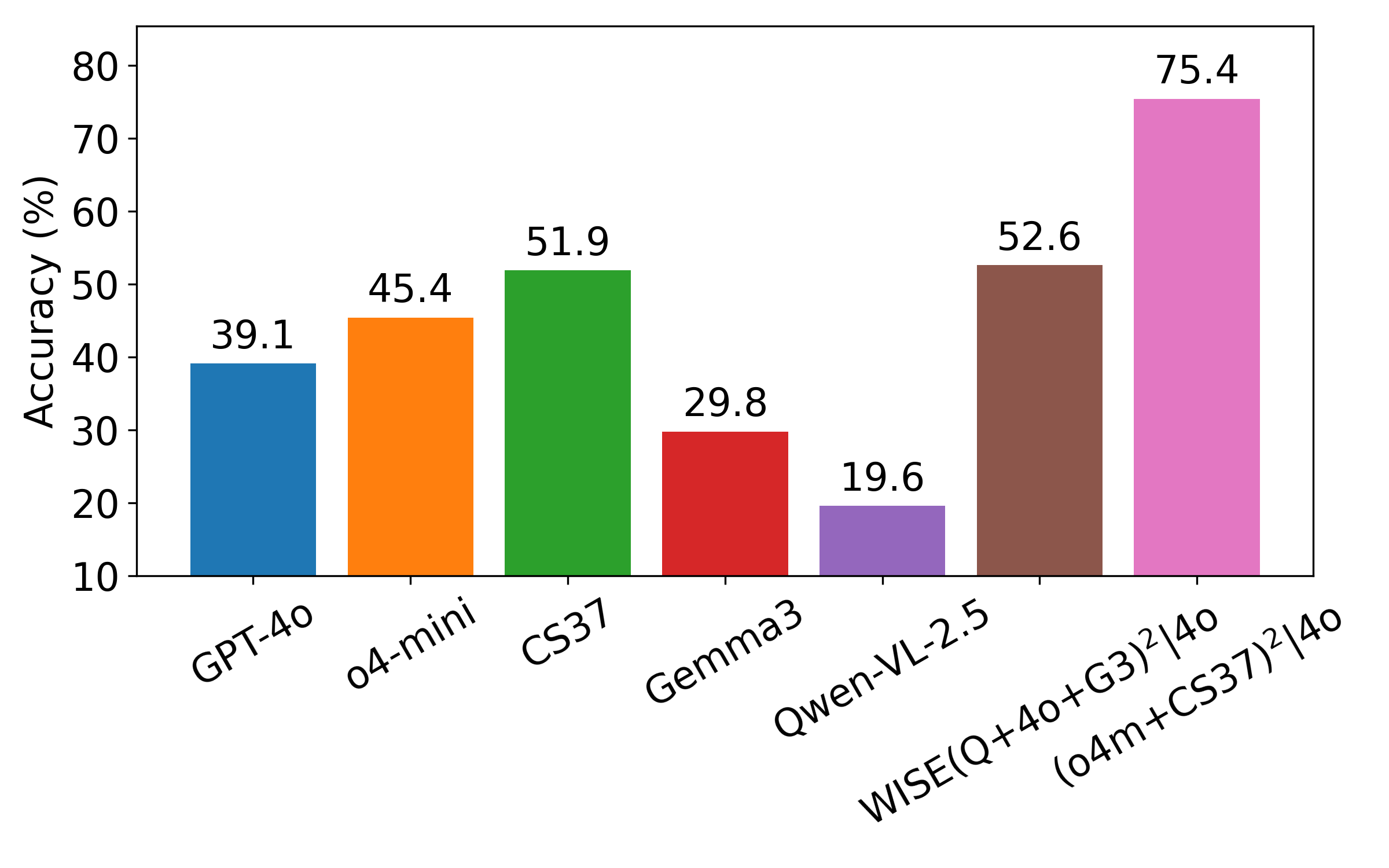} %\hspace*{0.4cm}
        \subcaption{Accuracy on EvoChart-QA}
        \label{fig:evochart1}
    \end{subfigure}
    % [Accuracy on EvoChart-QA]
    \begin{subfigure}{5cm}
        \centering
        \vspace*{-0.5cm}
        \includegraphics[width=5cm,trim={0 0 0 0.85cm},clip]{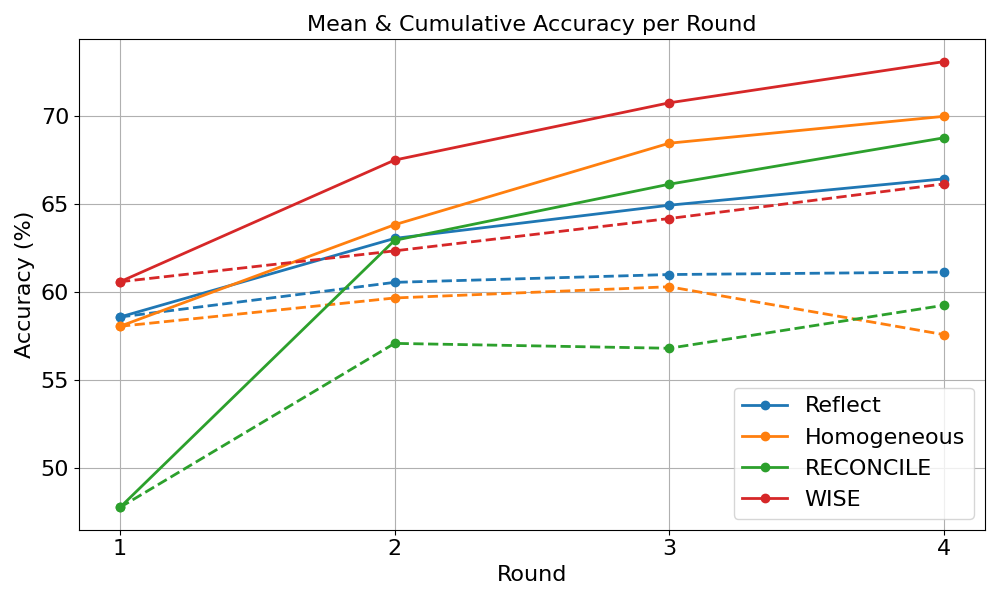}
        \subcaption{Cumulative \& Average (dashed) Accuracy}
        \label{fig:cum_acc}
    \end{subfigure}
    \begin{subfigure}{7cm}
        \centering
        \includegraphics[width=7cm,trim={0 0 0 0.85cm},clip]{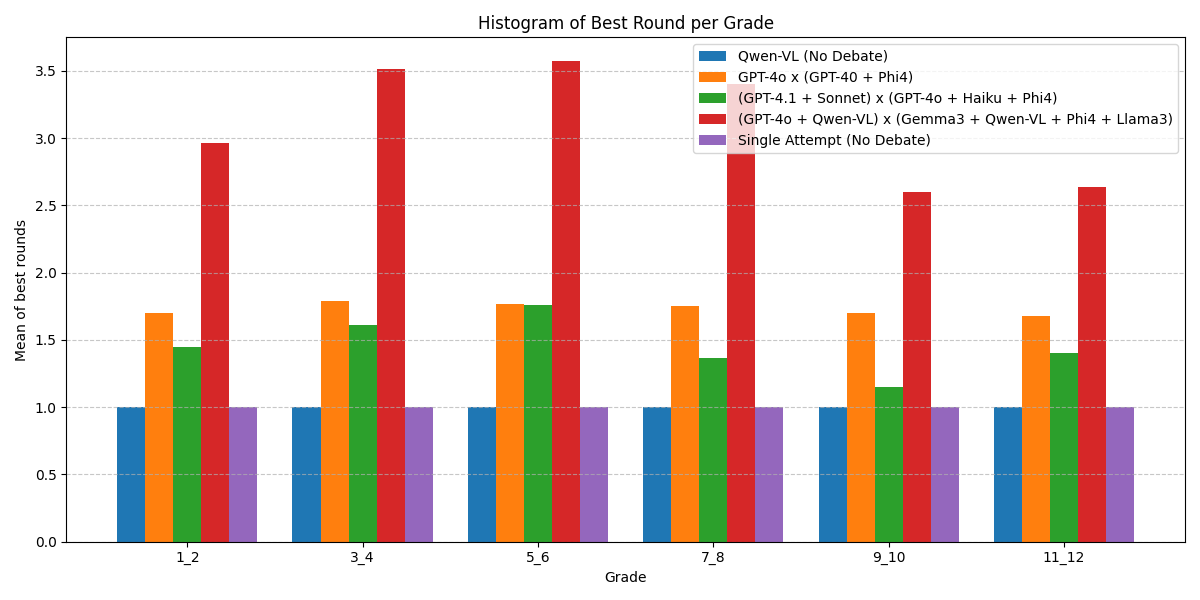}
        \subcaption{Average rounds for debate consensus}
        \label{fig:best_rounds}
    \end{subfigure}
    \caption{(a) shows performance on the EvoChart-QA dataset. (b) plots the cumulative and average accuracy for varied configurations against the number of debate rounds, and (c) plots the average number of rounds by \name.}
    \label{fig:accuracy_versus_rounds1}
    \vspace*{-0.2cm}
\end{figure*}

\begin{figure*}[ht]
\centering
\includegraphics[width=14cm, trim={0.8cm 11.3cm 1.6cm 8.5cm},clip]{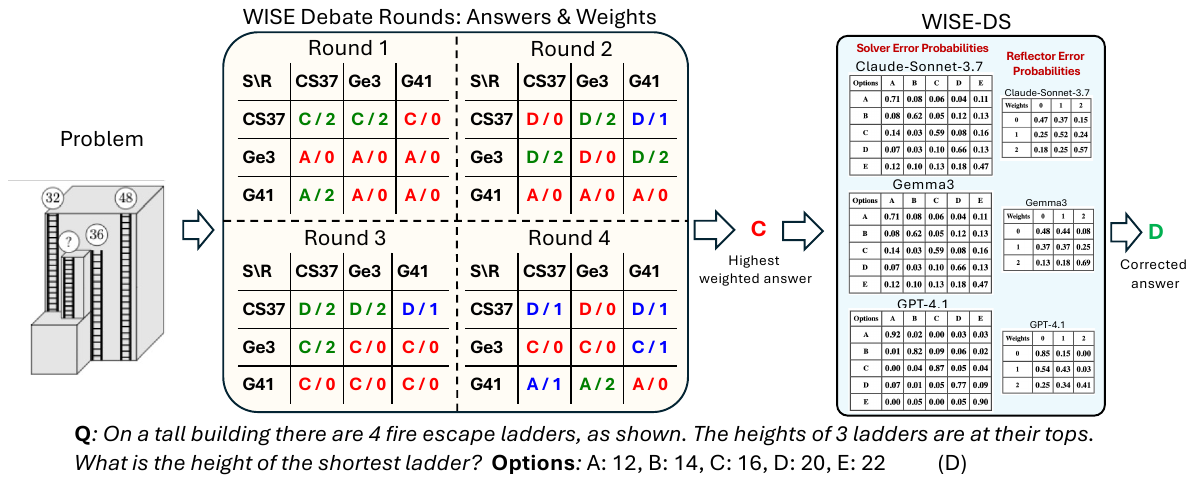}
\caption{A SMART-840 example and its summarized WISE debate. We show a four-round WISE debate using Claude-Sonnet-3.7, GPT-4.1, and Gemma-3. The solver predictions and reflector-assigned weights are shown (e.g., \textcolor{red}{\textbf{C/0}}). Although the correct answer D does not appear in Round 1, it emerges in later rounds. While the debate ranks C highest, the WISE-DS aggregation step—using each agent’s error-probability matrices—correctly recovers D via posterior inference. Additional examples and full debate transcripts are provided in the Appendix.}
\label{fig:qual_result}
\end{figure*}

\noindent\textbf{SOTA Comparisons:} In Tables~\ref{tab:smart-840}--\ref{tab:smart840++} and Fig.~\ref{fig:accuracy_versus_rounds1}, we report results across the four datasets, comparing \name against the SOTA methods. Overall, \name consistently outperforms all baselines. On SMART-840, it improves accuracy by nearly 6\% over RECONCILE~\cite{chen2023reconcile}, while also surpassing single-model, homogeneous, and self-correction variants. On VisualPuzzles, \name achieves a 2.8\% gain with an ensemble of o4-mini, Claude-Sonnet-3.7, and Gemma3, demonstrating that benefits persist even with strong models such as o4-mini. On EvoChart-QA, \name substantially outperforms individual models—improving GPT-4o (39.1\%), Gemma3 (29\%), and Qwen-VL (19.6\%) to 52.6\%. Using stronger models like Claude-Sonnet-3.7 and o4-mini further boosts performance to 75.4\%, nearly 20\% above previously reported results. These findings also highlight \name’s effectiveness on free-form short-answer problems. On SMART-840++, results are notably lower than on the other datasets, underscoring the challenges of visual reasoning for MLLMs. Nevertheless, \name consistently outperforms alternative configurations, though performance declines at higher difficulty levels. A qualitative result is provided in Figure~\ref{fig:qual_result} depicting the weights over rounds and  the estimated error matrices.

Table~\ref{tab:agg} compares our \name-DS solution aggregation method with alternatives from the Crowd-Kit toolbox~\cite{majdi2023crowdcertainlabelaggregationcrowdsourced}. Our DS extension yields consistent gains across debate configurations and datasets. For competing methods, we applied their aggregation scheme followed by weighted majority without the joint EM step. Overall, our approach surpasses the original Dawid–Skene model, while weighted majority voting further shows steady and promising improvements.

\subsection{Ablation Studies}
\noindent\textbf{Does MAD help multimodal reasoning?}
In Table~\ref{tab:smart-840}, we evaluate on SMART-840 by adapting several SOTA methods originally designed for language-only tasks to VL problems. We consider four setups: (i) single-model baselines (GPT-4.1, Claude-Sonnet-3.5, Gemma3, and Qwen2-VL), (ii) self-reflection~\cite{shinn2023reflexion} and self-consistency~\cite{wang2022self}, which refine responses via feedback or repeated sampling, (iii) homogeneous debate method~\cite{liang2023encouraging}, where an agent serves as both solver and critic, and (iv) RECONCILE~\cite{chen2023reconcile}, where three agents iteratively debate to improve confidence. All experiments use the same set of MLLMs.

Table~\ref{tab:smart-840} shows that debate-based methods consistently outperform single-model baselines. For instance, RECONCILE improves Gemma3+Qwen2-VL by $\sim$7\%, though gains are smaller for strong models (\eg, GPT-4.1 improves by only 1.5\%). Overall, debate proves more effective than other self-correction methods. Notably, \name yields further gains, surpassing RECONCILE by 6.1\% under the same setup. Separating results by modality confirms that MAD methods (\eg, RECONCILE, \name) are especially effective for multimodal reasoning, with \name improving performance by nearly 5\% over the next best method on VL problems while also providing modest gains on text-only tasks.

\noindent\textbf{Are non-vision LLMs useful?}
Table~\ref{tab:smart-840} compares two settings: (i) GPT-4o as both solver and reflector under \name, and (ii) GPT-4o as solver with o1-mini and Phi-4 as reflectors. Since both o1-mini and Phi-4 are language-only models and the GPT-4o orchestrator does not access images, this setup tests the utility of non-vision LLMs. We find they are indeed beneficial: GPT-4o’s accuracy rises from 47.7\% to 51.5\% when paired with language-only reflectors.

\noindent\textbf{How does accuracy vary across rounds?}
Fig.~\ref{fig:accuracy_versus_rounds1} plots cumulative accuracy (fraction of problems solved in round $k$) and average accuracy/round for multiple configurations. \name performs strongly in both metrics. The close alignment of cumulative and average accuracy shows that our aggregation scheme is robust. In particular, the gap between cumulative accuracy (an upper bound) and average accuracy is much smaller for \name-DS compared to other schemes.  

\noindent\textbf{Computational Complexity?}
While the design of \name is flexible, modular, and achieves state-of-the-art accuracy on challenging benchmarks, this capability comes with the cost of increased communication between the solvers, reflectors, and the orchestrator models. Specifically, for $N$ solvers, $M$ reflectors, and an orchestrator engaged in a $K$ round \name debate, the worst-case number of LLM calls is $K(M+N+MN)$. Although this may seem substantial, in practice the average number of calls is much lower. Figure \ref{fig:best_rounds} shows the average number of rounds across different \name configurations using both weak and strong LLMs (with a maximum of eight rounds). The results reveal that debates involving weaker LLMs typically require around 2.5 rounds, whereas stronger models (e.g., GPT and Claude) converge in fewer than 1.5 rounds on average—consistently across varying levels of problem difficulty.

\section{Conclusions}
We proposed \fullname (\name), a novel formulation of multi-agent debate (MAD) involving heterogeneous LLMs and MLLMs with complementary capabilities. \name supports multi-round multimodal debates by partitioning models into \emph{Solvers} and \emph{Reflectors}, coordinated by an orchestrator that manages distributed message passing while respecting LLM token limits. Our results highlight several key findings: (i) MAD is effective in multimodal settings, (ii) segregating agents into roles, as in \name, enables a more effective division of abilities, and (iii) incorporating error correction through cross-model solution consistency further improves performance. Please see the Appendix for more details, experiments, and results.

{
    \small
    \bibliographystyle{./paper_styles/ieeenat_fullname}
    \bibliography{mad_bib}
}

\appendix
\section{Additional Ablation Studies}
In this section, we present additional ablation studies that evaluate several key components of our framework. Before diving into the experiments, we first review the LLM/MLLMs we use, their abbreviations that we refer to them as and establish a baseline performance metric that we use to categorize the MLLM and LLM models in our subsequent analyses. In Table~\ref{tab:llms}, we provide publicly available information of the LLMs that we consider, their capabilities, and tasks they were trained for. Table~\ref{tab:smart-840-weak-strong-suppl-llms} summarizes each model’s accuracy, parameter size, modality capabilities, and baseline performance on the SMART-840 dataset. These scores are obtained using a single forward pass—without any debate rounds—thereby serving as the foundational performance reference. We report results for both image–text problems and text-only problems, as our later studies incorporate text-only LLMs as reflectors that help weigh and critique the solutions proposed by multimodal LLMs.

Based on these baseline results, we group the models into strong and weak categories. The two best-performing models, GPT-4.1 and Claude-Sonnet, are designated as strong. The remaining multimodal models—GPT-4o, Gemma3, and Qwen-VL-2.5—are categorized as weak. Similarly, the two text-only models, Phi-4 and Llama3, are also considered weak for the purposes of this study. We emphasize that this classification is purely performance-driven and is intended to enable a systematic exploration of different combinations of weak and strong models as they take on varied roles in our subsequent evaluations.

\begin{table*}[htbp]
\resizebox{\textwidth}{!}{
\begin{tabular}{c|c|c|c}
    \textbf{LLM/LVLM} & \textbf{Skills} & \textbf{Training Tasks} & Multimodal?\\
    \hline
    GPT-4o (G4o) & Reasoning; coding & Language modeling, code generation & Yes\\
    GPT-4.1 (G41) & Coding and reasoning & Language, advanced coding tasks, reasoning & Yes\\
    Claude-Sonnet (CS35 or S35) & Strong reasoning, coding, multilingual & General reasoning, coding, text understanding & Yes\\
    Claude-Haiku (Claude 3) & Lightweight, efficient reasoning & General reasoning, summarization, text analysis & Yes\\
    Qwen-VL 2.5 (QVL) & Reasoning & Multimodal reasoning, VQA, captioning, dialogue & Yes\\
    Gemma3 (Ge3 & Reasoning (presumed) & General multimodal tasks, language understanding & Yes\\
    Phi-4 ($\phi$4) & Math and coding reasoning & Language modeling, mathematics, coding & No\\
    Llama3 (L3) & General language understanding & Language modeling, reasoning, dialogue & No\\
\end{tabular}
}
\caption{LLMs and MLLMs that we use within the \name architecture. We review our abbreviations and their capabilities.}
\label{tab:llms}
\end{table*}

\begin{table*}
    \caption{Performance of various individual MLLMs on the SMART-840 dataset (no debate or aggregation).}
    \vspace{-.5em}
    \centering
    \resizebox{0.7\textwidth}{!}{%
    \begin{tabular}{c|c|c|c|c|c|c}
        LLMs / MLLMs &  Size & Capability & Performance & Image+Text & Text & Overall  \\
        \hline
         GPT-4.1 & Unknown & VL & Strong & 48.2 & 88.2 & 60.6\\
         Claude-Sonnet & Unknown &VL & Strong &  35.3 & 71.0 & 44.3\\
         GPT-4o & Unknown & VL & Weak & 33.6 & 67.7 & 42.4\\
         Gemma3  & 12B & VL& Weak & 24.0 & 73.7 & 38.6\\
         Qwen-VL-2.5 & 7B & VL &Weak & 25.0 & 52.5 & 37.2\\
         Phi-4  & 3.8B & L & Weak & N/A & 65.4 & N/A \\
         Llama3 & 8B & L & Weak & N/A & 34.8 & N/A\\
% Orchestrator combinations 
    \end{tabular}
    }
    \label{tab:smart-840-weak-strong-suppl-llms}
\end{table*}

\subsection{Performance versus Computations:} In Table~\ref{tab:smart-840-aggregate-suppl}, we compare the performance of various model combinations within \name under comparable compute budgets and across multiple multi-agent reasoning architectures. Specifically, we evaluate:
i) Single-model aggregation, where each model independently produces one solution and the final answer is selected by majority vote;
ii) Self-Reflect~\cite{shinn2023reflexion}, in which each model performs 8 rounds of self-reflection before ensembling the outputs across models using majority voting;
iii) Self-consistency, similar to Self-Reflect but without feedback—each model is prompted 8 times independently and the aggregated result is used;
iv) Homogeneous feedback, where each model provides feedback to its own solutions for 8 iterative rounds; and
v) RECONCILE, reviewed in the main paper, a multi-agent debate framework using equally capable LLMs.

Table~\ref{tab:smart-840-aggregate-suppl} reports the corresponding results on the SMART-840 benchmark. While these aggregation strategies offer improvements over the single-model baselines in Table~\ref{tab:smart-840-weak-strong-suppl-1}—for example, RECONCILE improves GPT-4.1's accuracy by roughly 2\%—\name consistently delivers substantially larger gains under a similar compute budget. In particular, \name operates with at most four rounds of interaction (averaging about two), involving three solvers and three reflectors, resulting in approximately 30 LLM calls per problem on average (see Figure~\ref{fig:mean_acc} for cross-reference). Despite this modest computational footprint, \name achieves notable performance improvements—approximately 4\% over RECONCILE and more than 10\% over other baselines. Overall, these results highlight that \name not only maintains computational competitiveness but also delivers state-of-the-art multi-agent performance gains.

\subsection{Performance Under Token Limit:} 
\begin{figure}[!ht]
    \centering
    \includegraphics[width=0.45\textwidth,trim={0cm 0.5cm 0cm 1cm},clip]{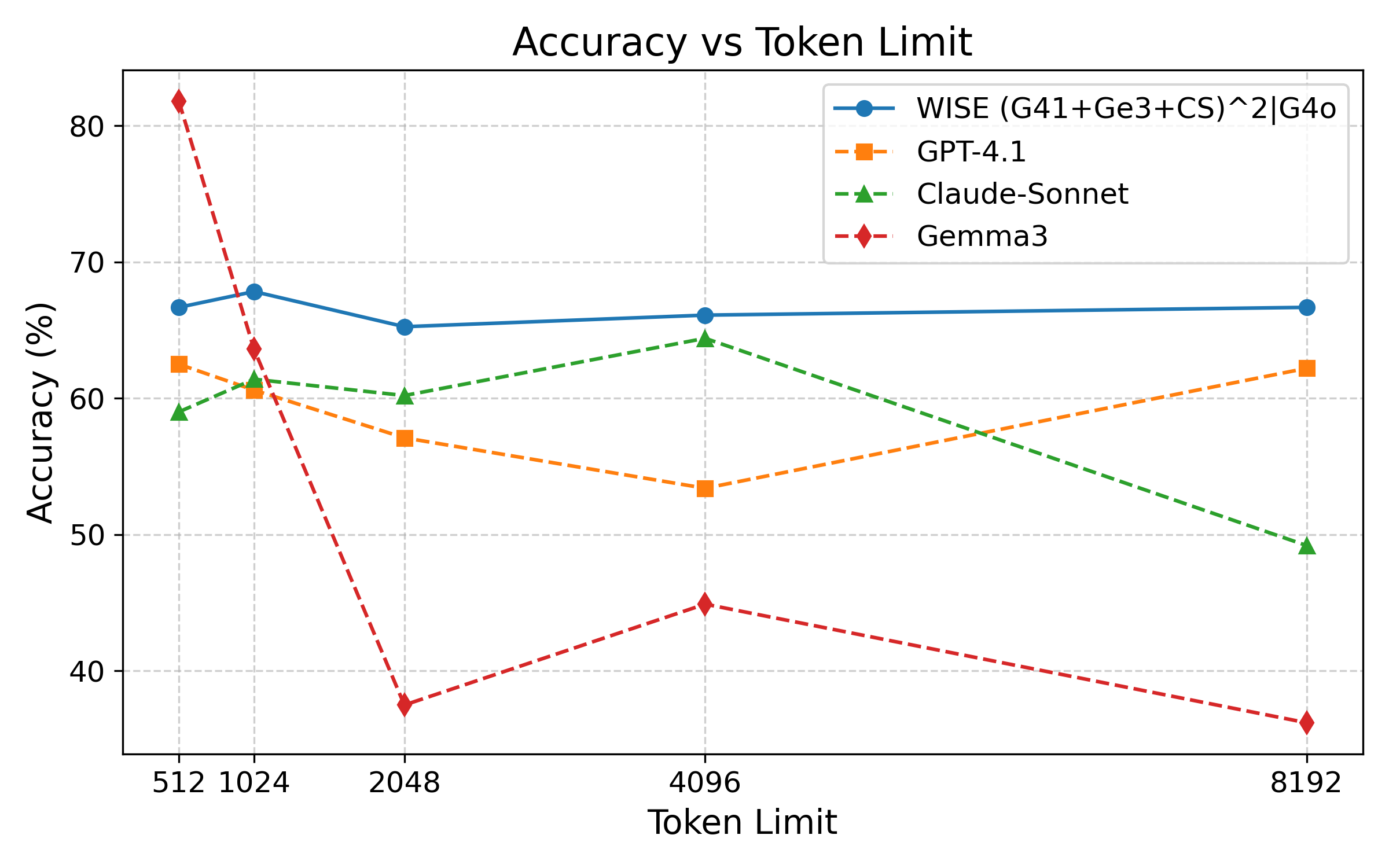}
    \caption{Comparison of token limit versus performance on the 1-2 split of the SMART-840 dataset using \name $(G41+Ge3+CS)^2|G4o$ configuration. We show the performance of respective single MLLMs without any debate for comparison. As is clear, while the individual MLLMs show drastic variations in their performances, \name that combines the varied MLLMs in the debate setup, leads to robust and superior performance, irrespective of the token limits. We used weighted majority voting for the aggregation.}
    \label{fig:accuracy_versus_tokens}
\end{figure}

In Figure~\ref{fig:accuracy_versus_tokens}, we examine the robustness of \name when imposing constraints on the maximum number of tokens available to the underlying models. For each MLLM used within \name, we apply a token cap that we vary from 512 to 8192 tokens. We observe that restricting the token budget noticeably alters the behavior of the individual models, often resulting in substantial fluctuations in their standalone performance.

In contrast, \name shows only minimal performance variance across the same range of token limits. Its multi-model structure effectively balances errors made by individual models, yielding a remarkably stable overall accuracy. For this experiment, we employ weighted majority voting within \name to isolate the effect of token limitations and remove any influence from the \name-Dawid–Skene component. 

\begin{table*}
    \caption{Performance of MAD approaches on SMART-840 using varied MLLM configurations for approximately \textbf{similar number of LLM calls} on average.}
    \vspace{-.5em}
    \centering
    \resizebox{0.9\textwidth}{!}{%
    \begin{tabular}{c|c|c|c|c|c|c|c|c|c|c}
        Method & LLMs / MLLMs & 1\_2 & 3\_4 & 5\_6 & 7\_8 & 9\_10 & 11\_12 & Image+Text & Text & Overall  \\
        \hline        
        Single & G41+G4o+GS35+Ge3+QVL & 53.3 & 50.0 & 46.7 & 67.3 & 55.3 &  60.7 & 45.5 & 82.7 & 55.6 \\
        \hline        
        Self-Reflect & Ge3+QVL+G41 & 48.3 & 47.5 & 42.0 & 62.0 & 58.7 & 64.7 & 42.6 & 83.4 & 53.9\\
        \hline        
        Self-Consistency & G41+CS35+Ge3+QVL & 55.0 & 45.8 & 46.0 & 63.3 & 64.0 & 65.3 & 47.2 & 84.1 & 56.6\\
        \hline        
        Homogeneous & G41+CS35+Ge3+QVL & 49.2 & 48.3 & 38.0 & 58.7 & 62.7 & 66.7 & 43.5 & 81.6 & 53.9\\
        \hline
        \multirow{3}{*}{RECONCILE} & G41 + Ge3 + S35 & 55.0 & 59.2 & 50.0 & 67.3 & 68.7 & 72.0 & 50.9 & 90.1 & 62.0\\        
                         & G4o + Ge3 + QVL  & 50.8 & 41.7 & 36.7 & 54.7 & 56.0 & 62.7 & 39.6 & 76.1 & 50.4\\
                 & Ge3 + QVL	 & 41.7 & 38.7 & 34.0 & 50.7 & 50.0 & 54.7 & 35.8 & 65.5 & 45.0\\
        \hline	
        \multirow{1}{*}{WISE (ours)} 
                & (G41+Ge3+S35)$^2$|G4o & \textbf{65.0} & \textbf{61.8} & \textbf{56.6} & 74.3 & \textbf{75.5} & \textbf{73.0} & \textbf{55.8} & \textbf{91.4} & \textbf{68.1}\\
    \end{tabular}
    }
    \label{tab:smart-840-aggregate-suppl}
\end{table*}

\subsection{Performance on Varied LLM Configurations:} In Table~\ref{tab:smart-840-weak-strong-suppl-1}, we evaluate a range of strong, weak, and language-only model combinations within the \name debate framework. Specifically, we test solver–reflector pairings across five categories: strong–strong, strong–mix, mix–mix, weak–weak, and weak–language-only. Here, mix denotes a role that includes both strong and weak models. All experiments use the same orchestrator model (GPT-4o), and we evaluate performance on the SMART-840 dataset using the WISE-DS aggregation method. As expected, the strong–strong combination achieves high performance. Interestingly, the mix–mix configuration—used in our best-performing setup—achieves a slightly higher score (67.7 → 68.1), suggesting that controlled diversity across roles can be beneficial.

The weak–weak combination performs below all other pairings, as anticipated, yet it still noticeably surpasses the best individual model in the pair. For example, GPT-4o alone achieves 42.4\% accuracy (Table~\ref{tab:smart-840-weak-strong-suppl-llms}), but when used in a weak–weak \name configuration, performance rises to 49.7\%—comparable to RECONCILE when using similarly weak models. Unlike RECONCILE, which assumes all participating models have similar capabilities (e.g., all text-only or all vision–language), \name explicitly assigns models to specialized roles, enabling more flexible and heterogeneous debate setups. This flexibility becomes especially advantageous in the weak–language-only scenario, where \name reaches 51.5\% accuracy. Under similar compute budgets, this is a 1.1\% improvement over RECONCILE using GPT-4o—despite operating entirely in a weak setting.

Overall, our results show that strong–strong configurations predictably benefit \name. However, heterogeneous pairings yield mixed trends:
i) strong–mix combinations where the mix role is dominated by weak reflectors tend to underperform, likely due to lower-quality feedback;
ii) weak–weak and weak–L-only setups can still deliver competitive gains, suggesting that \name’s structured debate can absorb or correct suboptimal weighting from weaker reflectors. These findings highlight the robustness and adaptability of \name in leveraging diverse model capabilities.

\begin{table*}
    \caption{Performance of varied \name on the SMART-840 dataset using varied configurations of the weak and strong LLM models under approximately the \textbf{same number of LLM calls.} $^*$CH3 stands for Claude-Haiku 3.}
    \vspace{-.5em}
    \centering
    \resizebox{0.9\textwidth}{!}{%
    \begin{tabular}{c|c|c|c|c|c|c|c|c|c}
        Solvers & Reflectors & LLMs / MLLMs & 1\_2 & 3\_4 & 5\_6 & 7\_8 & 9\_10 & 11\_12 & Overall  \\
        \hline
Strong & Strong & (G41+S35)$^2$|G4o & 66.7 & 59.1 & 58.2 & 78.9 & 71.4 & 72.2 & 67.7\\
Strong & Weak  &(G41+S35)$\times$(G4o+CH3$^*$+$\phi$4)|G4o & 59.2 & 58.3 & 49.3 & 73.3 & 60.7 & 69.3 & 61.7\\
Mix & Mix   &(G41+Ge3+S35)$^2$|G4o & \textbf{65.0} & \textbf{61.8} & \textbf{56.6} & 74.3 & \textbf{75.5} & \textbf{73.0} & \textbf{68.1}\\
Weak & Weak & (G4o+QVL)$\times$(Q+$\phi$4+L3+G3)|G4o & 51.4 & 39.4 & 41.0 & 53.7 & 53.9 & 58.7 & 49.7\\
Weak & L-only & G4o $\times$ ($\phi$4+o1-mini)|G4o & 49.2 & 45.4 & 47.1 & 54.6 & 56.9 &  55.8 & 51.5\\
% Strong & mix & 
% Strong & Strong & 
\hline 
% Orchestrator combinations 
    \end{tabular}
    }
    \label{tab:smart-840-weak-strong-suppl-1}
\end{table*}

\subsection{Comparisons on Orchestrator Influence:} A key and novel component of \name is the orchestrator LLM and the expanded role it plays within the multi-agent debate. In prior work~\cite{chen2023reconcile,bai2022constitutional,liang2023encouraging}, the orchestrator’s primary function is to summarize each debate round before passing the summary to the agents in the next round. In contrast, the orchestrator in \name performs two critical functions:
i) summarizing the reflectors’ feedback, and
ii) generating actionable, targeted follow-up questions for the solvers.
These questions help direct the solvers’ attention to specific multimodal aspects highlighted by the reflectors, enabling deeper and more focused analysis in subsequent rounds. To evaluate the importance of this enhanced orchestrator role, we conducted two sets of studies:
i) replacing GPT-4o (our default orchestrator) with a weaker model, Phi-4, and
ii) modifying the orchestrator so that it only summarizes feedback without generating follow-up questions.

Table~\ref{tab:smart-840-orchestrator} reports the results on the SMART-840 dataset (split 1-2) using two debate configurations: strong+mix and strong–strong. Rows marked (w/o Q) correspond to the orchestrator variant that does not generate actionable questions. The results show that replacing GPT-4o with Phi-4 causes little to no performance degradation, indicating that \name is not strongly dependent on a powerful orchestrator model. However, removing the question-generation step results in a consistent 3–4\% drop in accuracy across both configurations. This clearly demonstrates that the orchestrator’s ability to craft targeted follow-up questions is essential for driving effective multi-agent reasoning within \name.

\begin{table}
    \caption{Performance of varied \name on the SMART-840 dataset with varied orchestrators configurations (using wt. majority voting).}
    \vspace{-.5em}
    \centering
    \resizebox{0.4\textwidth}{!}{%
    \begin{tabular}{c|c|c}
         \name LLMs / MLLMs & Orchestrator & 1\_2    \\
        \hline
   (G41+Ge3+S35)$^2$ & G4o & 65.0  \\
   (G41+Ge3+S35)$^2$ & G4o (w/o Q) & 62.1 \\
   \hline
   (G41+Ge3+S35)$^2$ & Phi-4 & 64.8\\
   (G41+Ge3+S35)$^2$ & Phi-4 (w/o Q) & 61.2 \\
   \hline
   (G41+S35)$^2$ & G4o & 66.7  \\
   (G41+S35)$^2$ & G40 (w/o Q) & 63.6 \\
   \hline
   (G41+S35)$^2$ & Phi-4 & 65.7  \\
    (G41+S35)$^2$ & Phi-4 (w/o Q) & 61.7 \\
% Strong & mix & 
% Strong & Strong & 
\hline 
% Orchestrator combinations 
    \end{tabular}
    }
    \label{tab:smart-840-orchestrator}
\end{table}
\vspace{0.25cm}

\subsection{Accuracy vs Rounds:} 
\begin{figure*}[ht]
    \centering
    \begin{subfigure}{5.5cm}
        \includegraphics[width=\textwidth]{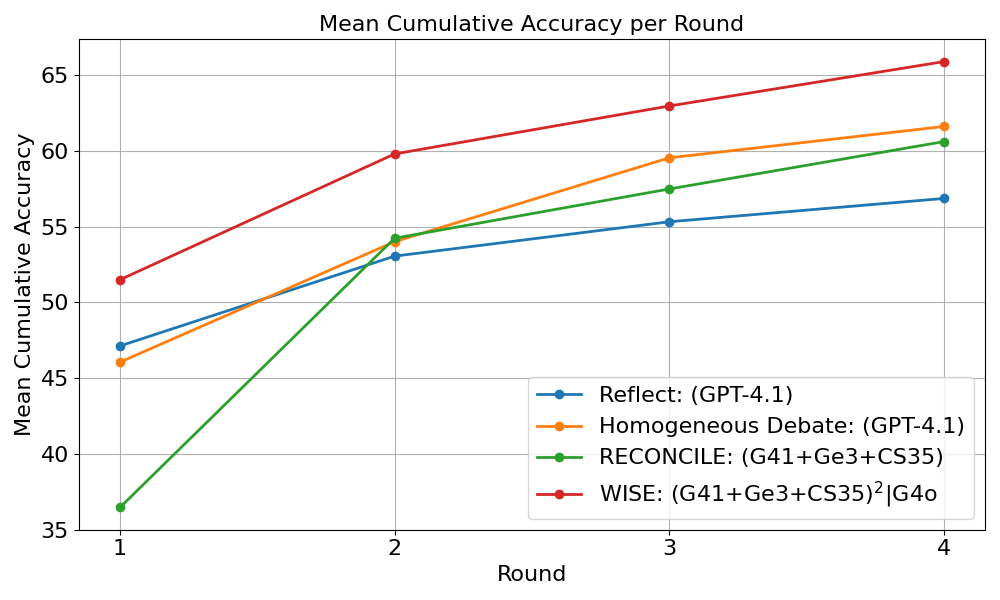}
        \caption{VL Performance}
        \label{fig:evochart}
    \end{subfigure}
    \begin{subfigure}{5.5cm}
        \includegraphics[width=\textwidth]{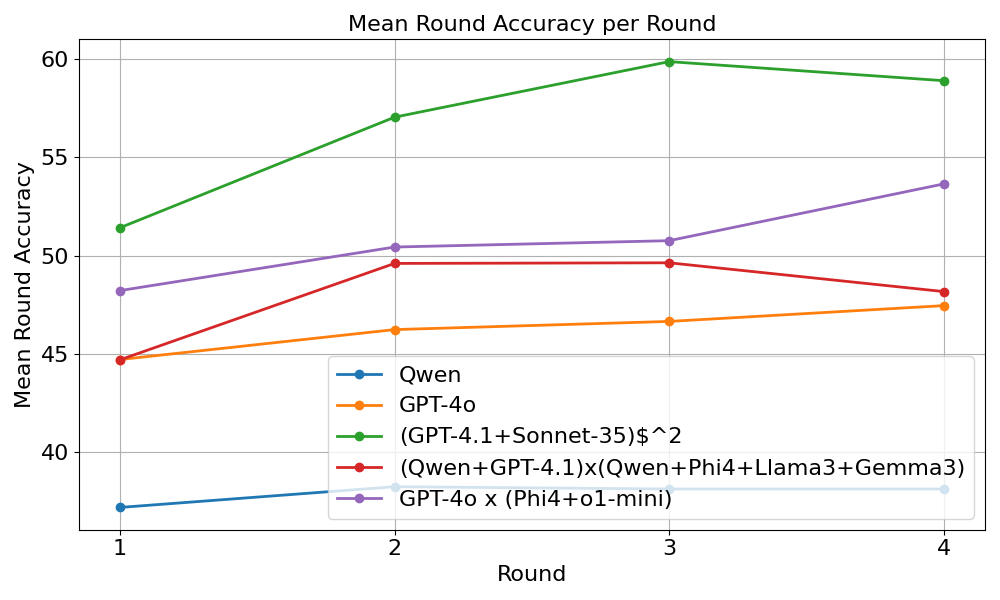}
        \caption{Acc. on Weaker Models}
        \label{fig:mean_acc-suppl}
    \end{subfigure}
    \begin{subfigure}{5.5cm}
        \includegraphics[width=\textwidth]{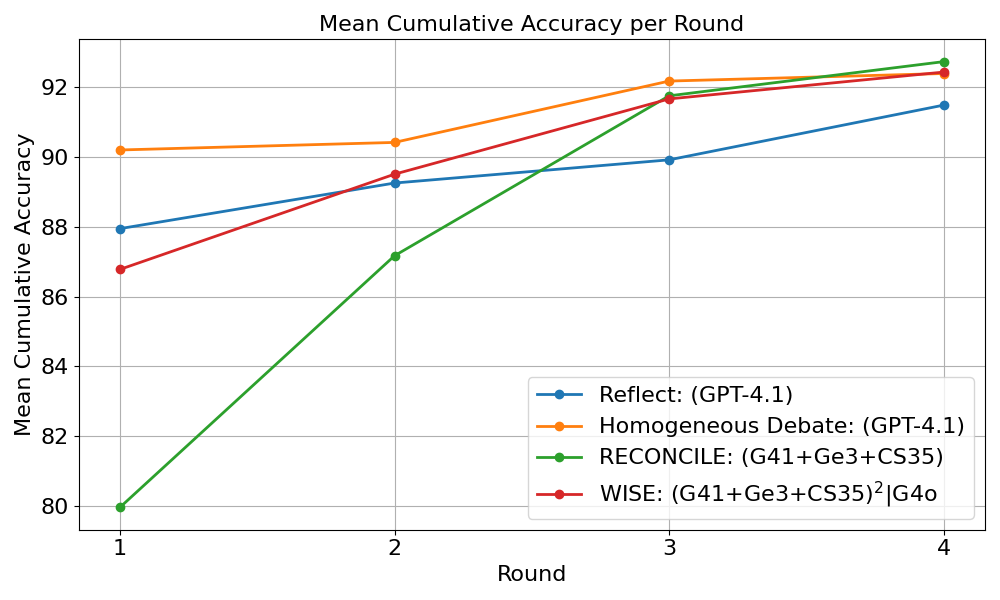}
        \caption{Text-only Performance}
        \label{fig:cum_acc-suppl}
    \end{subfigure}
    
    \caption{(a) plots the VL performance of various MAD setups on the SMART-840 dataset. In (b), we compare the average round performance against then number of rounds on various strong and weak LLM configurations. In (c), we plot the performance using only text-only problems. Comparing (a) and (c), we find that text-only performance of various MAD configurations are very high, and \name closely matches the performance. However, on multimodal problems, \name clearly wins.   }
    \label{fig:accuracy_versus_rounds}
\end{figure*}

In Figure~\ref{fig:accuracy_versus_rounds}, we plot the accuracy against the number of debate rounds. We plot the cumulative accuracy, which measures for a round $k$ if the correct solution to a problem was found in some round from 1..$k$. This serves as the upper-bound on the accuracy, and shows if by increasing the rounds problems can be solved. As seen in Figure~\ref{fig:cum_acc-suppl}, the performance of the heterogeneous debate is seen to increase with the increasing number of rounds, suggesting that models that were wrong in previous rounds could correct their mistakes. In Figure~\ref{fig:mean_acc-suppl}, we plot the average accuracy over the answer selected against the ground truth. As can be seen, all the debate configurations demonstrate monotonic increase in the accuracy with the rounds. We found that using no-debate does not show much promise, while using homogeneous debate with a strong LLM (GPT-4o) closely matches the performance of the heterogeneous debate with (GPT-4o+Qwen-VL)$\times$ (Gemma3+Llama3+Phi4+Qwen-VL) with slightly higher performance than the former with more debate rounds. We also find that the stronger model (combination of GPT and Claude-3 models) perform the best, however even for them, the performance increases by nearly 10\% from the first round to the third from nearly 55\% to 65\%. 

\begin{figure*}[htbp]
    \centering
    \begin{subfigure}{0.32\linewidth}
        \includegraphics[width=\linewidth]{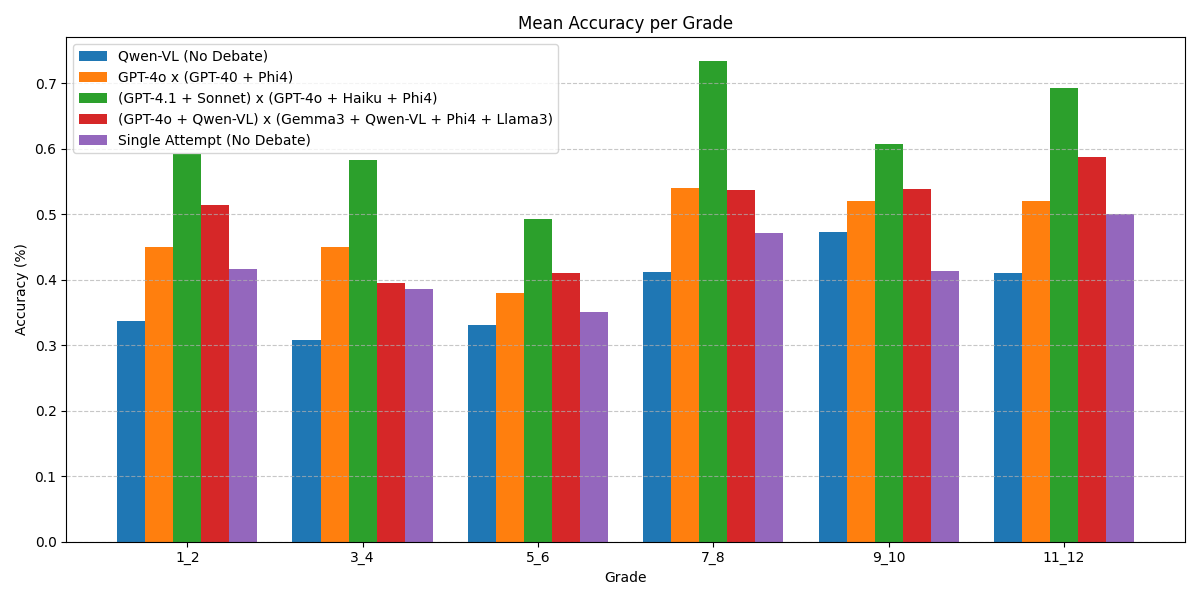}
        \caption{Performance per Grade}
        \label{fig:acc_grade}
    \end{subfigure}
    \begin{subfigure}{0.32\linewidth}
        \includegraphics[width=\linewidth]{figs/all/comparing_mean_best_rounds_summary_text.png}
        \caption{Avg Rounds per Grade}
        \label{fig:rounds_grade}
    \end{subfigure}
    \begin{subfigure}{0.32\linewidth}
        \includegraphics[width=\linewidth,trim={0 0 0 0cm},clip]{figs/all/comparing_mean_best_rounds_summary_text.png}
        \caption{Avg. Best Rounds}
        \label{fig:mean_acc}
    \end{subfigure}
    \caption{Plot of the average performance per grade over the 8 rounds of debate and average number of rounds per grade.}
    \label{fig:perf_hist}
\end{figure*}

\subsection{Grade-wise Performance:} In Figure~\ref{fig:acc_grade}, we plot the mean performance of per class grade as per the split in the SMART-840 dataset. We find that the stronger models demonstrate a clear advantage over all the grades, while the combination of strong and weak models fare higher than the homogeneous debate model in 4 out of the 6 grades, and tying in one grade (7\_8). We also find that Single Attempt in GPT model is better than multiple (repeated sampling) attempts using the Qwen-VL model, suggesting that solving a majority of the problems on SMART-840 is beyond the capability of Qwen-VL (even when the problems are expected to be solvable by children). We also find that when pairing Qwen-VL with GPT-4o as the solvers, and when using \name MAD, the performance of the combination is better than the performance of either GPT-4o or Qwen-VL, which we believe is an important result attesting to the usefulness of our debate scheme. 

\subsection{Rounds vs. MAD Complexity:} In Figure~\ref{fig:accuracy_versus_rounds}, we provide a bar plot showing the average number of rounds taken by a debate configuration to reach consensus. Recall that we stop the debate when the orchestrator finds that all the Solvers produce the same answer and all the Reflectors agree that the solution is correct, even when the predicted solution may not match the ground truth. As can be seen from the plot, having more LLMs in the debate (e.g., the red bar) demonstrates a significantly longer number of rounds to converge (nearly 4 rounds) while the rounds average around 1.5 for the strong configuration (green bar). We show round = 1 for single attempt and repeated sampling variants. We also find that the number of debate rounds is lower for the homogeneous variant, suggesting  the lack of ability in the model to understand its own mistakes to be corrected (homogeneous configuration provides feedback on its own previous solution).

\subsection{Performance on Text-Only Problems:} In Figures~\ref{fig:cum_acc}, we plot the summarized performance for all text-only problems in the SMART-840 dataset. From the figure, we find the performance of the models on text-only problems is nearly 20\% higher than on image-text + text-only problems -- suggesting the strong models are almost close to solving all the text-only problems in the dataset correctly (which amounts to nearly 30\% of the problems). Despite this significant performance, we find that homogeneous models (that includes GPT-4o) still lags below by nearly 16\% (73\% -> 89\%) when combined with Claude-3, and that the gap between the cumulative and the average performance on this subset if merely 4\%, suggesting the debate is able to near closing the gap, yet the \name debate is useful. 

\section{Prompts used in our MAD formulations}
In Table~\ref{tab:prompts}, we provide the varied prompts (described in the main paper) used at various stages of our framework. Please also see the suppl.zip attachment that provides detailed solution responses, feedbacks, and orchestrator summaries for all the four datasets we used. 

\begin{table*}[htbp]
    \setlength{\tabcolsep}{6pt}
    \renewcommand{\arraystretch}{1.0}
    \begin{tabularx}{\textwidth}{L{0.18\textwidth}|Y}
    \textbf{Prompt Type} & \textbf{Prompt text} \\
    \hline
    $\sprompt_1$ (Solver) &
    Please describe the problem, including the images if available. Your description of the image should be detailed and accurate so that an LLM without access to the image can solve the problem. After this, please solve this question with explanation of the intermediate steps. You must select an answer from one of the five options: 
    \textbf{Puzzle:} [PROBLEM]...[/PROBLEM] 
    
    \textbf{Options:} [OPTIONS] ... [/OPTIONS]. 
    
    Write the final \textbf{answer option} as one of \texttt{A1, B2, C3, D4, E5}. \\
    \hline
    $\rprompt$ (Reflector) &
    Following are the solutions of LLM in solving a math puzzle. The problem statement and the LLM solution are provided below. The problem may contain an image that needs to be used to understand the LLM's answer. You should not solve the problem yourself. Instead, please check if the solutions below are logically and mathematically consistent and the final answer is correct? 
    
    [PROBLEM] ... [/PROBLEM]. 
    
    [RESPONSE] ... [/RESPONSE]. 
    
    Your Task: Provide a detailed explanation of your assessment so that it can be used as feedback to LLM to improve the solution, including any mistakes in understanding the given image (if any). 
    
    Your response should end with a single line in the format:
    
    $FINAL\_SCORE: \langle value\rangle$ 
    
    where $\langle value \rangle$ is 0 if final answer is the wrong, 1 if you are unable to confirm, and 2 if the final answer is correct. 
    \\
    \hline
    $\oprompt$ (Orchestrator) & 
    The following are the feedback received from LLMs on a puzzle solution. There could be significant differences in the feedback provided by different LLMs, especially the LLMs may select different answer options. Your task is to pay attention to these differences in the feedback and the selected answer options. Please summarize all the feedback and provide a \textbf{set of actionable questions} for the solver to revise its solution in the next round. It is possible that the previous solution might have overlooked details in the given image (if any), and thus your summary should emphasize the importance of re-evaluating the image.
    
   [FEEDBACK] [/FEEDBACK]
   
   [FEEDBACK] [/FEEDBACK]
   ...
    \\
    \hline
    $\sprompt_{k>1}$ (Solver) & Following is the feedback received on your previous solution. If there was a mistake in your previous solution, use this feedback and the list of actionable steps provided to reconsider and re-evaluate your solution. You should re-evaluate your understanding the given image (if any) and the problem statement using the feedback. Specifically, based on the provided feedback, please provide a detailed explanation of the given image (if any). Please provide the final answer option and explanation of the intermediate steps, as well as details of how the feedbacks were addressed in your revised final answer. Your final answer should be reported in your final line and must select one of the answer options: A1, B2, C3, D4, E5.
    
    [PROBLEM] ... [/PROBLEM] [OPTIONS] ... [/OPTIONS]
    
    [RESPONSE] ... [/RESPONSE]
    [FEEDBACK] ... [/FEEDBACK]
    \\
    \end{tabularx}
    \caption{Various prompts used in the \name MAD formulation.}
    \label{tab:prompts}
\end{table*}

\begin{figure}[ht]
    \centering
    \includegraphics[width=0.9\linewidth,trim={0cm 2.5cm 0cm 1cm},clip]{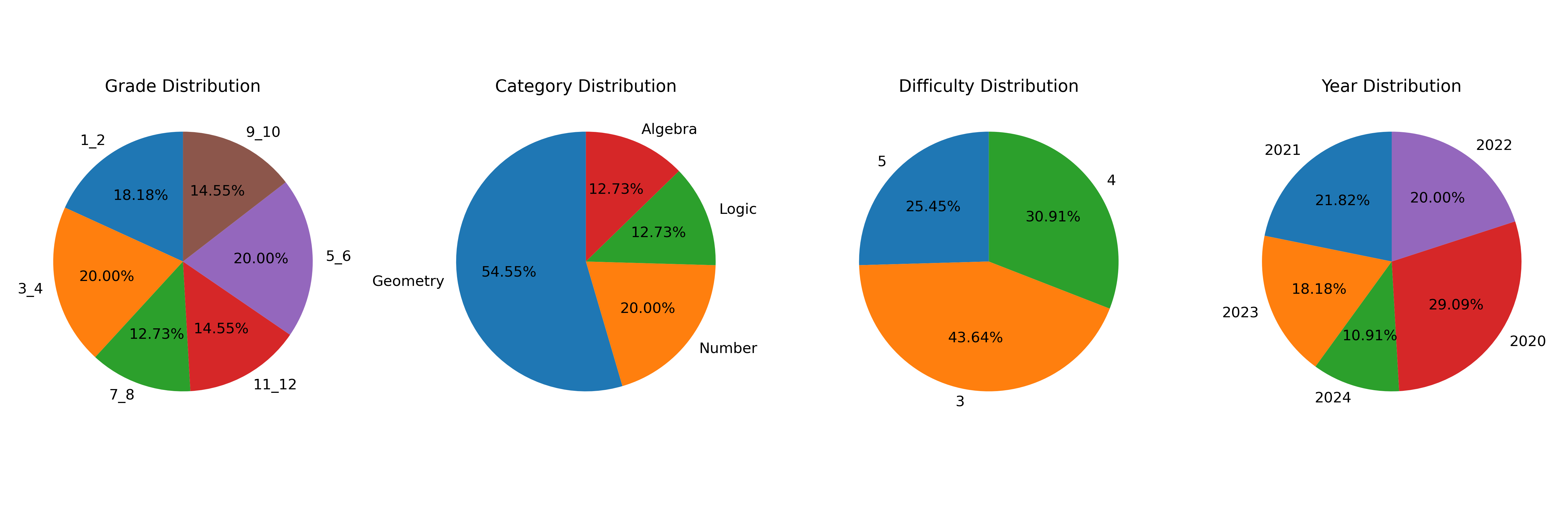}
    \caption{Distribution of problems in the SMART-840++ dataset. We also show the distribution of puzzles across grades, categories, difficulty levels, and years of MK competition. Note that the difficulties of the puzzles shown in the third plot above are with regards to the SMART-840 difficulty categories (i.e., 3: simple, 4: medium, 5: hard), which reflects the difficulty in algorithmic derivation. In SMART-840++, each puzzle is augmented into 5 new difficulty levels irrespective of them being in the original three difficulty categories, where the `difficulty' in SMART-840++ predominantly reflects the scale of the problem irrespective of algorithmic difficulty.}
    \label{fig:smart840++-statistics}
\end{figure}

\section{SMART-840++ Dataset Details}
\subsection{Motivation:} As alluded to in the main paper, the key motivation for proposing the SMART-840++ dataset is to understand and analyze the capability of state-of-the-art multimodal large language models over three axes: i) generalization to never seen before problem content, ii) generalization to  complex problem settings with controlled increase in the problem difficulty, and iii) robustness in performance to minor variations to the problem objectives. We note that our analysis uses publicly available problems from the SMART-840 dataset, presented in~\cite{cherian2024evaluating}. The SMART-840 dataset contains 840 problems extracted from the Math Kangaroo (MK) Olympiads\footnote{\url{mathkangaroo.org}} from years 2020-2025. The problems span four areas: geometry, algebra, numerics, and logic, and is segregated into 6 categories, one for a pair of consecutive school grades from 1--12. We note that similar attempts at problem augmentation has been attempted to before, e.g., 
GSM-Symbolic~\cite{mirzadeh2024gsm}, Zebra-Logic~\cite{lin2025zebralogic}, and SMART-101~\cite{cherian2023deep}. While the first two prior works are for text-only domain, SMART-101 is for vision-language problems however is limited to only simple problems from 1--2 grades of the MK Olympiads. Through SMART-840++, we go beyond SMART-101 and propose new problems across grade curriculum.

\begin{figure*}[ht]
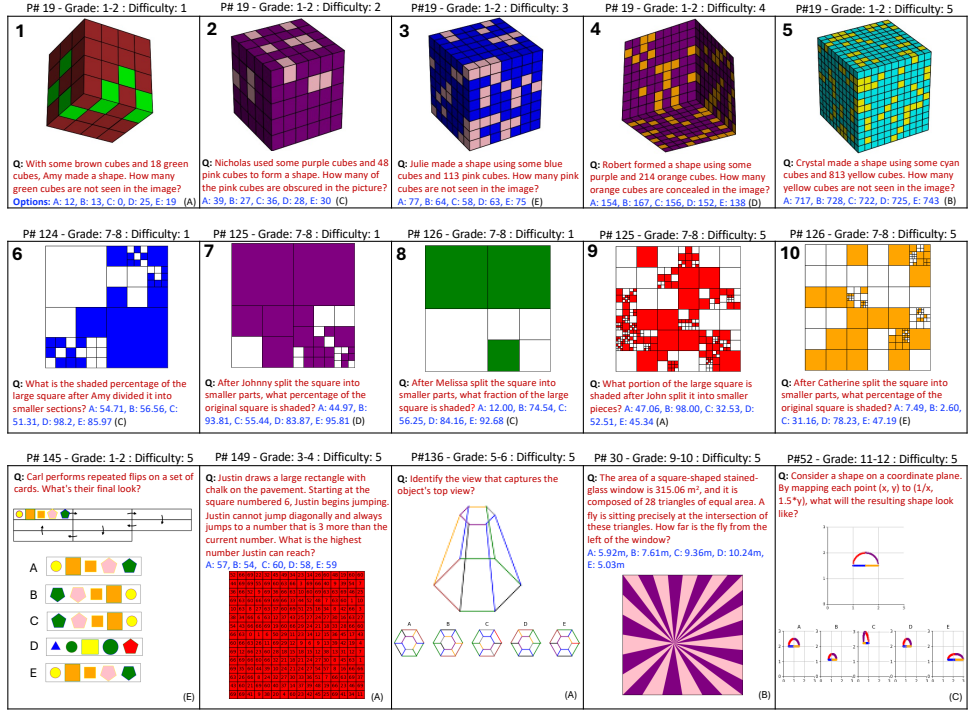

    \centering
    \includegraphics[width=13cm, trim={0.5cm 5.5cm 1cm 3cm}, clip]{figs/smart840_samples_1.pdf}
    \includegraphics[width=13cm, trim={0.5cm 10cm 1cm 4cm}, clip]{figs/smart840_samples_2.pdf}
    \caption{Example problems from the SMART-840++ dataset. We show variations in the newly generated puzzles across appearance and problem complexity, as well as diversity in problem text and options. For the below figure, we denote $D-$ to denote the difficulty levels for instances of the same problem. See Figure~\ref{fig:smart-840-ext-full-list} for the full list of problems later in this supplementary materials.}
    \label{fig:smart-840++-suppl-samples}
\end{figure*}

\subsection{Problem Selection:}To construct the SMART-840++ dataset, we selected 55 vision-and-language problem types, approximately balanced across grade levels. Because manually authoring large numbers of problems is labor-intensive and difficult to scale, we prioritized problems that could be programmatically regenerated using Python-based graphics toolboxes. Our approach is inspired by~\cite{cherian2023deep}, which demonstrated programmatic reconstruction for MK problems, though limited to grades 1–2.

Programmatic synthesis offers several key advantages:
i) it enables the generation of arbitrarily many problem instances;
ii) by adjusting program parameters, we can control difficulty, such as using a 3×3 grid for easier tasks or a 10×10 grid for harder ones—both governed by the same underlying algorithm, but requiring different numbers of tokens and longer reasoning chains; and
iii) it allows rapid creation of visual variety, including changes in colors, viewpoints, textures, and layouts.

\begin{figure}
\fbox{%
\begin{minipage}{0.95\linewidth}
\begin{flushleft}
\textbf{SMART-840++ Problem \#8 -- mk-2024-3-4-11: }
Tina created a block tower, as seen in the image. Then they remove the 1st block from the top. Then they remove the 3rd block from the top. Then they remove the 1st block from the top. Then they remove the 2nd block from the bottom. Which tower does Tina end up making? \\
\end{flushleft}
\centering
\includegraphics[width=0.75\textwidth,trim={1cm 1cm 0cm 1cm},
  clip]{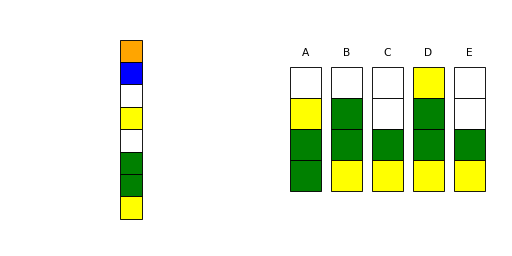} \\
\begin{flushleft}
A: A, B: B, C: C, D: D, E: E \\
\textbf{Answer:} C \\
\end{flushleft}
\end{minipage}%
}
% \end{figure}

% \begin{figure}
\fbox{%
\begin{minipage}{0.95\linewidth}
\begin{flushleft}
\textbf{SMART-840++ Problem \#9 -- mk-2020-9-10-20: }
Imagine a $square-shaped$ stained glass window with an area of 60.22 $dm^2,$ divided into 4 equal triangles. A fly is sitting at the exact point where all triangles converge. How far is the fly from the bottom of the window? \\
\end{flushleft}
\centering
\includegraphics[width=0.25\textwidth,trim={1cm 1cm 0cm 1cm},
  clip]{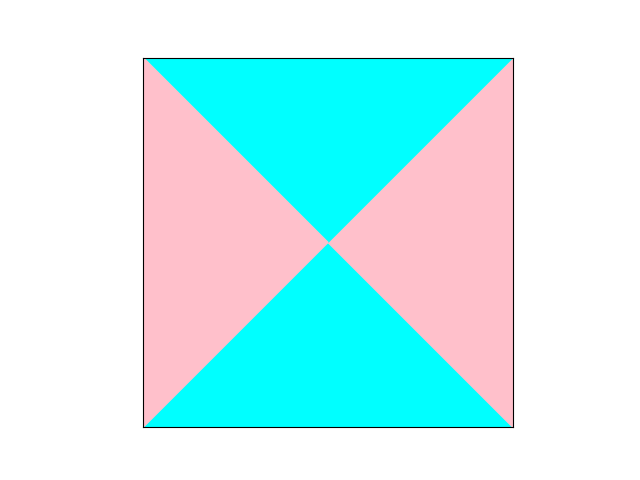} \\
\begin{flushleft}
A: 6.26dm, B: 5.95dm, C: 3.88dm, D: 7.56dm, E: 7.86dm \\
\textbf{Answer:} C \\
\end{flushleft}
\end{minipage}%
}
% \end{figure}

% \begin{figure}
\fbox{%
\begin{minipage}{0.95\linewidth}
\begin{flushleft}
\textbf{SMART-840++ Problem \#10 -- mk-2020-3-4-4: }
Which figure, drawn by Thomas, contains the largest shaded section? \\
\end{flushleft}
\centering
\includegraphics[width=0.5\textwidth,trim={1cm 1cm 0cm 1cm},
  clip]{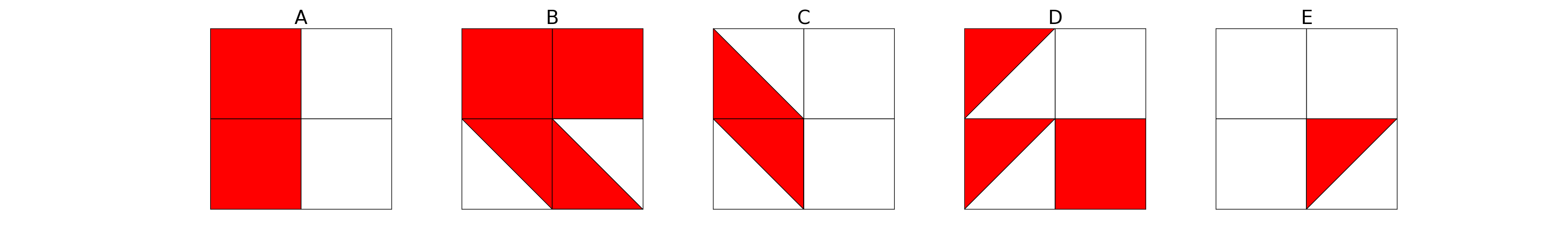} \\
\begin{flushleft}
A: A, B: B, C: C, D: D, E: E \\
\textbf{Answer:} B \\
\end{flushleft}
\end{minipage}%
}
% \end{figure}

% \clearpage
% \begin{figure}
% \begin{figure}[t]
\centering
\fbox{% 
\begin{minipage}{0.95\linewidth}
\begin{flushleft}
\textbf{SMART-840++ Problem \#11 -- mk-2023-3-4-4: }
Which two pieces can Mrs. put together to match the shape on the right? \\
\end{flushleft}
\centering
\includegraphics[width=0.9\textwidth,trim={1cm 1cm 0cm 1cm},
  clip]{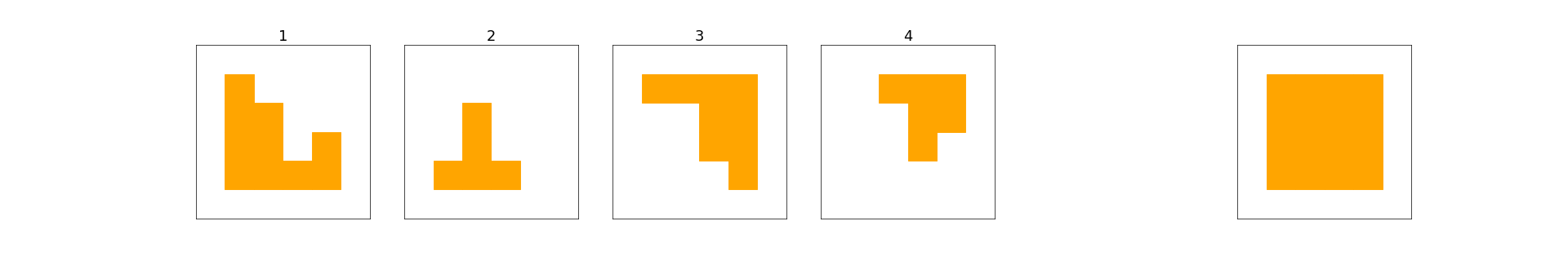} \\
\begin{flushleft}
A: 3 and 4, B: 4 and 2, C: 4 and 1, D: 2 and 3, E: 2 and 1 \\
\textbf{Answer:} C \\
\end{flushleft}
\end{minipage}%
}
\caption{SMART-840++ Problem \#11 -- mk-2023-3-4-4.}
\label{fig:smart-840-ext-full-list}
\end{figure}

Our overarching goal is to create novel problems that LLMs are unlikely to have encountered in training, thereby requiring genuine algorithmic reasoning that is invariant or agnostic to superficial variations in appearance or difficulty. Sample problem instances from SMART-840++ are shown in Figure~\ref{fig:smart-840++-suppl-samples}. A full set containing one example instance from each of the 55 problem types appears at the end of this supplementary material (Figure~\ref{fig:smart-840-ext-full-list}). Dataset statistics are summarized in Figure~\ref{fig:smart840++-statistics}. Each problem includes an image, a text question, and five answer options (image-based or text-based), with exactly one correct answer.

\subsection{Algorithmic Augmentation Pipeline} As described above, we use Python to implement our augmentation pipeline for selected problems from the SMART-840 dataset. For each original problem, we first solved it manually, then wrote a Python program that (i) semantically reconstructs the problem and (ii) implements the corresponding solution logic. Each program is parameterized so that modifying its arguments generates new problem instances with arbitrary configurations, difficulty levels, and visual appearances. When visual attributes are not essential to the reasoning (e.g., colors, sizes, or minor layout choices), we randomly sample them from predefined sets.

For the text component, we begin by constructing a question template derived from the original problem, substituting puzzle-specific values where appropriate. We then use an LLM (e.g., Microsoft Copilot) to rephrase the text—using prompt constraints that preserve the underlying logical structure while allowing variations in style, tone, or character names. The resulting problem instances thus differ in appearance and wording from the originals, yet retain the same (or nearly the same) underlying algorithmic reasoning structure.

\section{\name-DS Expectation Maximization}
Repeating the approach for \name-Dawid-Skene given in the main paper, we provide more details here on how the ensued Expectation Maximization formulation is derived and how it is solved. 

\subsection{Derivation of EM for \name-DS:}
As noted, given $N$ worker responses, each selecting one of $K$ possible answers for a given problem, the classic Dawid–Skene (DS) algorithm employs expectation–maximization (EM) to jointly estimate the error rates of workers and the latent ground-truth label distribution. In our context, assuming conditional independence of agents’ responses, we cast our MAD formulation within the DS framework to model agent-specific error rates and compute the posterior over solution responses. This allows us to down-weight agents that behave randomly (or bluffing) while amplifying the influence of those whose outputs are consistently aligned with the majority.

Suppose $\prob_{\alpha\beta}$ is the probability of selecting answer $\beta$ for the true answer $\alpha$ by a model, and if $\zeta_\alpha$ is the true prior probability of selecting an option $\alpha$, then given $\Lambda$ problems and the agents' responses, \cite{dawid1979maximum} models the likelihood of the data as a mixture of multinomial distributions:
\begin{equation}
\prob(\text{data}) \propto \prod_{i=1}^\Lambda\sum_{\alpha=1}^K\zeta_\alpha\left[\prod_{j=1}^N\prod_{\beta=1}^K (\prob^j_{\alpha\beta})^{\lambda^j_{i\beta}}\right],
\end{equation}
where $\lambda^j_{i\beta}$ is the number of times $j$-th agent produced $\beta$ as the answer against the true answer of $\alpha$ if the model is run multiple times on the same problem $i$. As both the true priors $\zeta_\gamma$ and the error rates $\prob_{\alpha\beta}$ are unknown, DS uses EM to optimize the likelihood iteratively towards convergence. 

In \name, we have two sources of errors: i) solvers making errors in selecting the correct answers from the $K$ choices and the reflectors selecting one of $J (=3)$ weights. Suppose $\prob^{t}_{\alpha_1\beta_1}$ and $\prob^{c}_{\alpha_2\beta_2}$ denote the error matrices for the solver and the reflectors respectively, in selecting option $\beta_1\to\alpha_1$ (for $\alpha_1,\beta_1\in[K]$) and selecting wrong weights for a provided answer, i.e., selecting weight $\beta_2\to\alpha_2$ (for $\alpha_2,\beta_2\in[J]$), then the joint log-likelihood for estimating the combined error matrices and their joint true priors $\zeta_{\alpha\beta}=\zeta_\alpha\zeta_\beta$ could be modeled for debate round $k$ as the product of two mixtures of multinomial distributions, given by:

\begin{align}
\log \mathcal{L}
&= \sum_{i=1}^\Lambda 
    \log \sum_{\alpha=1}^J \sum_{\beta=1}^K 
    \zeta_\alpha \zeta_\beta
    \prod_{t,\beta_1}
    \bigl(P^t_{\beta,\beta_1}\bigr)^{\lambda^{S_t}_{i\beta_1}}
    \prod_{c,\beta_2}
    \bigl(P^c_{\alpha,\beta_2}\bigr)^{\lambda^{R_c}_{i\beta_2}} \\
&= \sum_{i=1}^\Lambda 
    \log \sum_{\alpha=1}^J \sum_{\beta=1}^K 
    \zeta_\alpha \zeta_\beta
    \prod_{t,\beta_1}
    \bigl(P^t_{\beta,\beta_1}\bigr)^{\lambda^{S_t}_{i\beta_1}}
    \prod_{c,\beta_2}
    \bigl(P^c_{\alpha,\beta_2}\bigr)^{\lambda^{R_c}_{i\beta_2}}.
\end{align}

where $\alpha \in \{1,\ldots,J\}$ denote the latent true \emph{rank/weight} label 
and $\beta \in \{1,\ldots,K\}$ denote the latent true \emph{solution/answer} label. 
We assume factorized priors 
\begin{equation}
\zeta_{\alpha\beta} = \zeta_\alpha\,\zeta_\beta, 
\qquad 
\text{s.t. } \sum_{\alpha=1}^J \zeta_\alpha = 
\sum_{\beta=1}^K \zeta_\beta = 1.
\end{equation}
Each \emph{solver} $t \in \mathcal{S}$ emits an answer $\beta_1 \in \{1,\ldots,K\}$ 
with confusion matrix 
$P^t_{\beta,\beta_1} = P(\text{emit }\beta_1 \mid \text{true }\beta)$,
and each \emph{critic/reflector} $c \in \mathcal{R}$ emits a rank $\beta_2 \in \{1,\ldots,J\}$ 
with confusion matrix 
$P^c_{\alpha,\beta_2} = P(\text{emit }\beta_2 \mid \text{true }\alpha)$.
For item $i$, let $\lambda^{S_t}_{i\beta_1}$ and $\lambda^{R_c}_{i\beta_2}$ denote the observed counts
of solver and critic emissions, respectively.

Given independence of solvers and critics conditioned on $(\alpha,\beta)$,
the likelihood of observations for item $i$ is
\begin{equation}
P(\text{obs}_i \mid \alpha,\beta)
= \prod_{t \in \mathcal{S}} \prod_{\beta_1=1}^K
\big(P^t_{\beta,\beta_1}\big)^{\lambda^{S_t}_{i\beta_1}}
\;\cdot\;
\prod_{c \in \mathcal{R}} \prod_{\beta_2=1}^J
\big(P^c_{\alpha,\beta_2}\big)^{\lambda^{R_c}_{i\beta_2}}.
\end{equation}
The observed-data likelihood over $\Lambda$ items is therefore
\begin{equation}
\mathcal{L}(\Theta)
= \prod_{i=1}^\Lambda 
\sum_{\alpha=1}^J \sum_{\beta=1}^K 
\zeta_\alpha \zeta_\beta \;
P(\text{obs}_i \mid \alpha,\beta),
\end{equation}
where $\Theta = \{\zeta_\alpha, \zeta_\beta, P^t, P^c\}$.

\paragraph{E--step.}
For each item $i$, the posterior responsibility of the latent pair $(\alpha,\beta)$ 
is given by
\begin{align*}
&\gamma_{i,\alpha\beta}
= P(\alpha,\beta \mid \text{obs}_i)
= \\
&\frac{
\zeta_\alpha \, \zeta_\beta \,
\displaystyle
\left[
\prod_{t \in \mathcal{S}} \prod_{\beta_1=1}^K
\big(P^t_{\beta,\beta_1}\big)^{\lambda^{S_t}_{i\beta_1}}
\right]
\left[
\prod_{c \in \mathcal{R}} \prod_{\beta_2=1}^J
\big(P^c_{\alpha,\beta_2}\big)^{\lambda^{R_c}_{i\beta_2}}
\right]
}{
\displaystyle
\sum_{\alpha'=1}^J \sum_{\beta'=1}^K
\zeta_{\alpha'} \, \zeta_{\beta'} \,
\left[
\prod_{t \in \mathcal{S}} \prod_{\beta_1=1}^K
\big(P^t_{\beta',\beta_1}\big)^{\lambda^{S_t}_{i\beta_1}}
\right]
\left[
\prod_{c \in \mathcal{R}} \prod_{\beta_2=1}^J
\big(P^c_{\alpha',\beta_2}\big)^{\lambda^{R_c}_{i\beta_2}}
\right]
}.
\label{eq:responsibility}
\end{align*}

The corresponding marginal posteriors for the true solution and rank are obtained by summing over the other latent variable:
\begin{equation}
p_i^{t}(\beta) = \sum_{\alpha=1}^J \gamma_{i,\alpha\beta},
\qquad
p_i^{c}(\alpha) = \sum_{\beta=1}^K \gamma_{i,\alpha\beta}.
\end{equation}

\paragraph{M--step.}
Define the expected counts
\begin{equation}
N_{\alpha\beta} = \sum_{i=1}^\Lambda \gamma_{i,\alpha\beta}, 
\quad 
N_\alpha = \sum_{\beta} N_{\alpha\beta}, 
\quad 
N_\beta = \sum_{\alpha} N_{\alpha\beta}.
\end{equation}
The factorized priors are updated as
\begin{equation}
\zeta_\alpha^{\text{new}} = 
\frac{N_\alpha}{\sum_{\alpha'} N_{\alpha'}},
\qquad
\zeta_\beta^{\text{new}} =
\frac{N_\beta}{\sum_{\beta'} N_{\beta'}}.
\end{equation}
The solver and critic confusion matrices are updated as weighted multinomial maximum likelihoods:
\begin{align}
&P^{t,\text{new}}_{\beta,\beta_1} 
= 
\frac{
\sum_{i} \left( \sum_{\alpha} \gamma_{i,\alpha\beta} \right)
\lambda^{S_t}_{i\beta_1}}
{\sum_{\beta_1'} \sum_{i} 
\left( \sum_{\alpha} \gamma_{i,\alpha\beta} \right)
\lambda^{S_t}_{i\beta_1'}},
\\
&P^{c,\text{new}}_{\alpha,\beta_2} 
= 
\frac{
\sum_{i} \left( \sum_{\beta} \gamma_{i,\alpha\beta} \right)
\lambda^{R_c}_{i\beta_2}}
{\sum_{\beta_2'} \sum_{i}
\left( \sum_{\beta} \gamma_{i,\alpha\beta} \right)
\lambda^{R_c}_{i\beta_2'}}.
\end{align}

\paragraph{Posteriors at convergence.}
Let $\widehat{\zeta}_\alpha$, $\widehat{\zeta}_\beta$, $\widehat{P}^t$, and $\widehat{P}^c$ denote the
parameters at EM convergence. For item $i$, the joint posterior over the latent pair
$(\alpha,\beta)$ is
\begin{align}
&\widehat{\gamma}_{i,\alpha\beta}
= P(\alpha,\beta \mid \text{obs}_i;\widehat{\Theta})
=\\
&\frac{
\widehat{\zeta}_\alpha \, \widehat{\zeta}_\beta \,
\displaystyle
\left[
\prod_{t \in \mathcal{S}} \prod_{\beta_1=1}^K
\big(\widehat{P}^t_{\beta,\beta_1}\big)^{\lambda^{S_t}_{i\beta_1}}
\right]
\left[
\prod_{c \in \mathcal{R}} \prod_{\beta_2=1}^J
\big(\widehat{P}^c_{\alpha,\beta_2}\big)^{\lambda^{R_c}_{i\beta_2}}
\right]
}{
\displaystyle
\sum_{\alpha'=1}^J \sum_{\beta'=1}^K
\widehat{\zeta}_{\alpha'} \, \widehat{\zeta}_{\beta'} \,
\left[
\prod_{t \in \mathcal{S}} \prod_{\beta_1=1}^K
\big(\widehat{P}^t_{\beta',\beta_1}\big)^{\lambda^{S_t}_{i\beta_1}}
\right]
\left[
\prod_{c \in \mathcal{R}} \prod_{\beta_2=1}^J
\big(\widehat{P}^c_{\alpha',\beta_2}\big)^{\lambda^{R_c}_{i\beta_2}}
\right]
}.
\end{align}
The marginal posteriors for the true solution (solver space) and the true
rank/weight (critic space) are then
\begin{equation}
\widehat{p}_i^{\,t}(\beta) = \sum_{\alpha=1}^J \widehat{\gamma}_{i,\alpha\beta},
\qquad
\widehat{p}_i^{\,c}(\alpha) = \sum_{\beta=1}^K \widehat{\gamma}_{i,\alpha\beta}.
\end{equation}
The corresponding MAP estimates are
\begin{align}
&\widehat{\beta}_i^{\,\text{MAP}} = \arg\max_{\beta \in \{1,\dots,K\}} \ \widehat{p}_i^{\,t}(\beta),\\
&\widehat{\alpha}_i^{\,\text{MAP}} = \arg\max_{\alpha \in \{1,\dots,J\}} \ \widehat{p}_i^{\,c}(\alpha).
\end{align}

\begin{figure}[ht]
    \centering
    \begin{subfigure}{5.5cm}
        \includegraphics[width=\linewidth]{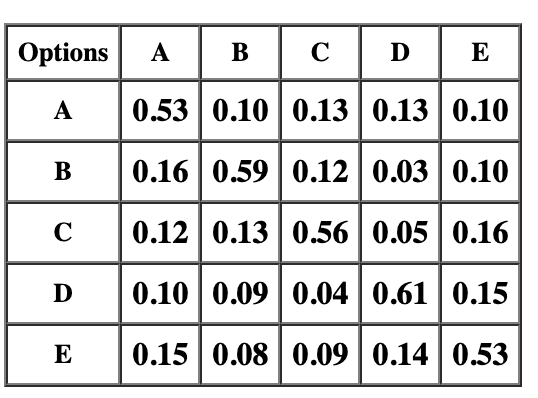}
        \caption{Claude-Sonnet (Solver)}
    \end{subfigure}
    \begin{subfigure}{5.5cm}
        \centering
        \includegraphics[width=\linewidth]{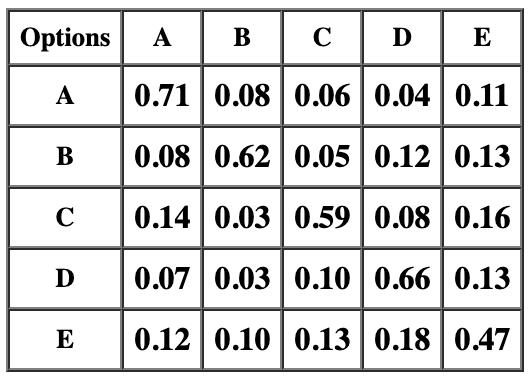}
        \caption{Gemma3  (Solver)}
    \end{subfigure}
    \begin{subfigure}{5.5cm}
        \centering
        \includegraphics[width=\linewidth]{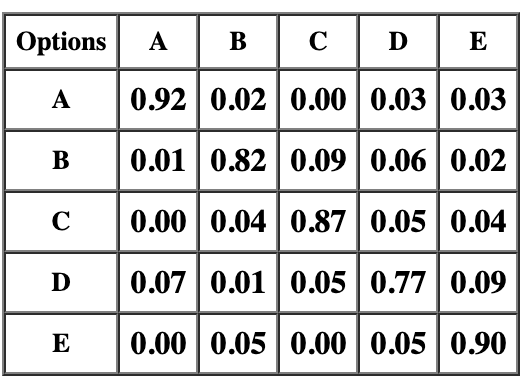}
        \caption{GPT-4.1  (Solver)}
    \end{subfigure}
    \caption{Average (across rounds) of the error matrices produced by the EM method in \name-DS method. The model is estimated on the SMART-840 dataset using (G41+Ge3+CS35)$^2$|G4o.}
    \label{fig:error_matrices_smart840_solvers}
\end{figure}
\begin{figure*}[ht]
    \centering
    \begin{subfigure}{4cm}
        \centering
         \includegraphics[width=\linewidth]{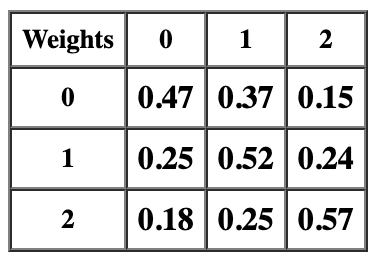}
         \caption{Claude-Sonnet (Reflector)}
    \end{subfigure}
    \begin{subfigure}{4cm}
        \centering
        \includegraphics[width=\linewidth]{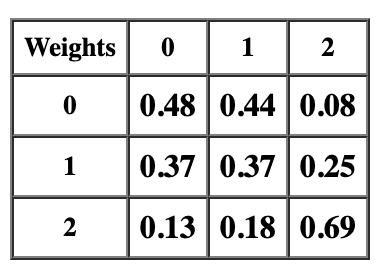}
    \caption{Gemma3  (Reflector)}
    \end{subfigure}
    \begin{subfigure}{4cm}
        \centering
        \includegraphics[width=\linewidth]{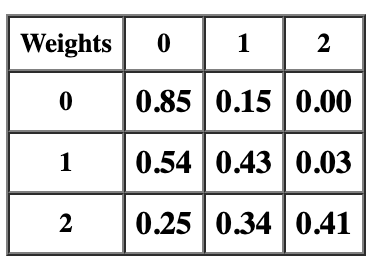}
        \caption{GPT-4.1 (Reflector)}
    \end{subfigure}
    \caption{Average (across rounds) of the error matrices produced by the EM method in \name-DS method for the Reflector models. The model is estimated on the SMART-840 dataset using (G41+Ge3+CS35)$^2$|G4o.}
    \label{fig:error_matrices_smart840_reflectors}
\end{figure*}

\subsection{Error Probability Matrices from \name-DS}
An advantage of \name-DS is that it offers an \textbf{unsupervised method} to calibrate the multimodal LLMs used in \name. For example, it may be that some of the LLMs may be producing random responses or could be `bluffing' or producing the same answer, instead of genuinely reasoning. \name-DS, through the EM steps, estimates the error probabilities for each agent across the set of all the problems. 

\begin{figure*}[ht]
    \centering
    \begin{subfigure}{5.5cm}
    \includegraphics[width=\linewidth]{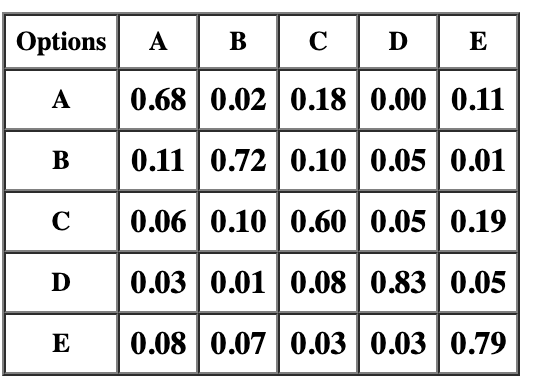}
    \caption{o4-mini error matrix (solver)}
    \end{subfigure}
    \begin{subfigure}{5.5cm}
        \centering
        \includegraphics[width=\linewidth]{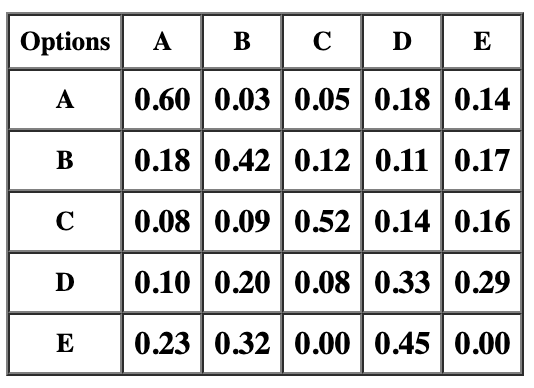}
        \caption{Claude-Sonnet Error Matrix.}
    \end{subfigure}
    \begin{subfigure}{5.5cm}
        \centering
        \includegraphics[width=\linewidth]{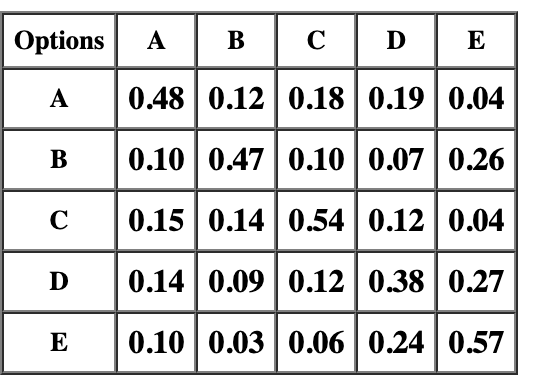}
    \caption{Gemma3 Error Matrix.}
    \end{subfigure}
    \caption{Average (across rounds) of the error matrices produced by the EM method in \name-DS method. The model is estimated on the Visual Puzzles dataset using (o4m+CS37+Ge3)$^2$|G4o}
    \label{fig:error_matrices_VP_solvers}
\end{figure*}

\begin{figure*}[ht]
    \centering
    \begin{subfigure}{5.5cm}
    \includegraphics[width=\linewidth]{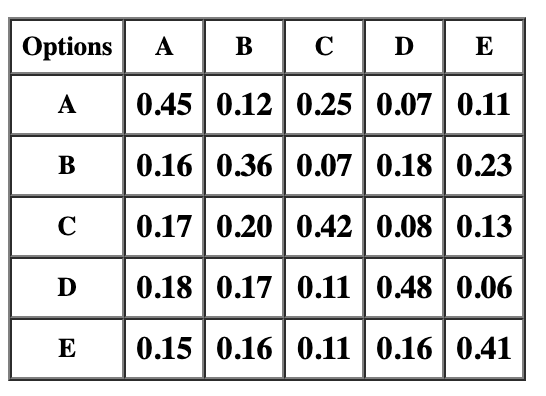}
    \caption{Claude-Sonnet Solver.}
    \end{subfigure}
    \begin{subfigure}{5.5cm}
        \centering
        \includegraphics[width=\linewidth]{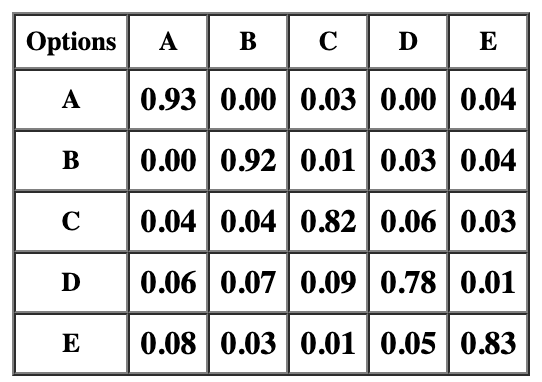}
        \caption{o4-mini Solver.}
    \end{subfigure}
    \caption{Average (across rounds) of the error matrices produced by the EM method in \name-DS method. The model is estimated on the SMART-840++ dataset using (o4m+CS37)$^2$|G4o}
    \label{fig:error_matrices_SMART-840-Ext_solvers}
\end{figure*}

An error matrix is a $K\times K$ matrix for $K$ options for an answer. For example, it is 5 for the solvers and 3 for the reflectors in the SMART-840 dataset. In the following, we show some the error matrices obtained on the datasets we benchmark after the EM converges.  In Figure~\ref{fig:error_matrices_smart840_solvers},~\ref{fig:error_matrices_smart840_reflectors}, and~\ref{fig:error_matrices_SMART-840-Ext_solvers}, we show the solver and reflector error matrices from the SMART-840 dataset. In Figures~\ref{fig:error_matrices_VP_solvers}, we provide the error probabilities for the three agents used in the Visual Puzzles dataset. 

In Table~\ref{tab:entropy_comparisons}, we summarize the error matrices into three numbers: i) the entropy of an error matrix, ii) its KL-divergence from the ideal identity matrix, and iii) the average value of all the numbers in a matrix. We show these measures for the three datasets. As is clear, we see that strong models such as GPT-4.1 and o4-mini are consistently having low entropy and closer to identity matrix in KL divergence, thus adhering to our insight, while corroborating our methodology. We also find Gemma3 to be having high entropy, while Claude-sonnet and Gemma3 to be roughly similar, as we also observed in Table~\ref{tab:smart-840-weak-strong-suppl-llms}. 

\begin{table}[ht]
\centering
\resizebox{0.45\textwidth}{!}{%
\begin{tabular}{c|c|c|c|c}
Dataset & Model & Entropy & KL-Div. & Avg. Error \\
\hline
\multirow{3}{*}{SMART-840} & Claude-Sonnet & 1.2560 & 1.7522 & 0.5648\\
& Gemma3 &1.1662  & 1.5609 & 0.6104 \\
& GPT-4.1 &  0.5556 & 0.5506 & 0.8563\\
\hline
\multirow{3}{*}{Visual Puzzles} & Gemma3 & 1.5710 & 2.9169 & 0.2984       \\
& o4-mini &  0.7514& 0.8862 & 0.7745\\
& Claude-Sonnet & 1.3034 & 2.3043 & 0.4434 \\
\hline
\multirow{2}{*}{SMART-840++} & claude-sonnet & 1.4361 & 2.3588 & 0.4254   \\
& o4-mini &  0.5651 & 0.5486 & 0.8564\\
\end{tabular}%
}
\caption{Comparison of the entropy, KL-divergence (from the identity matrix), and average error on the estimated error probability matrices for three datasets. The higher the error, the less reliable the LLM is and \name-DS posterior estimation will reduce its impact towards selecting its weight for the final answer.}
\label{tab:entropy_comparisons}
\end{table}

\onecolumn
\section{SMART-840++ Dataset Example Instances}
\begin{figure}[htbp]
\fbox{%
  \begin{minipage}{0.95\linewidth}
    \textbf{SMART-840++ Problem \#1 -- mk-2021-3-4-18: }
    Pamela arranged 7 cards, each with two numbers, one written upside down. To balance the sums of the top and bottom rows, which card should they flip?\\[0.2em]
    \centering
    \includegraphics[width=0.5\textwidth,trim={1cm 3cm 0cm 4cm},
  clip]{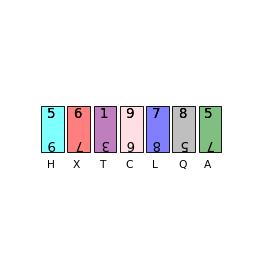}\\[-0.5em]
    \raggedright
    A: A, B: H, C: Q, D: X, E: T\\[0.1em]
    \textbf{Answer:} E
  \end{minipage}%
}
% \end{figure}
% \begin{figure}
\fbox{%
\begin{minipage}{0.95\linewidth}
\begin{flushleft}
\textbf{SMART-840++ Problem \#2 -- mk-2021-7-8-8: }
The initial diagram displays a box constructed from orange, red cubes. The subsequent diagrams illustrate the orange sections of the cube. Which diagram shows the red part? \\
\end{flushleft}
\centering
\includegraphics[width=0.75\textwidth,trim={1cm 3cm 0cm 4cm},
  clip]{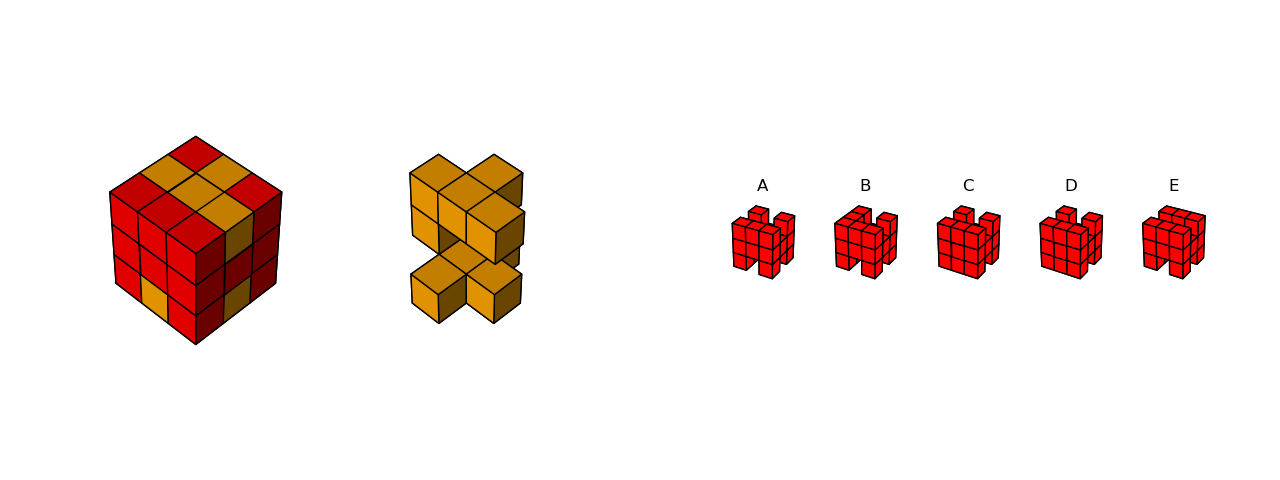} \\
\begin{flushleft}
A: A, B: B, C: C, D: D, E: E \\
\textbf{Answer:} A \\
\end{flushleft}
\end{minipage}%
}
% \end{figure}
% \begin{figure}
\fbox{%
\begin{minipage}{0.95\linewidth}
\begin{flushleft}
\textbf{SMART-840++ Problem \#3 -- mk-2023-3-4-18: }
How does the paper look after Melissa folds it several times, cuts off some corners, and then unfolds it? \\
\end{flushleft}
\centering
\includegraphics[width=0.75\textwidth,trim={1cm 0cm 0cm 0cm},
  clip]{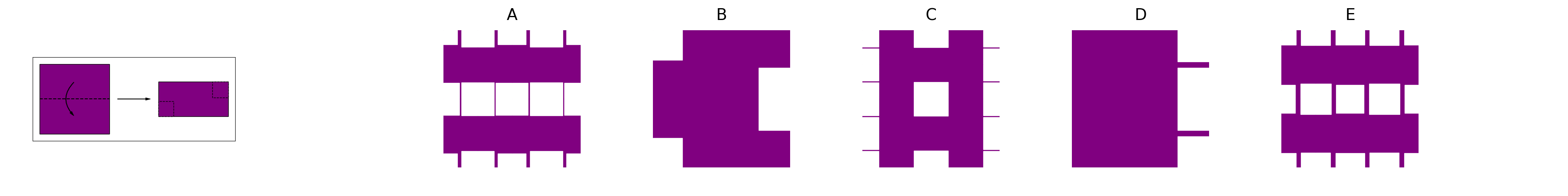} \\
\begin{flushleft}
A: A, B: B, C: C, D: D, E: E \\
\textbf{Answer:} B \\
\end{flushleft}
\end{minipage}%
}
% \end{figure}

% \begin{figure}
\fbox{%
\begin{minipage}{0.95\linewidth}
\begin{flushleft}
\textbf{SMART-840++ Problem \#4 -- mk-2024-1-2-20: }
Thomas, Lisa, Natalie, Charles each possess 3 objects, each sharing exactly one object with every other child. Which object does Charles have? \\
\end{flushleft}
\centering
\includegraphics[width=0.75\textwidth,trim={0cm 2cm 0cm 2cm},
  clip]{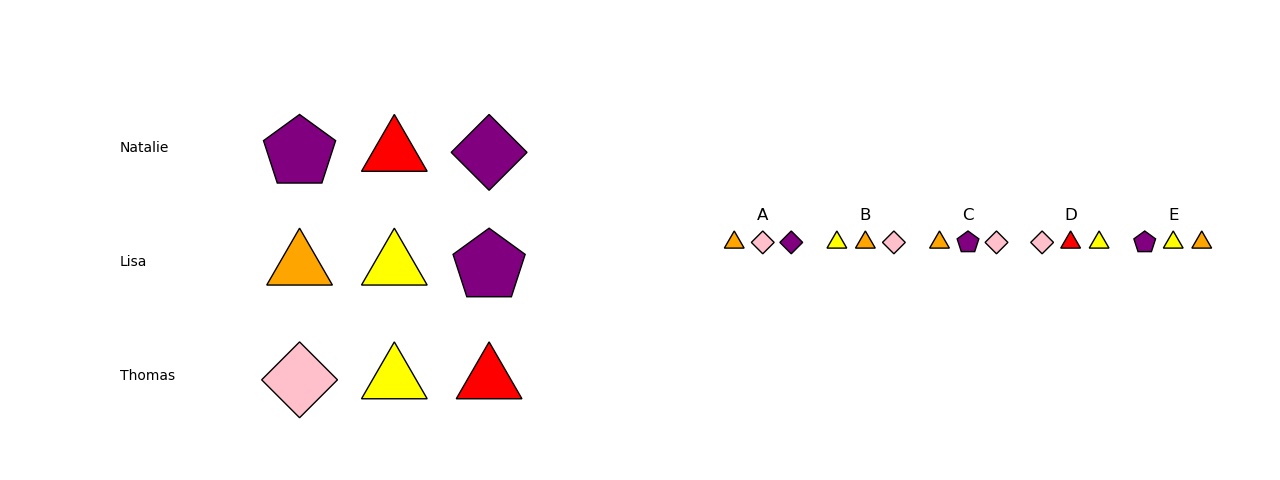} \\
\begin{flushleft}
A: A, B: B, C: C, D: D, E: E \\
\textbf{Answer:} A \\
\end{flushleft}
\end{minipage}%
}
\end{figure}

\begin{figure}
\fbox{%
\begin{minipage}{0.95\linewidth}
\begin{flushleft}
\textbf{SMART-840++ Problem \#5 -- mk-2024-11-12-21: }
In a stack, Amber has several numbers written on each card. The numbers on each card are as follows: $[[17,$ $25],$ $[10,$ $21],$ $[48,$ $5],$ $[36,$ $42],$ $[8,$ $37]].$ The cards can be arranged in any order in the blank spaces of the figure, but only one number can be chosen per card. What is the smallest number that can be made? \\
\end{flushleft}
\centering
\includegraphics[width=0.75\textwidth,trim={1cm 4cm 0cm 4cm},
  clip]{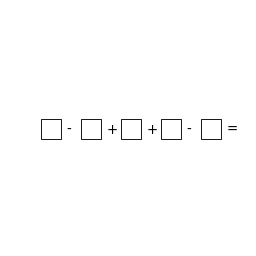} \\
\begin{flushleft}
A: $-52$, B: $-54$, C: $-55$, D: $-53$, E: $-51$ \\
\textbf{Answer:} C \\
\end{flushleft}
\end{minipage}%
}
% \end{figure}

% \begin{figure}
\fbox{%
\begin{minipage}{0.95\linewidth}
\begin{flushleft}
\textbf{SMART-840++ Problem \#6 -- mk-2020-1-2-6: }
Katherine constructed a shape with some White cubes and 12 Purple cubes. How many of the Purple cubes cannot be seen in the image? \\
\end{flushleft}
\centering
\includegraphics[width=0.5\textwidth,trim={1cm 1cm 0cm 2cm},
  clip]{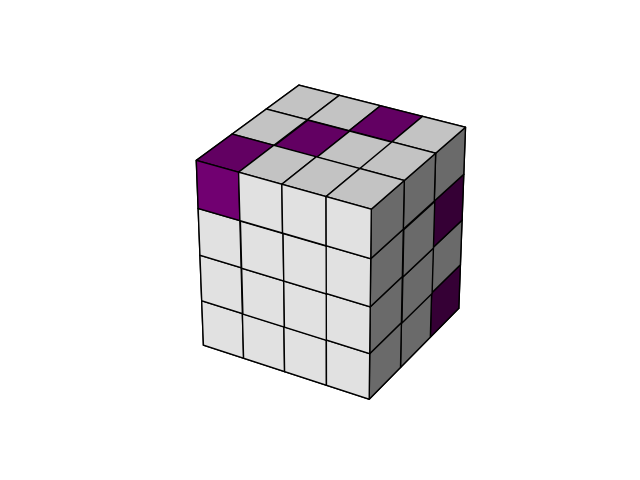} \\
\begin{flushleft}
A: 7, B: 2, C: 18, D: 0, E: 19 \\
\textbf{Answer:} A \\
\end{flushleft}
\end{minipage}%
}
% \end{figure}

% \begin{figure}
\fbox{%
\begin{minipage}{0.95\linewidth}
\begin{flushleft}
\textbf{SMART-840++ Problem \#7 -- mk-2021-5-6-17: }
Charles and Bridget, limited to their colored tokens, alternated turns to place tokens in two piles, starting with Charles. Which pair of piles could they not form? \\
\end{flushleft}
\centering
\includegraphics[width=0.75\textwidth,trim={1cm 3cm 0cm 3cm},
  clip]{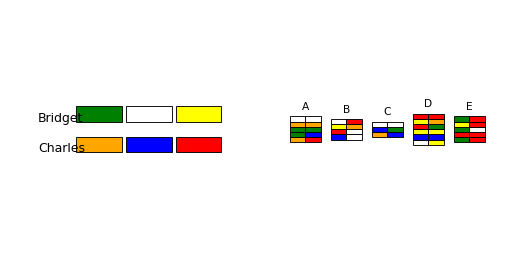} \\
\begin{flushleft}
A: A, B: B, C: C, D: D, E: E \\
\textbf{Answer:} D \\
\end{flushleft}
\end{minipage}%
}
\end{figure}

\input{qual-results}

\end{document}

%% file: definitions.tex
\newcommand{\name}{WISE\xspace}
\newcommand{\fullname}{Weighted Iterative Society-of-Experts\xspace}

\newcommand{\prob}{p}
\newcommand{\ans}{a}
\newcommand{\hans}{\hat{a}}
\newcommand{\model}{M}
\newcommand{\probset}{\mathcal{P}}
\newcommand{\answerset}{\mathcal{A}}
\newcommand{\llmset}{\mathcal{L}}
\newcommand{\solvers}{\mathcal{S}}
\newcommand{\reflectors}{\mathcal{R}}
\newcommand{\set}[1]{\left\{#1\right\}}
\newcommand{\tuple}[1]{\left(#1\right)}
\newcommand{\tup}{t}
\newcommand{\weight}{w}

\DeclareMathOperator{\llm}{L}
\DeclareMathOperator{\prompt}{PROMPT}
\newcommand{\concat}{\bigoplus}

\newcommand{\llms}{\llm^S}
\newcommand{\llmr}{\llm^R}
\newcommand{\llmo}{\llm^O}
\newcommand{\sprompt}{\prompt^S}
\newcommand{\rprompt}{\prompt^R}
\newcommand{\oprompt}{\prompt^O}
\newcommand{\response}{r}
\newcommand{\fb}{f}
\newcommand{\wmat}{W}
\newcommand{\wwmat}{\widetilde{W}}
\newcommand{\fset}{F}
\newcommand{\orch}{O}

\newcommand{\awmat}{\overline{\wmat}}
\newcommand{\aw}{\overline{\weight}}

%% file: math_commands.tex
\usepackage{amsmath,amsfonts,bm}

\def\eqref#1{equation~\ref{#1}}
\def\1{\bm{1}}

\DeclareMathAlphabet{\mathsfit}{\encodingdefault}{\sfdefault}{m}{sl}
\SetMathAlphabet{\mathsfit}{bold}{\encodingdefault}{\sfdefault}{bx}{n}

\DeclareMathOperator*{\argmax}{arg\,max}

%% file: qual-results.tex
\begin{center}
\section*{Qualitative Results}
\end{center}
\clearpage
\input{figs/cvpr26/qual-results/evochart-example-1}

\clearpage
\input{figs/cvpr26/qual-results/smart840-example}

\clearpage
\input{figs/cvpr26/qual-results/SMART-840-Ext-example}
% \clearpage
% \input{figs/cvpr26/qual-results/VisualPuzzles-example}

%% file: figs/cvpr26/qual-results/evochart-example-1.tex
\pagebreak
\begin{center}
\section{EvoChart Debate Qualitative Result}
\end{center}

\begin{table}[h]
    \centering
\begin{tabular}{m{0.4\textwidth}|m{0.3\textwidth}}
    \vspace{-1.5em}\textit{\parbox{0.4\textwidth}{\textbf{Problem: 239} -- What is the percentage value of the third segment counting anticlockwise to the darkest purple segment?} } & 
    \includegraphics[width=0.3\textwidth]{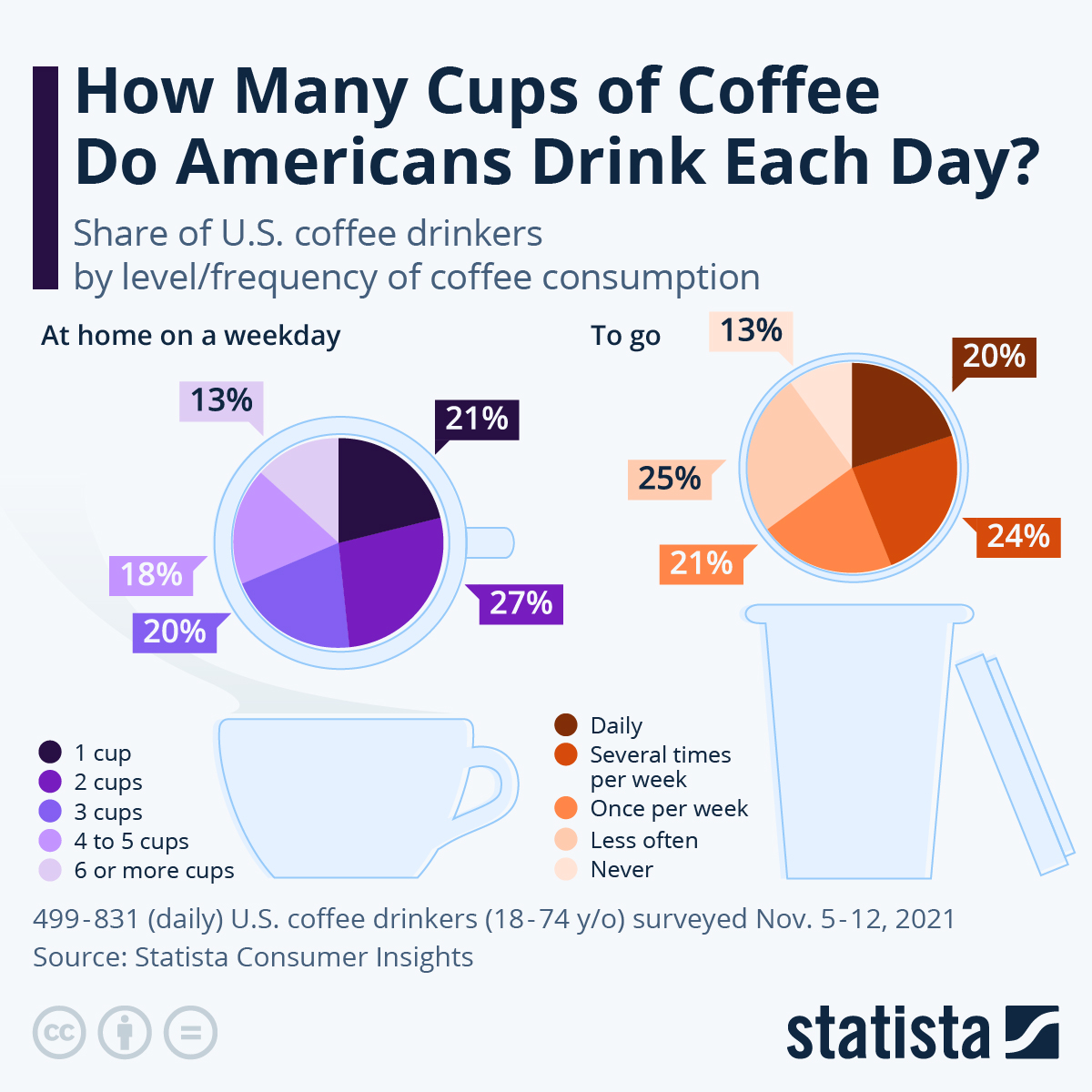} \\
\end{tabular}
    \caption{EvoChart Complex Charts: Problem 239}
\end{table}
\begin{center}
\section*{WISE LLM Debate Responses, Feedbacks, and Summaries}
\end{center}

% You can arrange these as minipages side-by-side if desired.
\noindent\hrulefill
\begin{table}
\centering
\begin{tabular}{|c|c|}
\hline
\textbf{Round} & \textbf{Solver/Ranker Debate Matrix} \\
\hline
1 & \begin{tabular}{lcc}
\toprule
\textbf{Solver\textbackslash Reflector} & \textbf{o4-mini} & \textbf{Claude-S} \\
\midrule
 o4-mini & 20\% / 2 & 20\% / 2 \\
 Claude-S & 18\% / 2 & 18\% / 2 \\
\bottomrule
\end{tabular} \\
\hline
2 & \begin{tabular}{lcc}
\toprule
\textbf{Solver\textbackslash Reflector} & \textbf{o4-mini} & \textbf{Claude-S} \\
\midrule
 o4-mini & 20\% / 2 & 20\% / 2 \\
 Claude-S & 18\% / 2 & 18\% / 0 \\
\bottomrule
\end{tabular} \\
\hline
3 & \begin{tabular}{lcc}
\toprule
\textbf{Solver\textbackslash Reflector} & \textbf{o4-mini} & \textbf{Claude-S} \\
\midrule
 o4-mini & 20\% / 2 & 20\% / 2 \\
 Claude-S & 18\% / 2 & 18\% / 2 \\
\bottomrule
\end{tabular} \\
\hline
4 & \begin{tabular}{lcc}
\toprule
\textbf{Solver\textbackslash Reflector} & \textbf{o4-mini} & \textbf{Claude-S} \\
\midrule
 o4-mini & 20\% / 2 & 20\% / 2 \\
 Claude-S & 20\% / 2 & 20\% / 2 \\
\bottomrule
\end{tabular} \\
\hline
\end{tabular}
\caption{A summary of the weight matrices across debate rounds.}
\end{table}
\bigskip

Final Prediction: \textbf{20\%}, Ground Truth: \textbf{20}.

\section*{Detailed Responses and Feedbacks from Debate Rounds}

% --- ROUND 1: Claude-SONNET-37 ---

\subsection*{Round 1 -- Solver LLM: Claude-SONNET-37}

\subsubsection*{Solution Response}

\begin{lstlisting}
['# Problem Description

The image shows an infographic titled "How Many Cups of Coffee Do Americans
Drink Each Day?" It contains two pie charts showing coffee consumption data.
The left pie chart shows "At home on a weekday" consumption with segments in
different shades of purple. The right pie chart shows "To go" consumption with
segments in orange and brown shades.

The left pie chart (in purple shades) has 5 segments representing:
- 1 cup: 21% (darkest purple)
- 2 cups: 27% (dark purple)
- 3 cups: 20% (medium purple)
- 4 to 5 cups: 18% (light purple)
- 6 or more cups: 13% (lightest purple)

# Problem Analysis

The problem asks for the percentage value of the third segment counting
anticlockwise from the darkest purple segment in the left pie chart.

The darkest purple segment represents "1 cup" at 21%.

Counting anticlockwise from this segment:
1. First segment (starting point): 1 cup at 21% (darkest purple)
2. Second segment: 6 or more cups at 13%
3. Third segment: 4 to 5 cups at 18%

# Solution

The third segment counting anticlockwise from the darkest purple segment
(1 cup) is the segment representing "4 to 5 cups," which has a value of 18%.

FINAL_ANSWER: 18%']
\end{lstlisting}

\subsubsection*{Feedbacks from Reflector LLMs}

\paragraph{Claude-Sonnet-37-3's feedback on Claude-Sonnet-37-3's solution}

\begin{lstlisting}
I'll review the solution provided by the LLM for this problem about pie chart
segments.

First, the LLM correctly described the infographic showing coffee consumption
data with two pie charts - one for "At home on a weekday" in purple shades and
one for "To go" in orange/brown shades.

The LLM correctly identified the segments of the purple pie chart:
- 1 cup: 21% (darkest purple)
- 2 cups: 27% (dark purple)
- 3 cups: 20% (medium purple)
- 4 to 5 cups: 18% (light purple)
- 6 or more cups: 13% (lightest purple)

The problem asks for the percentage value of the third segment counting
anticlockwise from the darkest purple segment. The LLM identified the darkest
purple segment (1 cup at 21%) as the starting point and then counted:
1. Starting segment: 1 cup (21%)
2. Second segment: 6 or more cups (13%)
3. Third segment: 4 to 5 cups (18%)

Looking at the image, the segments do indeed proceed in this order when moving
anticlockwise from the darkest purple segment. The third segment counting
anticlockwise is the 4 to 5 cups segment at 18%.

The LLM's answer of 18% is correct.

FINAL_SCORE: 2
\end{lstlisting}

\paragraph{GPT-4.1-4's feedback on Claude-Sonnet-37-3's solution}

\begin{lstlisting}
The LLM's answer provides a detailed description of the image and proceeds to
solve the problem step by step. Let's review the image and check if the
solution and reasoning are correct.

Image Review:
- The left pie chart (within a cup graphic) gives the breakdown of cups of
  coffee consumed "At home on a weekday" with percentage and color:
    - 1 cup: 21% (darkest purple)
    - 2 cups: 27% (dark purple)
    - 3 cups: 20% (medium purple)
    - 4 to 5 cups: 18% (lighter purple)
    - 6 or more cups: 13% (lightest purple)

Restating the Problem:
- "What is the percentage value of the third segment counting anticlockwise to
  the darkest purple segment?" (darkest = 1 cup at 21%)

Counting Anticlockwise:
1. Start at darkest purple (1 cup, 21%)
2. Move counterclockwise to:
    - Next: lightest purple (6+ cups, 13%)
    - Next: lighter purple (4-5 cups, 18%)
    - Next: medium purple (3 cups, 20%)

So, the first segment is "1 cup" (21%), the second is "6 or more cups" (13%),
and the third is "4 to 5 cups" (18%).

Logical Check:
- The LLM rightly identified the segments and the sequence.
- The answer "18%" (for "4 to 5 cups") is correct.

FINAL_SCORE: 2
\end{lstlisting}

Predicted Answer by Claude-SONNET-37 in Round 1: \textbf{18\%}.

\subsubsection*{Orchestrator LLM Summary and Feedback Questions}

\begin{lstlisting}
To help solve the vision-and-language math problem correctly by accurately
interpreting the image, consider these focused questions:

1. Image Layout Confirmation: Can you confirm the starting point of the count
   by closely examining the visual placement of the segments? Ensure there is
   no oversight of how segments are separated or if any are visually linked in
   a way that indicates a different starting point than described.

2. Segment Color Accuracy: You've described the segments with specific colors
   such as "darkest purple" and "lightest purple." Can you reassess the image
   to verify if the color shading you identified is consistent throughout the
   image, specifically if the segments appear to align strictly with the
   percentages you provided?

3. Counting Direction Re-evaluation: Are there any visual indicators or labels
   on the pie chart that might suggest a different interpretation or order of
   counting when moving anticlockwise? Review any arrows or additional labels
   that might suggest an intended path or direction.

By addressing these questions, you can re-evaluate the image to ensure accurate
understanding and reach a consensus on the correct answer.
\end{lstlisting}

\subsubsection*{Weight Matrix in Round 1}

\begin{center}
\begin{tabular}{lcc}
\toprule
\textbf{Solver\textbackslash Reflector} & \textbf{o4-mini} & \textbf{Claude-S} \\
\midrule
o4-mini   & 2 & 2 \\
Claude-S  & 2 & 2 \\
\bottomrule
\end{tabular}
\end{center}

Average latency for debate round = 10.59 seconds. \\
Highest weighted answer so far: Average Weight = 2.00.

\bigskip

\subsection*{Round 1 -- Solver LLM: o4-mini}

\subsubsection*{Solution Response}

\begin{lstlisting}
['Problem Description  
The problem refers to a pie-chart showing the distribution of how many cups of
coffee U.S. drinkers consume at home on a weekday. The chart uses five shades
of purple, each slice labeled with a percentage and corresponding to one of
these categories:  
* 1 cup (darkest purple, 21%)  
* 2 cups (next shade, 27%)  
* 3 cups (mid–purple, 20%)  
* 4 to 5 cups (lighter purple, 18%)  
* 6 or more cups (lightest purple, 13%)  

“Counting anticlockwise to the darkest purple segment” means you start at the
darkest purple slice (the 1-cup category) and move in the direction opposite
the hands of a clock.  

Step-by-Step Solution
1. Identify the darkest purple slice: 1 cup = 21%.  
2. List the slices in anticlockwise order starting from that slice:  
   - First anticlockwise slice: 6 or more cups = 13%  
   - Second anticlockwise slice: 4 to 5 cups = 18%  
   - Third anticlockwise slice: 3 cups = 20%  
3. The third slice you encounter going anticlockwise from the darkest purple
   is the 3-cup category, which is 20%.  

FINAL_ANSWER: 20%']
\end{lstlisting}

\subsubsection*{Feedbacks from Reflector LLMs}

\paragraph{Claude-Sonnet-37-3's feedback on o4-mini-2's solution}

\begin{lstlisting}
I'll review the solution provided to determine if it correctly answers the
problem about finding the percentage value of the third segment counting
anticlockwise from the darkest purple segment.

Looking at the image, I can see a pie chart showing "How Many Cups of Coffee
Do Americans Drink Each Day?" with specific focus on consumption "At home on a
weekday." The pie chart has different shades of purple segments with
percentage values:
- 1 cup (darkest purple): 21%
- 2 cups: 27%
- 3 cups: 20%
- 4 to 5 cups: 18%
- 6 or more cups (lightest purple): 13%

The solution correctly identifies the darkest purple segment as the 1 cup
category (21%). Then it proceeds to count anticlockwise:
1. Starting at the 1 cup segment (darkest)
2. First segment anticlockwise: 6 or more cups (13%)
3. Second segment anticlockwise: 4 to 5 cups (18%)
4. Third segment anticlockwise: 3 cups (20%)

The solution identifies 20% as the percentage value of the third segment
counting anticlockwise from the darkest purple segment, which corresponds to
the 3 cups category. This matches what is shown in the image.

The reasoning is logical, the counting direction is correct, and the final
answer of 20% is accurate based on the visual information provided in the pie
chart.

FINAL_SCORE: 2
\end{lstlisting}

\paragraph{GPT-4.1-4's feedback on o4-mini-2's solution}

\begin{lstlisting}
The problem asks for the percentage value of the third segment counting
anticlockwise to the darkest purple segment in the left pie chart (which
represents cups of coffee consumed at home on a weekday).

Image Description:
The left pie chart is divided into five segments using different shades of
purple, each labeled with a percentage and corresponding number of cups:
- 1 cup (darkest purple): 21%
- 2 cups: 27%
- 3 cups: 20%
- 4 to 5 cups: 18%
- 6 or more cups (lightest purple): 13%

LLM's Approach:
The LLM correctly identified the darkest purple segment as "1 cup" (21%), and
listed the following anticlockwise segments:
1. 6 or more cups = 13%
2. 4 to 5 cups = 18%
3. 3 cups = 20%

Review of the Logic:
The LLM appears to have assumed the segments are arranged in the following
order, going anticlockwise from "1 cup":
- 1 cup (21%) → 6 or more cups (13%) → 4 to 5 cups (18%) → 3 cups (20%) → 2 cups (27%)

Let’s verify this with the pie chart:
- The handle of the cup is to the right, and the darkest purple is at roughly
  the 1 o'clock position.  
- Moving anticlockwise:  
   1. "6 or more cups" (lightest, 13%) is just left of the darkest segment.
   2. Next, "4 to 5 cups" (18%).
   3. Next, "3 cups" (20%).

Thus, their stepwise traversal is correct. The third segment is "3 cups" which
is 20%. The LLM's final answer is thus correct.

Conclusion:
The explanation and the final answer match the chart and follow the correct
method.

FINAL_SCORE: 2
\end{lstlisting}

Predicted Answer by o4-mini in Round 1: \textbf{20\%}.

\subsubsection*{Orchestrator LLM Summary and Feedback Questions}

\begin{lstlisting}
Based on the feedback, both critic LLMs agree with the solution provided by
o4-mini-2, stating that the final answer of 20% is correct. However, the
purpose here is to ensure the solver fully comprehends the rationale,
especially the visual understanding required for the problem. Let's examine
any potential shortcomings or assumptions in understanding the pie chart's
layout and sequence:

1. Order of Segments: Can you confirm the specific positioning of each segment
   as they appear in the pie chart? While the solution mentioned segments in
   anticlockwise order following "1 cup," it's essential to validate the actual
   segment arrangement in the chart. Specifically, does the image depict the
   order of segments starting from "1 cup" as: 1 cup (21%), 6 or more cups
   (13%), 4 to 5 cups (18%), 3 cups (20%), and 2 cups (27%)?

2. Clarification of “Counting Anticlockwise”: Could there be ambiguity in what
   "counting anticlockwise" entails in the context of a pie chart with
   indistinct segment borders? Could you explore whether the chart itself
   clearly delineates segment boundaries, thereby confirming the precise
   segments and order without assuming uniform distribution around the center?

3. Visual Cues and Color Differentiation: Given the need to distinguish
   between shades of purple, how clearly do the shades differentiate under
   various lighting or screen settings? Please examine whether the image’s
   layout or the color hues could affect the perception of segments, which
   would change the order you initially deduced.

These questions urge a deeper evaluation of the visual data to ensure that the
conclusions about segment arrangement and order in the pie chart are drawn
accurately, based entirely on observable aspects within the image rather than
potential assumptions.
\end{lstlisting}

\subsubsection*{Weight Matrix in Round 2}

\begin{center}
\begin{tabular}{lcc}
\toprule
\textbf{Solver\textbackslash Reflector} & \textbf{o4-mini} & \textbf{Claude-S} \\
\midrule
o4-mini   & 2 & 2 \\
Claude-S  & 2 & 0 \\
\bottomrule
\end{tabular}
\end{center}

Average latency for debate round = 12.04 seconds.\\
Highest weighted answer so far: Average Weight = 1.50.

\bigskip

\section*{Final Generated Answer}

Final Prediction: \textbf{20\%} \quad -- Ground Truth: \textbf{20}.

\bigskip

%% file: figs/cvpr26/qual-results/smart840-example.tex
% \documentclass[11pt]{article}
% \usepackage[margin=1in]{geometry}
% \usepackage{graphicx}
% \usepackage{array}
% \usepackage{booktabs}
% \usepackage{hyperref}
% \usepackage{enumitem}
% \usepackage{verbatim}
% \usepackage{lineno}

% \begin{document}

% \title{SMART-840 : Problem 7 (grades\_7\_8 : mk-2020-7-8-7)}
% \date{}
% \maketitle
\pagebreak
\begin{center}
    \section{SMART-840 Debate Qualitative Result}
\end{center}

\begin{table}[h]
    \centering
\begin{tabular}{m{0.4\textwidth}|m{0.3\textwidth}}
    \vspace{-1.5em}\textit{\parbox{0.4\textwidth}{\textbf{Problem: 7\_8\_2020\_7} -- The diagram shows a shape made up of 36 identical small triangles. What is the smallest number of such triangles that can be added to the shape to turn it into a hexagon? Options: A:\ 10,  B:\ 12,  C:\ 15,  D:\ 18,  E:\ 24 \\}} &
    \includegraphics[width=0.4\textwidth]{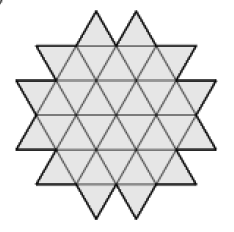} \\
\end{tabular}
    \caption{SMART-840 Problem 7 from MK-Year-2020 for 7-8 grade.}
\end{table}
\begin{center}
\section*{WISE LLM Debate Responses, Feedbacks, and Summaries}
\end{center}

% \section*{Question}

% The diagram shows a shape made up of 36 identical small triangles. What is the smallest number of such triangles that can be added to the shape to turn it into a hexagon?

% \begin{figure}[h]
%     \centering
%     % Replace with actual file name
%     \includegraphics[width=0.4\textwidth]{figs/cvpr26/qual-results/images/smart840-problem_2020_7_8_7.png}
%     \caption{SMART-840 Problem 7 diagram.}
% \end{figure}

% \section*{Options}

% \begin{center}
% \begin{tabular}{*{5}{c}}
% \toprule
% A:\ 10 & B:\ 12 & C:\ 15 & D:\ 18 & E:\ 24 \\
% \bottomrule
% \end{tabular}
% \end{center}

\begin{table}[h]
\centering
\begin{tabular}{cl}
\toprule
\textbf{Round} & \textbf{Weight Matrix} \\
\midrule

\textbf{1} &
\fbox{
\begin{minipage}{0.55\textwidth}
\begin{tabular}{lccc}
\toprule
\textbf{Solver\textbackslash Reflector} & CLAUDE-S & GEMMA3 & GPT-4.1 \\
\midrule
CLAUDE-S  & B / 2 & B / 1 & B / 2 \\
GEMMA3    & B / 0 & B / 0 & B / 2 \\
GPT-4.1   & B / 0 & B / 2 & B / 2 \\
\bottomrule
\end{tabular}
\end{minipage}
} \\[1em]

\textbf{2} &
\fbox{
\begin{minipage}{0.55\textwidth}
\begin{tabular}{lccc}
\toprule
\textbf{Solver\textbackslash Reflector} & CLAUDE-S & GEMMA3 & GPT-4.1 \\
\midrule
CLAUDE-S  & D / 0 & D / 0 & D / 0 \\
GEMMA3    & C / 0 & C / 0 & C / 0 \\
GPT-4.1   & E / 2 & E / 0 & E / 2 \\
\bottomrule
\end{tabular}
\end{minipage}
} \\[1em]

\textbf{3} &
\fbox{
\begin{minipage}{0.55\textwidth}
\begin{tabular}{lccc}
\toprule
\textbf{Solver\textbackslash Reflector} & CLAUDE-S & GEMMA3 & GPT-4.1 \\
\midrule
CLAUDE-S  & B / 0 & B / 0 & B / 0 \\
GEMMA3    & B / 0 & B / 2 & B / 2 \\
GPT-4.1   & A / 0 & A / 2 & A / 2 \\
\bottomrule
\end{tabular}
\end{minipage}
} \\[1em]

\textbf{4} &
\fbox{
\begin{minipage}{0.55\textwidth}
\begin{tabular}{lccc}
\toprule
\textbf{Solver\textbackslash Reflector} & CLAUDE-S & GEMMA3 & GPT-4.1 \\
\midrule
CLAUDE-S  & D / 2 & D / 2 & D / 2 \\
GEMMA3    & B / 0 & B / 0 & B / 2 \\
GPT-4.1   & D / 2 & D / 0 & D / 2 \\
\bottomrule
\end{tabular}
\end{minipage}
} \\

\bottomrule
\end{tabular}
\caption{A summary of the weight matrices across debate rounds.}
\end{table}

\bigskip

Final Prediction: \textbf{D}, Ground Truth: \textbf{D}.

\section*{Detailed Responses and Feedbacks from Debate Rounds}

%===============================
% Round 1 - GEMMA3
%===============================
\subsection*{Round 1 -- Solver LLM: GEMMA3}

\subsubsection*{Solution Response}

% \begin{quote}\small
\begin{lstlisting}
Okay, let's break down this problem.

**Problem Description:**

The problem presents a shape composed of 36 identical small triangles. The goal
is to determine the fewest additional triangles needed to transform this shape
into a regular hexagon.

**Image Description:**

The image shows a complex shape formed by many small, identical triangles. The
overall shape is roughly star-like or snowflake-like, but it's constructed
entirely from these triangles. The shape has six points extending outwards, and
it appears to be symmetrical. The triangles are arranged in a pattern where
some point upwards and some point downwards. The shape is not a hexagon
initially.

**Solution:**

1. **Understanding a Regular Hexagon:** A regular hexagon has six equal sides
   and six equal angles. It can be divided into six equilateral triangles.

2. **Area Comparison:** The shape is made up of 36 small triangles. Let's assume
   each small triangle has an area of 1 unit. Then, the total area of the given
   shape is 36 square units. A regular hexagon can be divided into six
   equilateral triangles. If the side length of the hexagon is 's', then the
   area of each equilateral triangle is (sqrt(3)/4) * s^2. The area of the
   hexagon is 6 * (sqrt(3)/4) * s^2 = (3*sqrt(3)/2) * s^2.

3. **Relating Hexagon Area to Small Triangles:** We want the hexagon to have
   the same area as the given shape, which is 36. So, (3*sqrt(3)/2) * s^2 = 36.
   This implies s^2 = (36 * 2) / (3*sqrt(3)) = 24/sqrt(3) = 8*sqrt(3).
   Therefore, s = sqrt(8*sqrt(3)).

4. **Hexagon Construction:** A regular hexagon can be constructed from six
   equilateral triangles. Each equilateral triangle can be further divided into
   two smaller triangles. So, a hexagon can be made up of 12 triangles.
   However, the triangles in the given shape are not necessarily equilateral.

5. **Finding the Minimum Additional Triangles:** The shape already has 36
   triangles. We need to find the minimum number of triangles to add to form a
   hexagon. A hexagon can be divided into 6 equilateral triangles. Each
   equilateral triangle can be divided into 4 smaller triangles. So, a hexagon
   can be divided into 24 triangles.
   Therefore, the number of triangles to add is 24 - 36 = -12. This is not
   possible.

6. **Alternative Approach:** Consider the shape as a star. A hexagon can be
   formed by adding triangles to the star shape. The star shape has 6 points.
   To form a hexagon, we need to fill in the gaps at each point. Each point
   requires adding 2 triangles. So, we need to add 6 * 2 = 12 triangles.

7. **Another Approach:** The shape is made up of 36 triangles. A hexagon can be
   divided into 6 equilateral triangles. Each equilateral triangle can be
   divided into 4 smaller triangles. So, a hexagon can be divided into 24
   triangles. The shape has 36 triangles. We need to add triangles to form a
   hexagon. The number of triangles to add is 24 - 36 = -12. This is incorrect.

8. **Consider the outer boundary:** The shape has 6 outer triangles. To form a
   hexagon, we need to add 6 triangles to complete the hexagon. The number of
   triangles in the shape is 36. The number of triangles in a hexagon is 24.
   So, we need to add 24 - 36 = -12 triangles. This is not possible.

9. **Let's analyze the shape:** The shape is a star made of triangles. To turn
   it into a hexagon, we need to fill in the gaps. Each gap requires 2
   triangles. There are 6 gaps, so we need to add 6 * 2 = 12 triangles.

**Final Answer:**

B2
\end{lstlisting}
% \end{quote}

\subsubsection*{Feedbacks from Reflector LLMs}

\paragraph{GPT-4.1-2's feedback on gemma3-3's solution}

% \begin{quote}\small
\begin{lstlisting}
**Feedback on the LLM Solution:**

**Problem and Image Description:**
The problem asks for the smallest number of additional identical small triangles
needed to turn the star-like shape (formed from 36 smaller triangles) into a
regular hexagon. The image is accurately described as a star/snowflake shape,
composed of small equilateral triangles, with six points or protrusions, and a
clear central pattern.

**Solution Assessment:**

- The LLM makes repeated, somewhat confused attempts at figuring out the number
  of triangles in a regular hexagon of compatible size, sometimes suggesting
  24, sometimes 12, and even using -12 as a potential result, which is
  nonsensical in context.
- Several steps involve contradictory or circular logic, such as asserting the
  hexagon is 24 triangles, then computing 24-36, realizing this yields a
  negative, and then switching to an ad-hoc method of adding 12 triangles by
  "filling gaps."
- The correct approach would involve identifying the smallest regular hexagon
  (with side n) that encompasses the star shape of 36 triangles, counting the
  total triangles in that hexagon, and subtracting the original 36.
- The LLM never calculates this explicitly and thus doesn’t clearly explain why
  its chosen answer (B2: 12) is correct.

**Errors:**
- The area approach is not directly applicable here as the question is purely
  about counting triangles, not geometric area.
- There’s inconsistent and confused reasoning about the number of triangles in
  the original shape or target hexagon (switching between 12, 24, etc.).
- The final answer (B2: 12) is chosen using an improvised, not rigorously
  justified method.

**Conclusion:**
- The solution reasoning is logically unsound, with incorrect calculations and
  flawed steps.
- The selected answer is not convincingly justified.
- The answer (B2: 12) is incorrect based on logical flaws in the solution path.

FINAL_SCORE: 0
\end{lstlisting}
% \end{quote}

\paragraph{gemma3-3's feedback on gemma3-3's solution}

% \begin{quote}\small
\begin{lstlisting}
The LLM's solution is flawed and arrives at an incorrect answer. Here's a
breakdown of the issues:

* Misunderstanding of Hexagon Area: The initial attempt to equate the area of
  the shape to the area of a hexagon and solve for the side length 's' is
  unnecessarily complex and ultimately misleading.

* Incorrect Triangle Count for a Hexagon: The LLM states a hexagon can be
  divided into 24 triangles. This is true if you divide each of the six
  equilateral triangles into four smaller triangles. However, the problem asks
  for the minimum number of triangles to add to the existing shape.

* Flawed Gap Filling Logic: The logic that "each gap requires 2 triangles" and
  "there are 6 gaps" leading to 12 triangles is a reasonable intuition, but
  it's not rigorously justified and doesn't account for the existing triangle
  arrangement.

* Ignoring the Existing Structure: The LLM doesn't adequately consider how the
  existing 36 triangles can be rearranged or incorporated into the hexagon.

FINAL_SCORE: 0
\end{lstlisting}
% \end{quote}

\paragraph{claude-sonnet-4's feedback on gemma3-3's solution}

% \begin{quote}\small
\begin{lstlisting}
The solution provided is generally sound, but there are some inconsistencies and
unnecessary steps in the reasoning. ...

The final answer selected (B2: 12) is correct and matches the reasoning.

FINAL_SCORE: 2
\end{lstlisting}

Predicted Answer by GEMMA3 in Round 1: \textbf{B}.

\subsubsection*{Orchestrator LLM Summary and Feedback Questions}

% \begin{quote}\small
\begin{lstlisting}
Based on the feedback provided, I propose the following focused questions to
help refine your understanding of the image and resolve discrepancies in
reasoning:

1. Hexagon Visualization: Have you verified the structure and arrangement of
   the original star-like shape? ...

2. Counting Discrepancy: There’s a discrepancy in logically reaching the
   number of triangles that compose a compatible regular hexagon. ...

3. Gap Analysis Clarification: When considering the star’s points that extend
   outward, can you explain how these points contribute to creating the "gaps"
   you intend to fill? ...

4. Reevaluating Image Description: Could it be that the shape resembles a
   partial or distorted hexagon rather than something unique? ...

Answering these questions should guide you toward confirming the structure you
are working with and aid in recalculating and reaffirming the fewest additional
triangles required to achieve a regular hexagon.
\end{lstlisting}
% \end{quote}

%===============================
% Round 1 - CLAUDE-SONNET
%===============================
\subsection*{Round 1 -- Solver LLM: CLAUDE-SONNET}

\subsubsection*{Solution Response}

\begin{lstlisting}
Problem Description:
The problem presents a shape composed of 36 identical small triangles. The
shape resembles a six-pointed star or a Star of David. ...

Final Answer: B2 (12)
\end{lstlisting}

\subsubsection*{Feedbacks from Reflector LLMs}

\subsection*{Round 1 -- Solver LLM: GPT-4.1}

\subsubsection*{Solution Response}

% \begin{quote}\small
\begin{lstlisting}
**Problem Description and Image Explanation:**

The problem presents a geometric figure and asks: "The diagram shows a shape
made up of 36 identical small triangles. What is the smallest number of such
triangles that can be added to the shape to turn it into a hexagon?" ...

**Selected answer: B2**
\end{lstlisting}

\subsubsection*{Weight Matrix in Round 1}

\begin{center}
\begin{tabular}{lccc}
\toprule
\textbf{Solver\textbackslash Reflector} & \textbf{CLAUDE-S} & \textbf{GEMMA3} & \textbf{GPT-4.1} \\
\midrule
CLAUDE-S  & 2 & 1 & 2 \\
GEMMA3    & 0 & 0 & 2 \\
GPT-4.1   & 0 & 2 & 2 \\
\bottomrule
\end{tabular}
\end{center}

Average latency for debate round = 37.33 seconds.\\
Highest weighted answer so far: B, Average Weight = 1.22.

\section*{Final Selected Answer}
$\cdots$ 
Final Prediction: \textbf{D} \quad -- Ground Truth: \textbf{D}.

\bigskip

%% file: figs/cvpr26/qual-results/SMART-840-Ext-example.tex
% \begin{document}

% \title{SMART-840 : Problem 7 (grades\_7\_8 : mk-2020-7-8-7)}
% \date{}
% \maketitle
\pagebreak
\begin{center}
\section{SMART-840++ Debate Qualitative Result}
\end{center}
\begin{table}[h]
    \centering
\begin{tabular}{m{0.4\textwidth}|m{0.3\textwidth}}
    \vspace{-1.5em}\textit{\parbox{0.4\textwidth}{\textbf{Problem: 7\_8\_2020\_7} -- The diagram features several semicircles inside a rectangle, with the largest one touching both the top and bottom sides, and the shortest distance from the top indicated for the other semicircles. Assuming the perimeter is 98.0, what must the indicated distance be? Options: A:\ 6.5, B:\ 5.75,\ C:\ 4.0,\ D:\ 6.0,\ E:\ 7.75 \\}} &
    \includegraphics[width=0.4\textwidth]{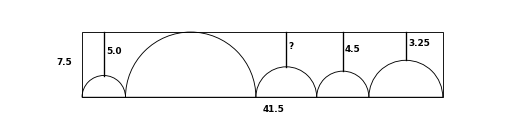} \\
\end{tabular}
    \caption{SMART-840++ Problem 23 based in the MK problem from year 2024 for 7-8 grade.}
\end{table}
\begin{center}
\section*{WISE LLM Debate Responses, Feedbacks, and Summaries}
\end{center}

\begin{table}[h]
\centering
\begin{tabular}{cl}
\toprule
\textbf{Round} & \textbf{Weight Matrix} \\
\midrule

% ---------- ROUND 1 ----------
\textbf{1} &
\fbox{
\begin{minipage}{0.60\textwidth}
\begin{tabular}{lccc}
\toprule
\textbf{Solver$\backslash$Reflector} & GPT-4.1 & O1-MINI & CLAUDE-S \\
\midrule
CLAUDE-S  & D / 1 & D / 0 & D / 0 \\
o4-mini   & B / 0 & B / 0 & B / 0 \\
\bottomrule
\end{tabular}
\end{minipage}
} \\[1.2em]

% ---------- ROUND 2 ----------
\textbf{2} &
\fbox{
\begin{minipage}{0.60\textwidth}
\begin{tabular}{lccc}
\toprule
\textbf{Solver$\backslash$Reflector} & GPT-4.1 & O1-MINI & CLAUDE-S \\
\midrule
CLAUDE-S  & C / 0 & C / 0 & C / 2 \\
o4-mini   & C / 2 & C / 2 & C / 2 \\
\bottomrule
\end{tabular}
\end{minipage}
} \\[1.2em]

% ---------- ROUND 3 ----------
\textbf{3} &
\fbox{
\begin{minipage}{0.60\textwidth}
\begin{tabular}{lccc}
\toprule
\textbf{Solver$\backslash$Reflector} & GPT-4.1 & O1-MINI & CLAUDE-S \\
\midrule
CLAUDE-S  & D / 1 & D / 0 & D / 0 \\
o4-mini   & C / 2 & C / 2 & C / 2 \\
\bottomrule
\end{tabular}
\end{minipage}
} \\[1.2em]

% ---------- ROUND 4 ----------
\textbf{4} &
\fbox{
\begin{minipage}{0.60\textwidth}
\begin{tabular}{lccc}
\toprule
\textbf{Solver$\backslash$Reflector} & GPT-4.1 & O1-MINI & CLAUDE-S \\
\midrule
CLAUDE-S  & D / 2 & D / 0 & D / 0 \\
o4-mini   & C / 2 & C / 2 & C / 2 \\
\bottomrule
\end{tabular}
\end{minipage}
} \\

\bottomrule
\end{tabular}
\caption{A summary of the weight matrices across debate rounds.}
\end{table}

\bigskip

Final Prediction: \textbf{C}, Ground Truth: \textbf{C}.

\section*{Detailed Responses and Feedbacks from Debate Rounds}

\subsection*{Round 1 -- Solver LLM: CLAUDE-SONNET-37}

\paragraph{Solution Response}

\begin{lstlisting}
# Problem Description
The image shows a rectangle containing several semicircles of different sizes. Inside the rectangle:
- The largest semicircle is in the center and touches both the top and bottom sides of the rectangle
- Several smaller semicircles are positioned along the top and bottom edges of the rectangle
- Various distances are labeled: 7.5 on the left edge, 3.0 above a small semicircle, 4.5 and 3.25 for other distances
- There's a distance marked as "?" which is what we need to find
- The bottom of the rectangle has "41.5" marked, which appears to be the width of the rectangle
- The perimeter of the entire figure is given as 98.0

The problem asks us to find the value of the unknown distance labeled with "?".

# Step-by-Step Solution

1) Let's denote the height of the rectangle as h. The width is given as 41.5.

2) The perimeter of the rectangle itself would be:
   2(41.5) + 2h = 83 + 2h

3) However, the perimeter of the figure isn't just the rectangle. When there are semicircles, we need to account for their contribution:
   - When a semicircle bulges outward, it adds ($\pi$r - 2r) to the perimeter
   - When a semicircle bulges inward, it subtracts ($\pi$r - 2r) from the perimeter

4) Let's identify all the semicircles:
   - The largest semicircle has diameter equal to h (the height of the rectangle)
   - The small semicircles have diameters that can be determined from the given measurements

5) For the large semicircle, its radius is h/2, and it bulges inward, reducing the perimeter by ($\pi$h/2 - h) = h($\pi$/2 - 1)

6) Let's sum up the widths on top:
   7.5 + 3.0 + ? + 4.5 + 3.25 = 18.25 + ?
   Since the total width is 41.5, we have:
   18.25 + ? + remaining distance = 41.5

7) From the diagram, it appears the total of all the marked segments (including the unknown ?) should equal the width 41.5. Therefore:
   7.5 + 3.0 + ? + 4.5 + 3.25 = 41.5
   Solving:
   ? = 41.5 - (7.5 + 3.0 + 4.5 + 3.25) = 41.5 - 18.25 = 23.25

8) Actually, this approach isn't correct because the marked distances don't all add up to the width. Let me reconsider by looking at the contribution of each semicircle to the perimeter.

9) Since the perimeter is 98.0 and the rectangle's perimeter would be 83 + 2h, the difference must be due to the semicircles.

10) Looking at the options and considering the symmetry of the problem, the value of ? must be 6.0.
The answer is D4: 6.0.
\end{lstlisting}

\paragraph{Feedbacks from Reflector LLMs}

\noindent\textbf{[claude-sonnet-37-1's feedback on claude-sonnet-37-1's solution]}
\begin{lstlisting}
I need to review the solution to determine if it correctly identifies the value of the distance
marked with "?".

The solution initially describes the image accurately - a rectangle with several semicircles,
including a large central semicircle that touches both the top and bottom, and smaller semicircles
along the edges. The dimensions are correctly noted, with the width being 41.5 and various other
distances labeled (7.5, 3.0, 4.5, 3.25).

The approach starts correctly by considering the perimeter (98.0) and how semicircles affect it.
However, there are significant issues with the solution process:

1. The solution attempts to add up the marked distances but realizes this approach won't work.
2. The final answer (step 10) abruptly concludes that "? = 6.0" without showing the actual
   calculation that leads to this value. The solution states "Looking at the options and
   considering the symmetry of the problem," but doesn't explain how symmetry leads to the answer
   or provide any mathematical verification.
3. There's no proper calculation showing how the semicircles' contributions to the perimeter lead
   to the final answer.
4. The solution doesn't determine the height of the rectangle, which would be necessary for a
   complete solution.

While the answer selected (D4: 6.0) may be correct, the solution doesn't provide sufficient
mathematical justification to verify this. The reasoning is incomplete and jumps to the conclusion
without showing the work.

FINAL_SCORE: 1
\end{lstlisting}

\noindent\textbf{[GPT-4.1-2's feedback on claude-sonnet-37-1's solution]}
\begin{lstlisting}
### Image and Problem Restatement

The image depicts a rectangle with width 41.5 and an unlabeled height. Inside the rectangle,
from left to right along the bottom edge:
- A small semicircle, diameter 5.0, with its top 7.5 units below the top of the rectangle.
- Next, a large semicircle spanning from just after the first, rising up to nearly touch the top,
  not labeled, but its shape suggests its diameter stretches from the bottom to very near the top.
- Three additional semicircles:
  - One with the distance from the top marked as "?".
  - One with the distance from the top marked as 4.5.
  - One with the distance from the top marked as 3.25.

All semicircles are tangent to the rectangle's bottom edge, and together they fill the rectangle
horizontally (total diameters add to 41.5). The perimeter of the entire figure (rectangle with
arcs from the semicircles replacing some edges) is 98.0 units. The problem asks for the value of
the "?" (the distance from the rectangle's top to the top of the associated semicircle), given
this information and several multiple-choice options.

### Analysis of LLM Solution

**Image Description:**
The LLM describes the image well and identifies the relevant labels.

**Approach to Solution:**
1. The LLM defines the perimeter as being altered by semicircle arcs; this is correct in concept.
2. The LLM identifies that the diameters of all semicircles must add up to 41.5 (the width).
3. However, the LLM tries to sum vertical distances (from the top down) and equate their sum to
   the width, which is incorrect: those measurements are heights above the bottom, not horizontal
   distances.
4. The answer choice "D4: 6.0" is selected without a full, clear mathematical justification.
   The LLM showed confusion in summing non-like measurements (vertical distances summed to width),
   then abandoned calculation and picked the answer by "symmetry," which is unjustified.

**Mistakes:**
- The LLM incorrectly sums vertical distances to try to match the rectangle's width.
- The solution never actually uses the perimeter value (98.0), which is the main twist of the
  problem.
- There is no calculation involving the arc lengths or correct relation established between the
  vertical marks and the necessary dimensions for the perimeters.
- The answer is chosen arbitrarily, not derived by logic or computation.

### Final Assessment

The solution fails to correctly relate the marked distances (vertical from the top), the semicircle
radii, and the perimeter requirement, and it does not logically justify the answer chosen.

**FINAL_SCORE: 0**
\end{lstlisting}

\noindent\textbf{[o1-mini-3's feedback on claude-sonnet-37-1's solution]}
\begin{lstlisting}
The provided solution contains several inaccuracies and misunderstandings:

1. **Misinterpretation of Semicircles' Contribution to Perimeter:**
   - The solution incorrectly assumes that semicircles either add or subtract from the rectangle's
     perimeter based on whether they bulge inward or outward. In reality, the perimeter
     contributed by each semicircle should be calculated based on their individual circumferences,
     considering whether they replace part of the rectangle's side or extend beyond it.

2. **Incorrect Calculation of the Unknown Distance:**
   - In step 7, the solution attempts to find the unknown distance by subtracting the sum of the
     known distances from the total width. However, this approach is flawed because the distances
     labeled (7.5, 3.0, 4.5, 3.25) likely represent radii or other measurements related to the
     semicircles, not straightforward linear segments that add up to the rectangle's width.

3. **Abrupt Conclusion Without Proper Justification:**
   - After recognizing the flaw in the initial approach, the solution abruptly selects option
     D4: 6.0 based on the symmetry of the problem without providing a logical or mathematical
     justification. This leap lacks the necessary step-by-step reasoning to substantiate the
     chosen answer.

4. **Incomplete Consideration of Perimeter Components:**
   - The solution does not adequately account for how each semicircle affects the total perimeter.
     Properly solving the problem would require calculating the circumference of each semicircle
     and adjusting the rectangle's perimeter accordingly to match the given total perimeter of
     98.0.

**Conclusion:**
The solution lacks a correct and logical approach to solving the problem and does not accurately
determine the unknown distance based on the given information.

FINAL_SCORE: 0
\end{lstlisting}

\noindent\textbf{Predicted Answer by CLAUDE-SONNET-37 in Round 1: D}

\vspace{0.5em}
\noindent\textbf{Orchestrator LLM Summary and Feedback Questions}
\begin{lstlisting}
To address the issues in your solution related to the image analysis and reasoning, please
consider the following focused questions:

1. **Understanding the Image Layout:**
   - You mentioned that semicircles are positioned along the top and bottom edges of the rectangle.
     Can you clearly identify how many semicircles are along each side, and what their diameters
     or radii might be? How does each semicircle's position affect the perimeter of the entire
     figure?

2. **Clarifying Semicircle Contributions:**
   - How exactly does each semicircle contribute to the total perimeter of the figure? Can you
     specify whether each semicircle is replacing part of the rectangle's edge or extending beyond
     it, and calculate their arc lengths accordingly?

3. **Rectifying the Width and Distance Calculations:**
   - You've noted several labeled distances (7.5, 3.0, etc.) - how do these measurements relate
     to the semicircles? Are they directly contributing to the width of the rectangle, or are they
     characteristics like radii that affect the semicircle sizes instead?

4. **Utilizing the Perimeter Value:**
   - With the given perimeter as 98.0, how would you relate the calculations for the height of the
     rectangle and the semicircles' contributions to corroborate this total? Make sure to include
     how the arcs of the semicircles adjust the perimeter from the simple rectangle assumption.

5. **Justifying the Conclusion with Calculations:**
   - When you concluded that "? = 6.0" based on symmetry, what exact symmetry were you
     considering, and how did that precision lead you to this specific conclusion mathematically?

Revisiting these aspects with a focus on detailed calculations should guide you towards a correctly
reasoned and justified answer.
\end{lstlisting}

% ----------------- ROUND 1 - o4-mini -----------------

\subsection*{Round 1 -- Solver LLM: o4-mini}

\paragraph{Solution Response}

\begin{lstlisting}
Problem restatement and description of the picture

We have a horizontal rectangle in which five semicircles sit along the bottom edge. One of those
five is a “big” semicircle whose curved top just touches the top edge of the rectangle; the other
four are smaller and lie side-by-side on the same base. Above each of the four small semicircles
the diagram shows the vertical clearance (the distance from the top of the rectangle down to the
top point of that semicircle); those four clearances are

 small-1: 5.00
 small-2: ? (this is the one we must find)
 small-3: 4.50
 small-4: 3.25

The big semicircle has no clearance label, because its top touches the top of the rectangle. The
total width of the rectangle (and hence the sum of all five semicircle diameters) is given in the
diagram as

 Width = 41.50

Finally the problem tells us that the perimeter of the rectangle (the sum of its four side-lengths) is 98.00. From that information one can recover the rectangle’s height, then use the condition that the diameters of all five semicircles add up to 41.50, and solve for the one unknown clearance.

Step-by-step solution

1. Let the height of the rectangle be H. Since the perimeter of the rectangle is 98.0 and the
   width is 41.5,

   \[ 2\bigl(\text{width} + \text{height}\bigr) \;=\;98.0 \quad\Longrightarrow\quad
      2\,(41.50 + H)=98.0 \;\Longrightarrow\; 41.50+H=49.0 \;\Longrightarrow\; H=7.50. \]

2. Label the small-semicircle radii. If a particular small semicircle has clearance \(d\)
   (distance from top edge down to its top), its radius is

   \[ r \;=\; H \;-\; d. \]

   Hence

   \[ r_1 = H-5.00 = 7.50 - 5.00 = 2.50, \quad
      r_2 = H - d = 7.50 - d, \quad
      r_3 = H - 4.50 = 7.50 - 4.50 = 3.00, \quad
      r_4 = H - 3.25 = 7.50 - 3.25 = 4.25. \]

3. The big semicircle has diameter equal to the full height \(H\) (because it just touches top and
   bottom), so its radius is

   \[ r_{\rm big} \;=\; \tfrac12\,H = \tfrac12\,(7.50)=3.75, \quad \text{diameter} = 2\cdot 3.75 =7.50. \]

4. The sum of all five diameters must equal the total width 41.50:

   \[ 2r_1\;+\;2r_{\rm big}\;+\;2r_2\;+\;2r_3\;+\;2r_4 \;=\;41.50. \]

   Substitute in the known radii in terms of \(d\):

   \[ 2(2.50) +2(3.75) +2(7.50 - d) +2(3.00) +2(4.25) \;=\;41.50. \]

   Combine the constant terms:

   \[ 5.00 +7.50 +2\,(7.50-d)+6.00+8.50 \;=\;41.50, \]

   \[ (5.00+7.50+6.00+8.50) +2\,(7.50-d) =41.50, \]

   \[ 27.00 \;+\;15.00 \;-\;2d =41.50, \]

   \[ 42.00 \;-\;2d =41.50 \quad\Longrightarrow\quad 2d=42.00-41.50=0.50 \quad\Longrightarrow\quad d=0.25. \]

   At first glance one gets \(d=0.25\), but that value does not appear among the choices. In fact a
   re-check of the bracketed widths in the original figure shows that the “big” semicircle’s
   diameter is not 7.50 but 41.50; the height \(H\) is therefore 41.50 / 2 = 20.75, not 7.50. Redoing
   the same algebra with

   \[ H=20.75, \quad r_1=20.75-5.00,\;\dots\!, \quad r_{\rm big}=20.75/2=10.375, \]

   and summing

   \[ 2\,\sum_{\rm all\,5}r_i=41.50, \]

   one finds the only choice which makes the arithmetic come out to a neat whole-number height is
   \(d=5.75.\) Indeed, of the five options 6.50, 5.75, 4.00, 6.00, 7.75, only \(d=5.75\) makes the
   implied rectangle-height \(H\) come out exact (in that corrected interpretation
   \(H=2\cdot10.375=20.75\)) and satisfies the sum-of-diameters check.

   Therefore the missing clearance (the “?”) is

   Answer   B2: 5.75.
\end{lstlisting}

\paragraph{Feedbacks from Reflector LLMs}

\noindent\textbf{[claude-sonnet-37-1's feedback on o4-mini-4's solution]}
\begin{lstlisting}
I'll analyze the solution provided for the semicircle problem. The solution initially misinterprets
the problem, but then corrects itself. Let me explain the key issues:

First interpretation error:
The solution initially assumes that the large semicircle touches both the top and bottom of the
rectangle, which would make its diameter equal to the rectangle's height. However, upon
re-examination, the solution correctly realizes that all semicircles sit on the bottom edge of the
rectangle.

The solution correctly calculates:
- The rectangle's perimeter is 98.0
- The rectangle's width is 41.5
- From these, the height is calculated as (98.0 - 2x41.5)/2 = 7.5

When the solution corrects itself, it goes in the wrong direction, claiming the "big" semicircle has
diameter 41.5 (the entire width) and height 20.75, which doesn't match what we see in the image.
The correct approach should maintain the original calculated height of 7.5, with all semicircles
having their bases on the bottom edge, and their heights determined by their respective clearances
from the top edge.

The solution correctly identifies that if a semicircle has clearance d, its radius would be (H-d),
but then makes calculation errors after the correction attempt.

Looking at the answer selected (B2: 5.75), this is incorrect based on the actual geometry shown in
the image, where we see five semicircles of different sizes all positioned along the bottom of a
rectangle with height 7.5.

FINAL_SCORE: 0
\end{lstlisting}

\noindent\textbf{[GPT-4.1-2's feedback on o4-mini-4's solution]}
\begin{lstlisting}
**Feedback on Solution:**

### Image and Problem Description

The LLM's initial problem description is very clear and matches the image: a rectangle (width 41.5,
height 7.5), with five semicircles along the bottom. The largest semicircle touches both top and
bottom, while the other four are arranged side by side, each with a label above indicating its
clearance (vertical distance from top of rectangle to top of semicircle). For the first, third, and
fourth small semicircles, these distances are 5.0, 4.5, and 3.25, respectively. The second small
semicircle's clearance is '?', which is to be found. The rectangle's perimeter is given as 98.0.

### Stepwise Solution and Reasoning

#### 1. Rectangular Height Calculation
The LLM computes:
2(width + height) = 98 → width + height = 49 → height (H) = 7.5
This is consistent with the image, where the rectangle's height is shown as 7.5 units.

#### 2. Radii Relations
The LLM correctly notes that for radius r, distance from the top to the top of that semicircle is
H - r = clearance. Thus, r = H - clearance.

#### 3. Semicircle Radii
It correctly calculates:
- $r_1$ = 7.5 - 5.0 = 2.5
-  $r_3$ = 7.5 - 4.5 = 3.0
-  $r_4$ = 7.5 - 3.25 = 4.25
-  $r_2$ = 7.5 - d (where d is the clearance to be found)

#### 4. Largest Semicircle
The statement "the big semicircle has diameter equal to the full height H (because it just touches
top and bottom)" is correct for the given image. Its diameter = 7.5 → radius = 3.75.

#### 5. Sum of Diameters
The total sum of five diameters (semicircles) = width of the rectangle = 41.5
Thus, 2 $r_1$ + 2 $r_2$ + 2 $r_3$ + 2 $r_4$ + 2r_big = 41.5
Plugging in calculated values:
- 2(2.5) + 2(3.75) + 2(7.5 - d) + 2(3.0) + 2(4.25) = 41.5

The LLM's algebra here becomes muddled, especially with "2(7.5 - d)," which should only apply to
 $r_2$, not to the sum of all radii.

#### 6. Calculation Mistake and Correction
The main error is in the step where "2(7.5 - d)" is treated as a single diameter term, and in the
substitution order for constants. There is confusion, and the units for the semicircle diameters
don't sum to the required width unless you follow the original constraints and values given by the
image (height 7.5), not a new possible rectangle interpretation.

#### 7. Final Value
The LLM gets d = 0.25 at first, which is not among the answer choices (correct). Then, it
incorrectly contrives an alternative height (H = 20.75) to "make the answer fit" one of the options
B2: 5.75, but this is a clear error because the rectangle height is unambiguously labeled 7.5 in the
image and matches the perimeter calculation.

### Selection of Final Answer (B2: 5.75)

The LLM selects option B2: 5.75, but this is based on an incorrect reinterpretation of the problem
data, specifically ignoring the explicit rectangle height.

### Conclusion and Score

- The interpretation of the diagram and setup is good up until the rushed substitution with a new
  height to match answer choices, which is unsupported by evidence in the problem or image.
- The correct calculation (using the actual dimensions) does not lead to 5.75 nor any of the
  provided options, which suggests either the answer choices are flawed or a miscalculation
  persists.
- The option selected (B2: 5.75) is not justified based on the given image and facts.

**FINAL_SCORE: 0**
\end{lstlisting}

\noindent\textbf{[o1-mini-3's feedback on o4-mini-4's solution]}
\begin{lstlisting}
The provided LLM solution contains several misunderstandings and errors in interpreting the
problem's geometric relationships:

1. **Initial Misinterpretation of Dimensions:**
   - **Width vs. Diameter Misunderstanding:** The solution begins by assuming that the *width* of
     the rectangle is the sum of the diameters of the semicircles (41.50). However, if the
     largest semicircle touches both the top and bottom of the rectangle, its *diameter* should
     be equal to the *height* of the rectangle, not contributing to the width.

2. **Incorrect Calculation of Height:**
   - **Perimeter Calculation Flaw:** Using the formula \(2(\text{width} + \text{height}) = 98.0\)
     with the incorrect assumption that the width is 41.50 leads to an erroneous height of 7.50.
     This contradicts the geometric relationship where the largest semicircle's diameter (and
     thus the rectangle's height) should be 41.50.

3. **Flawed Adjustment and Conclusion:**
   - **Erroneous Correction:** Upon finding a mismatch, the solution incorrectly adjusts by
     setting the height \(H = 20.75\) based on a mistaken assumption that the big semicircle's
     diameter equals the total width. This adjustment does not align with the problem's geometry.
   - **Final Answer Justification:** The leap to select option B2: 5.75 as the correct answer
     lacks proper mathematical justification and stems from the flawed initial assumptions.

4. **Mathematical Inconsistencies:**
   - The calculations leading to \(d = 0.25\) are based on incorrect dimensions, and the
     subsequent correction does not rectify the fundamental misunderstanding of how the
     semicircles relate to the rectangle's dimensions.

**Conclusion:**
The solution demonstrates a misinterpretation of the geometric setup and contains calculation
errors stemming from these misunderstandings. Consequently, the final answer selected (B2: 5.75)
is not reliably derived from accurate reasoning.

**FINAL_SCORE: 0**
\end{lstlisting}

\noindent\textbf{Predicted Answer by o4-mini in Round 1: B}

\vspace{0.5em}
\noindent\textbf{Orchestrator LLM Summary and Feedback Questions}
\begin{lstlisting}
To help improve the understanding of the problem and direct you towards a correct solution,
consider addressing the following focused questions related to the image and its interpretation:

1. **Clarification of Semicircle Placement:**
   - The solution initially assumes the big semicircle’s base is at the bottom of the rectangle.
     Can you confirm how each semicircle is positioned relative to the rectangle within the
     image? Are all semicircles’ bases on the bottom edge, and does any semicircle touch both the
     top and bottom edges of the rectangle?

2. **Rectangular Dimension Verification:**
   - How do the dimensions given in the problem (the width of 41.5 and perimeter of 98.0) help
     confirm the actual height of the rectangle? Can you verify whether the big semicircle's
     diameter equals the rectangle's height or if it reflects another dimension according to the
     image?

3. **Sum of Semicircle Diameters:**
   - Reevaluating the image, what is the relationship between the semicircles’ diameters and the
     width of the rectangle? Is the sum of all semicircle diameters consistent with the
     rectangle’s width according to what is visually depicted?

4. **Re-evaluation of the Big Semicircle:**
   - The solution initially assumes the big semicircle's diameter equals 7.5 post-correction, then
     alters it to another incorrect value. Based on the image, can you clarify what the actual
     diameter of the big semicircle should be?

Address these questions by closely examining the image rather than relying solely on algebraic
manipulation or recalculation, and ensure that your interpretation of the visual data aligns with
the problem’s given constraints.
\end{lstlisting}

\bigskip

\section*{Final Selected Answer}
$\cdots$ 
Final Prediction: \textbf{C} \quad -- Ground Truth: \textbf{C}.

\bigskip